\documentclass[conference]{IEEEtran}
\IEEEoverridecommandlockouts
\usepackage{cite}
\usepackage{amsmath,amssymb,amsfonts}
\usepackage{algorithm, algorithmic}
\usepackage{graphicx}
\usepackage{textcomp}
\usepackage{marvosym}
\usepackage{subfigure}
\usepackage{booktabs}
\usepackage{multirow}
\usepackage{listings}
\usepackage{colortbl}
\usepackage[dvipsnames]{xcolor}
\usepackage{hyperref}
\hypersetup{hidelinks=False}

\def\BibTeX{{\rm B\kern-.05em{\sc i\kern-.025em b}\kern-.08em
    T\kern-.1667em\lower.7ex\hbox{E}\kern-.125emX}}
\begin{document}

\title{TriMLP: Revenge of a MLP-like Architecture in Sequential Recommendation}

\author{
    \IEEEauthorblockN{
    Yiheng Jiang\textsuperscript{1,2},
    Yuanbo Xu\textsuperscript{1,2 \Letter},
    Yongjian Yang\textsuperscript{1,2},
    Funing Yang\textsuperscript{1,2},
    Pengyang Wang\textsuperscript{3},
    Hui Xiong\textsuperscript{4,5}}
    \IEEEauthorblockA{\textsuperscript{1}\textit{Lab of Mobile Intelligence (MIC), Jilin University, Changchun, China}}
    \IEEEauthorblockA{\textsuperscript{2}\textit{College of Computer Science and Technology, Jilin University, Changchun China}}
    \IEEEauthorblockA{\textsuperscript{3}\textit{Department of Computer and Information Science, University of Macau, Macao SAR, China}}
    \IEEEauthorblockA{\textsuperscript{4}\textit{Thrust of Artificial Intelligence, Hong Kong University of Science and Technology (Guangzhou), Guangzhou, China}}
    \IEEEauthorblockA{\textsuperscript{5}\textit{Department of Computer Science and Engineering, Hong Kong University of Science and Technology, Hong Kong SAR, China}}
    \IEEEauthorblockA{
    \textsuperscript{1,2}\texttt{jiangyh22@mails.jlu.edu.cn};
    \textsuperscript{1,2}\texttt{\{yuanbox, yyj, yfn\}@jlu.edu.cn};\\
    \textsuperscript{3}\texttt{pywang@um.edu.mo};
    \textsuperscript{4,5}\texttt{xionghui@ust.hk}}}

\maketitle

\begin{abstract}
Sequential recommenders concentrate on modeling the transmission patterns shrouded in sequences of historical user-item interactive behaviors (or referred as token) and inferring dynamic preferences over candidate items. Fueled by diverse advanced neural network architectures like RNN, CNN and Transformer, existing methods have enjoyed rapid performance boost in the past years. Recent progress on MLP lights on an efficient method, token-mixing MLP, to establish contact among tokens. However, due to the unrestricted cross-token communications, i.e., information leakage issue, caused by the inherent fully-connection structure, we find that directly migrating these modern MLPs in recommendation task would neglect the chronological order of historical sequences and lead to subpar performances. In this paper, we present a \underline{MLP}-like architecture for sequential recommendation, namely TriMLP, with a novel \underline{Tri}angular Mixer for cross-token communications. In designing Triangular Mixer, we simplify the cross-token operation in MLP as the basic matrix multiplication, and drop the lower-triangle neurons of the weight matrix to block the anti-chronological order connections from future tokens. Accordingly, the information leakage issue can be remedied and the prediction capability of MLP can be fully excavated under the standard auto-regressive mode. Take a step further, the mixer serially alternates two delicate MLPs with triangular shape, tagged as global and local mixing, to separately capture the long range dependencies and local patterns on fine-grained level, i.e., long and short-term preferences. Empirical study on 12 datasets of different scales (50K\textasciitilde 10M user-item interactions) from 4 benchmarks (Amazon, MovieLens, Tenrec and LBSN) show that TriMLP consistently attains promising accuracy/efficiency trade-off, where the average performance boost against several state-of-the-art baselines achieves up to 14.88\% with 8.65\% less inference cost. Our code is available at \href{https://github.com/jiangyiheng1/TriMLP}{https://github.com/jiangyiheng1/TriMLP}.
\end{abstract}


\vspace{-2.5pt}

\section{Introduction}
Sequential recommendation processes sequences of historical user-item interactive behaviors (or referred as tokens in this paper), concentrates on mining dependencies among tokens and inferring preferences over time, and provides pleasing suggestions \cite{dualrec}. The performances of sequential recommenders are closely tied with the reliant neural network architectures, which serve as essential components for establishing contact among tokens and capturing transformation patterns.

\begin{figure}[t]
\centering
	\includegraphics[scale=0.3025]{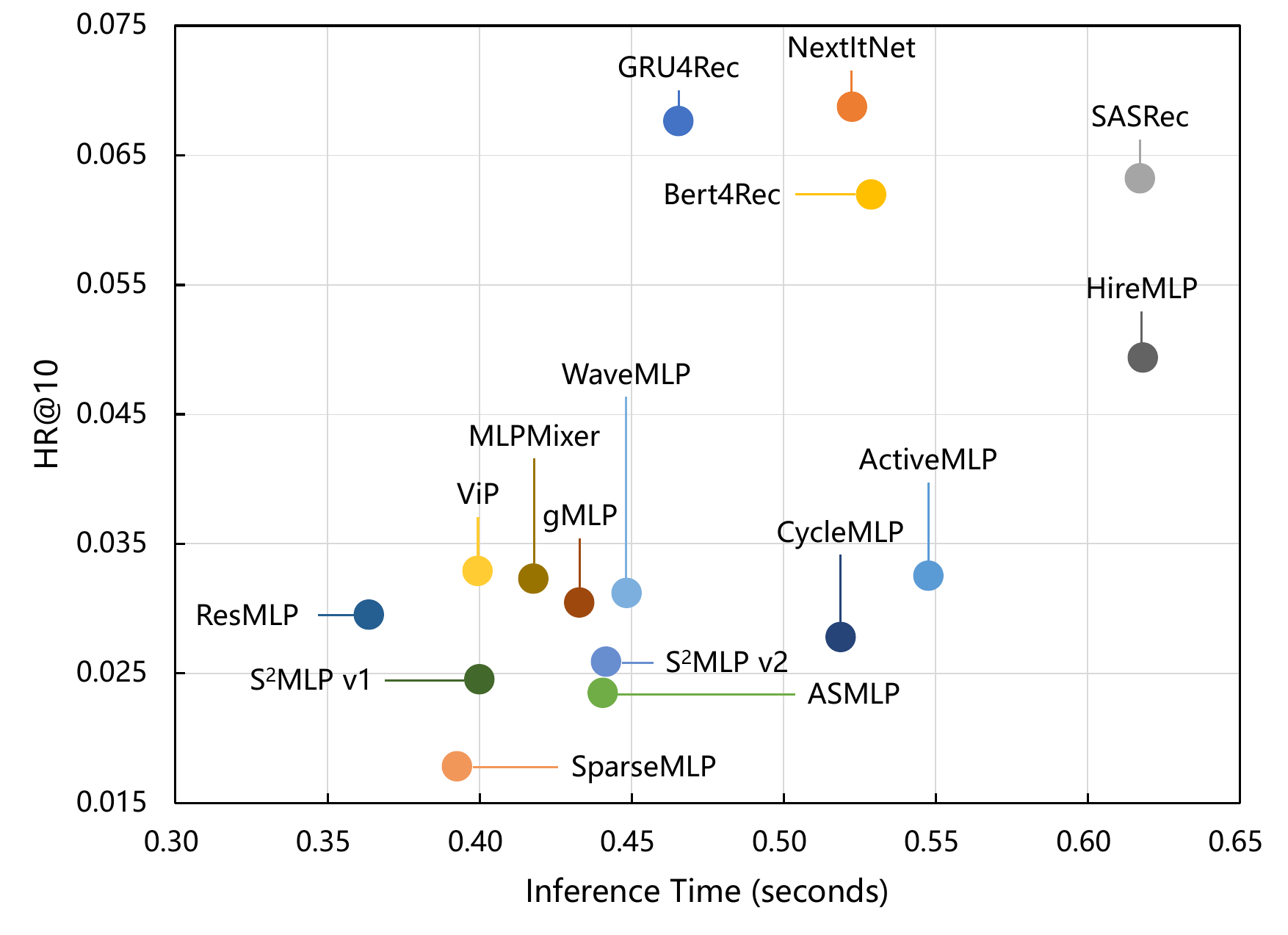}
\caption{Accuracy/Efficiency traded-off on \texttt{QB-Video}. Along the vertical axis, the higher, the better recommendation performance; along the horizontal axis, the more left, the less inference cost.}
\label{pilot exp 1}
\vspace{-20pt}
\end{figure}

Innovations in neural network architectures have consistently played a major role in sequential recommendation. Recurrent neural network (RNN)-based sequential recommenders, represented by \cite{gru4rec, flashback}, transmits the information in tokens step-by-step. Methods like \cite{nextitnet, caser}, adopting convolutional neural network (CNN), aggregate the local spatial features with sliding filters. Credited to the superb adaptability with sequential tasks, Transformer architecture \cite{transformer} that dynamically re-weights and integrates tokens through the self-attention mechanism has become the \textit{de-facto} backbone in modern sequential recommenders \cite{sasrec, bert4rec, stisan}.

In most recent, ``retrospective'' research on purely multi-layer perceptron-based (MLP) models, pioneered by MLPMixer \cite{mlpmixer} and ResMLP \cite{resmlp}, investigates an conceptually simple and computationally efficiency idea to realize the cross-token communication, namely token-mixing MLP, where tokens interact with each other independently and identically across dimensions \cite{mlp-survey}. In terms of structural properties, token-mixing MLP holds the global reception field as Transformer while reserves the learned sequential dependency as static weights like RNN and CNN. Despite token-mixing MLP is originally derived from the vision community, it is intuitive that such MLP owns promising potential for sequential tasks.

\begin{figure}[t]
\centering
	\includegraphics[scale=0.23]{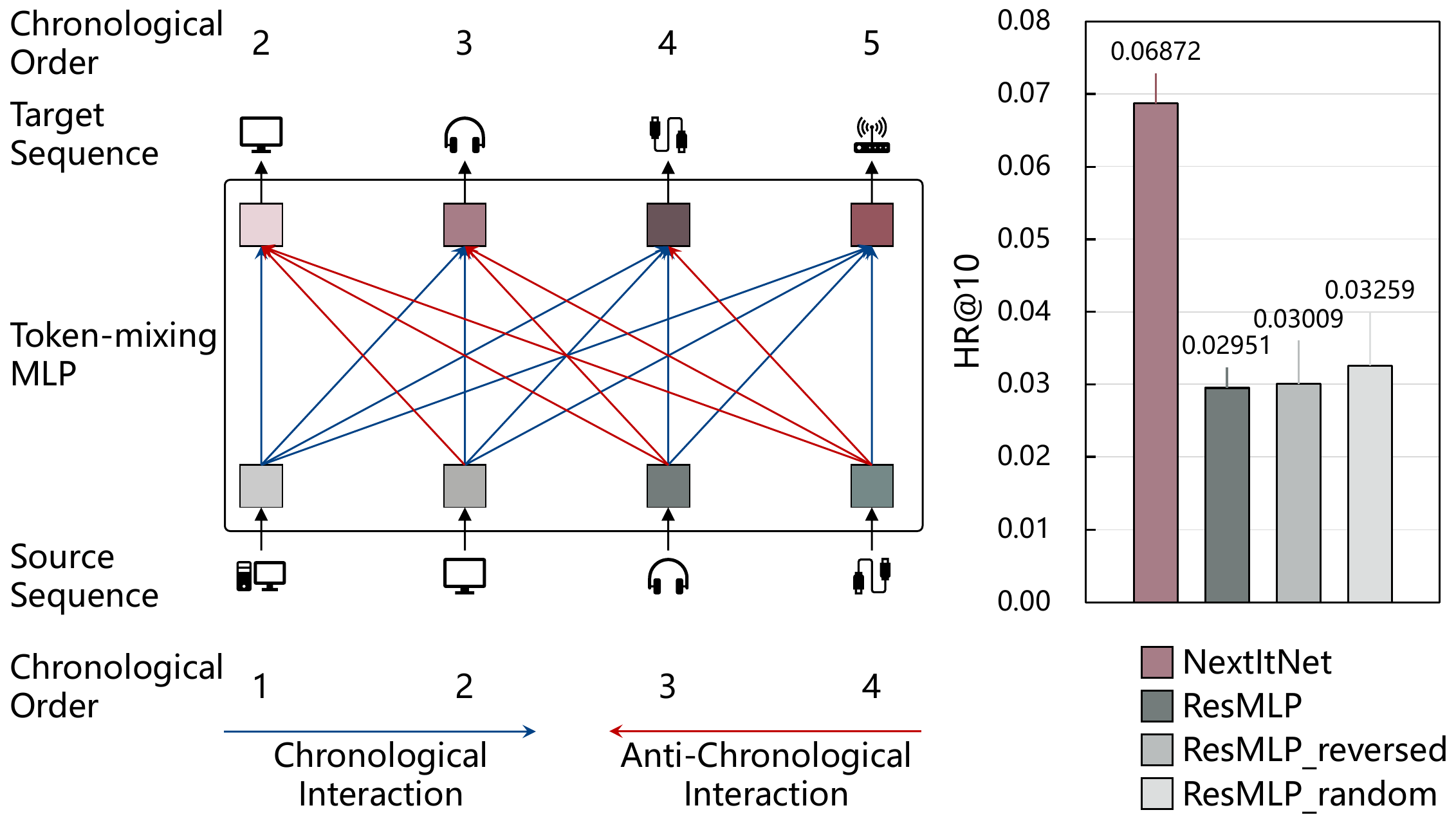}
\caption{The left part illustrates the unrestricted cross-token interactions in MLP. The blue arrows denote the interaction in chronological order where the current token can only attend to itself and previous tokens, while the red ones are in contrast which lead to the information leakage. The right histogram reveals that the fully-connected MLP is insensitive to the sequence order.}
\label{pilot exp 2}
\vspace{-20pt}
\end{figure}

However, the empirical observation draws apart from the exception when we refer and explore token-mixing MLP in sequential recommendation. Following the implementations in \cite{tenrec}, we reproduce RNN-based GRU4Rec \cite{gru4rec}, CNN-based NextItNet \cite{nextitnet}, Transformer-based SASRec \cite{sasrec} and Bert4Rec \cite{bert4rec} and various modern MLP-based models \cite{mlpmixer, resmlp, vip, gmlp, wavemlp, sparsemlp, asmlp, hiremlp, activemlp, s2mlpv1, s2mlpv2, asmlp} on \texttt{QB-Video} \cite{tenrec}. As shown in Figure \ref{pilot exp 1}, albeit most MLPs inherit the merit of efficiency (along the horizontal axis), their recommendation performances stand far behind other neural network architecture-based sequential recommenders, e.g. the strongest HireMLP lags behind NextItNet by 28.19\% on the metric of HR@10. 

We argue such subpar performance is oriented from the inherent \textit{Fully-Connection Design} in MLPs. As the example in the left part of Figure \ref{pilot exp 2}, a user has historically interacted with \texttt{1: Computer}$\rightarrow$\texttt{2: Monitor}$\rightarrow$\texttt{3: Headphone}$\rightarrow$\texttt{4: USB}, and the MLP aims at predicting the $i+1$ th token at step $i$ under the standard auto-regressive training mode. Unfortunately, except for the cross-token interactions in chronological order (denoted as blue arrows in Figure \ref{pilot exp 2}) consistent with the natural behavior pattern \cite{gru4rec, caser, sasrec, nextitnet}, the fully-connected MLP inevitably conducts the anti-chronological ones (red arrows) which would leak the future information to the current time step and suppress the prediction ability throughout the training procedure \footnote{We also consider utilizing the bidirectional attribute of MLP and conduct the auto-encoding training mode \cite{bert}. See Section \ref{atrg vs atec} for more details.}. To further verify whether or not MLP is sensitive to the sequence order, we train ResMLP on \texttt{QB-Video} with different ordered sequences, i.e., chronological, reserved and random. As summarized in the histogram of Figure \ref{pilot exp 2}, the results of MLPs are sharing inferior performances to CNN-based NextItNet with negligible standard deviations. It supports our opinion about the incompatibility between MLP and auto-regressive manner, that the fully-connection nullifies the capacity of implicitly encoding and differentiating the position of each token \cite{resmlp}.

In this paper, we propose to build a MLP-like architecture for sequential recommendation, with the aim of persisting the computationally efficiency advantage, gearing to the auto-regressive training fashion and catching up to the performances obtained using advanced neural network architectures.

In designing the MLP-based token mixer, we present Triangular Mixer to remedy the issues brought by fully-connection. It is inspired by the use of masking strategy in Transformer-based methods \cite{stisan}. In principle, since the cross-token interactions endowed by MLP can be simplified as the matrix multiplication, undesirable interactions can be forbidden by disabling specific neurons in MLP. In practice, we drop the lower-triangle elements in the weight matrix of MLP to block the connections from future tokens and ensure that each token can only attend to itself and previous ones. Naturally, the information leakage issue is avoided, and the potential of MLP can be fully excavated under the auto-regressive training.

Take a step further, since MLP with global reception field excels in modeling the long-range relations among tokens while fails in capturing local patterns \cite{repmlpnet}, we derive two mixing layers based on the above delicate MLP with triangular shape, tagged as global mixing and local mixing. The global mixing follows the vanilla triangular shape and attaches importance to all tokens in sequences for inferring long-term preference. The local one further drops specific upper-triangle neurons of weight matrix and treats the input sequence as multiple non-overlapped independent sessions with equal length. Specifically, the shape of active neurons is converted as several isosceles right sub-triangles arranged along the main diagonal whose sides are equal to the session length. Each sub-triangle is responsible for capturing the short-term preference contained in the corresponding session. Triangular Mixer serially alternates global mixing and local mixing for the fine-grained sequential dependency modeling.

To this end, we present a \underline{MLP}-like sequential recommender TriMLP based on the proposed \underline{Tri}angular Mixer. In summary, our major contributions can be listed as follows:

\begin{itemize}
\item We refer and explore the idea of all-MLP architecture in sequential recommendation. To the best of our knowledge, we are the first to empirically point out that the fully-connection in MLP is not compatible with the standard auto-regressive training mode.
\item We present a MLP-like sequential recommender, namely TriMLP, with a novel Triangular Mixer which endows the chronological interactions among tokens.
\item We put forward Triangular Mixer with the global mixing and local mixing, to reconcile the long-rang dependencies and local patterns in sequences.
\item We evaluate TriMLP with 12 datasets of different scales from 4 benchmarks (MovieLens, Amazon, Tenrec and LBSN) which contain 50K\textasciitilde10M user-item interactive behaviors. The experimental results demonstrate that TriMLP attains stable and promising accuracy/efficiency trade-off over all validated datasets, i.e., averagely surpasses the performance of several state-of-the-art baselines by 14.88\% with 8.65\% less inference cost.
\end{itemize}

\section{Related Work}

\subsection{Sequential Recommendation}
Sequential recommendation aims at capturing dynamic preferences from sequences of historical user-item interactive behaviors and providing pleasant suggestions \cite{dualrec}. Building upon technological breakthroughs in the past decade \cite{ncf, deepfm, din, dien}, this field has ushered a new era of deep learning. Hidasi et al. \cite{gru4rec} leveraged RNN to model the sequential dependency which transmits the information contained in token step-by-step. The spatial local information aggregation in CNN also benefits sequential recommenders \cite{caser, nextitnet}. SASRec \cite{sasrec} and Bert4Rec \cite{bert4rec} separately employed unidirectional and bidirectional Transformer-base encoder \cite{transformer, bert} to dynamically extract the relationship between target and relevant items. Towards all-MLP methods, FMLP4Rec \cite{f-mlp4rec} referred the learnable filter-based denoising manner and encoded sequential patterns with Fourier transform, and MLP4Rec \cite{mlp4rec} incorporated contextual information (e.g., item characteristics) into the MLPMixer architecture \cite{mlpmixer}. In separate lines of research, \cite{srgnn, gcsan} utilized graph neural network (GNN) to enhance item representations,\cite{hirestructure} adopted hierarchical structures, \cite{dataagument1, dataagument2} introduced data augmentation and \cite{pretrain1, pretrain2, pretrain3} exploited pre-training techniques. 

TriMLP architecture focuses on improving the primary sequential modeling capacity of MLP under the essential auto-regressive training mode without assistance of any auxiliary information. Credited to the triangular design, TriMLP successfully merges the performance gap between MLP and other advanced neural network-based sequential recommenders.

\subsection{Toke-mixing MLP}
Since the pioneering MLPMixer \cite{mlpmixer} and ResMLP \cite{resmlp} have been proposed in the early 2020s, all-MLP models are staging a comeback in vision community. These models rely on the novel MLP-based token mixer where tokens interact independently and identically across the channel dimension. Due to the simple concept with less intensive computation, such deep MLP models have stirred up a lot of interest \cite{mlp-survey} and derive a surge of variants. According to the dimensions of mixing tokens, these variants can be divided into three categories: (i) employing both the axial direction and the channel dimension \cite{raftmlp, vip, sparsemlp, dynamixer, wavemlp} which proposes to orthogonally decompose the cross-token mixing, maintain long-range dependency and encode cross-token relation along axial directions, (ii) considering only the channel dimension \cite{s2mlpv1, s2mlpv2, asmlp, cyclemlp, hiremlp, msmlp, activemlp}  which aligns features at different spatial locations to the same channel by axial shifting, and then interacting with spatial information through channel projection. (iii) reserving the entire spatial and channel dimensions \cite{mlpmixer, resmlp, feedforward, gmlp, repmlpnet, sparsemlp} which retains the global reception field and channel projection.

In the pilot experiments, we empirically investigate that the inherent fully-connection design in MLP is incompatible with sequential tasks especially under the auto-regressive training fashion. In contrast, the proposed Triangular Mixer provides a simple, effective and efficient alternative to remedy the issue.

\section{Preliminary}

\subsection{Basic Definition}
Let $\mathcal{U}$ and $\mathcal{I}$ denote the user and item set, respectively. Accordingly, we have the following basic definitions.

\textbf{Definition 1: (User-Item Interactive Behavior)} A user-item interactive behavior, or referred as token in this paper, is represented as a triplet $x=\left\langle u, i, t\right\rangle$, which denotes that the user $u\in\mathcal{U}$ interacted with the item $i\in\mathcal{I}$ at time $t$.

\textbf{Definition 2: (Historical Sequence)} A historical sequence, tagged as $X^u=x_1 \rightarrow x_2 \rightarrow \cdots \rightarrow x_{|X^u|}$, chronologically records $|X^u|$ user-item interactive behaviors of the user $u$.

\subsection{Problem Statement}
\textbf{Sequential Recommendation} Given the specific user $u$ and his/her historical sequence $X^u$. Sequential recommendation problem infers the dynamic preferences and provides the top-K recommendation list, which contains $K$ items that the user might be most likely to interact in the next time step. It can be formulated as the following equation,
\begin{equation}
\mathcal{F}\left(X^u\right) \rightarrow TopK^u,
\end{equation}
where $TopK^u$ denotes the top-K recommendation list and $\mathcal{F}\left(\cdot\right)$ is the abstract symbol of any sequential recommender.

\section{Methodology}
\subsection{Architecture Overview}
\begin{figure}[t]
	\centering
	\includegraphics[scale=0.4125]{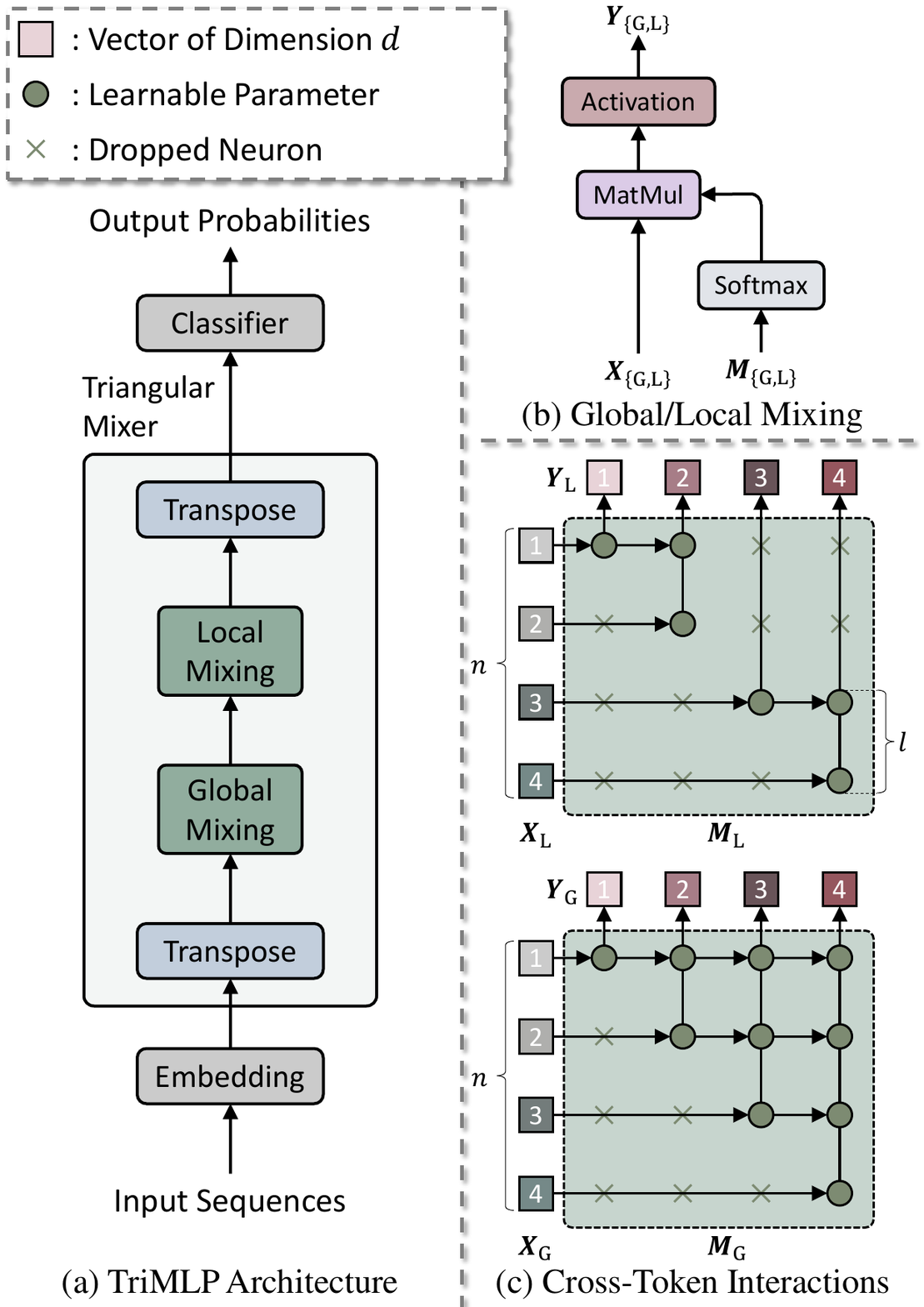}
	\caption{(a) depicts the proposed MLP-like architecture TriMLP. (b) reveals the details of global and local mixing in Triangular Mixer. (c) presents an illustrative example with sequence length $n=4$ and session length $l=2$ to explain the cross-token interactions in global and local mixing.}
	\label{architecture}
	\vspace{-20pt}
\end{figure}

The macro overview of TriMLP architecture is depicted in the Figure \ref{architecture} (a). TriMLP takes a historical sequence of $n$ user-item interactive behaviors (or tokens) as input, where $n$ is the maximum sequence length. The tokens are independently pass through the Embedding layer to form the $d$-dimension sequence representation matrix. The resulting embedding is then fed into the Triangular Mixer to produce cross-token interactions. The Classifier takes these encoded representations as input, and predicts the probabilities over all candidate items.

\subsection{Embedding}
Considering that the historical sequences of different users are inconsistent in length, we set the maximum sequence length to $n$. Following the operation in \cite{sasrec}, we split the longer sequences into several non-overlapping sub-sequences of length $n$. For the shorter ones, we repeatedly add the ``padding'' token in the head until their lengths grow to $n$. For the clarity and conciseness, we omit the superscript that denotes the specific user $u$, and the embedding layer can be formulated with the following equation,
\begin{equation}
{\rm Embed}(X)\rightarrow\emph{\textbf{X}}\in\mathbb{R}^{n \times d},
\end{equation}
where \emph{\textbf{X}} is the sequence representation matrix. Note that the padding tokens are encoded with constant zero vectors \cite{stisan} and excluded from the gradient update.

\begin{algorithm}[t]
\caption{PyTorch-like Pseudo-code of Triangular Mixer}
\label{code}
\begin{lstlisting}
import torch
import torch.nn as nn
# n: input sequence length 
# l: session length
# s: number of sessions
# n = l * s
def generate_kernel(n, l, s):
  mask = torch.zeros([n, n])
  for i in range(0, s):
    mask[i*l: i*l+l, i*l: i*l+l] = torch.ones(l, l)
  M_G = torch.triu(torch.ones([n, n]))
  M_L = M_G.masked_fill(mask == 0.0, 0)
  M_G = nn.parameter.Parameter(M_G, requires_grad=True)
  M_L = nn.parameter.Parameter(M_L, requires_grad=True)
  return M_G, M_L

class TriangularMixer(nn.Module):
  def __init__(self, n, l, s, act):
    super().__init__()
    assert l * s == n
    self.M_G, self.M_L = generate_kernel(n, l, s)
    self.act = act

  def forward(self, X):
    # X: input sequence embedding, [b, n, d]
    X_G = X.permute(0, 2, 1)
    Y_G = self.act(torch.matmul(X_G, self.M_G).softmax(dim=-1))
    Y_L = self.act(torch.matmul(Y_G, self.M_L).softmax(dim=-1))
    Y = Y_L.permute(0, 2, 1)
    return Y
\end{lstlisting}
\end{algorithm}

\subsection{Triangular Mixer}
Triangular Mixer endows the cross-token communication in strict compliance with chronological order. As shown in Figure \ref{architecture} (a), the mixer takes as input the sequence representations \emph{\textbf{X}}, and encodes the sequential dependency through the global mixing layer and local mixing layer, successively. Formulaically, it can be expressed as,
\begin{equation}
\emph{\textbf{Y}} = 
{\rm TriMix}(\emph{\textbf{X}})=
{\rm Mix}_{ \rm L}\left({\rm Mix}_{ \rm G}\left(\emph{\textbf{X}}^\top\right)\right)^\top,
\end{equation}
where ``$\top$'' is the matrix transposition and $\emph{\textbf{Y}}\in\mathbb{R}^{n \times d}$ is the encoded sequence representations. The global mixing ${\rm Mix}_{ \rm G}(\cdot)$ injects the long-range sequential dependency and the local mixing ${\rm Mix}_{ \rm L}(\cdot)$ further captures the local patterns. These two mixing layers share the identical structure (as Figure \ref{architecture} (b)), and can be expressed as,
\begin{equation}
\begin{split}
\label{mixing eq}
\emph{\textbf{Y}}_{\rm \left\{G, L\right\}} &= 
{\rm Mix}_{\rm \left\{G, L\right\}}(\emph{\textbf{X}}_{\rm \left\{G, L\right\}})\\
&=
{\rm Act}(
\emph{\textbf{X}}_{\rm \left\{G, L\right\}}
\cdot
{\rm Softmax}(
\emph{\textbf{M}}_{\rm \left\{G, L\right\}}))
\end{split}
\end{equation}
where $\emph{\textbf{M}}_{\{{\rm G, L}\}}\in\mathbb{R}^{n \times n}$ are the mixing kernels in global and local mixing, separately, i.e., the learnable weight matrices in MLPs. Since the cross-token operations in MLP is actually re-weighting and integrating tokens based on the weight matrix, i.e., linear combination, we utilize activation function ${\rm Act}(\cdot)$ to inject the non-linearity. We also adopt ${\rm Softmax}(\cdot)$ to convert the parameters in MLP as the probabilities over tokens. Note that we employ the unified notations in Eq. \ref{mixing eq} for simplicity. Specifically, the input for global mixing $\emph{\textbf{X}}_{\rm G}\in\mathbb{R}^{d \times n}$ is the transposed sequence embedding, and the corresponding output $\emph{\textbf{Y}}_{\rm G}\in\mathbb{R}^{d \times n}$ also serves as the input for local mixing $\emph{\textbf{X}}_{\rm L}\in\mathbb{R}^{d \times n}$. The output of local mixing $\emph{\textbf{Y}}_{\rm L}\in\mathbb{R}^{d \times n}$ is transposed to form the final output of the mixer $\emph{\textbf{Y}}$.

Next, we devote into the details of the delicate $\emph{\textbf{M}}_{\{{\rm G, L}\}}$ in global and local mixing, respectively.

\subsubsection{Global Mixing}
The triangular design in $\emph{\textbf{M}}_{\rm G}$ gets insight from the utilization of mask strategy in Transformer-based sequential recommenders \cite{sasrec, stisan}, which masks the lower-triangular elements in the attentive map to prevent the information leakage. Similarly, we drop the lower-triangular neurons in $\emph{\textbf{M}}_{\rm G}$ to cut off the contact from future tokens. The lower part of Figure \ref{architecture} (c) provides an illustrative example with the input sequence of length $n=4$ to explain the cross-token communication, where ``1, 2, 3, 4'' denote the chronological order. ${\rm Mix}_{ \rm G}(\cdot)$ compels that the 2 nd token can only attend to the previous 1 st token and itself. Mathematically, the $i$ th token in the global mixing interact with each other as

\begin{equation}
\emph{\textbf{Y}}_{{\rm G}_{*,i}}=
\sum^i_{j=1}\emph{\textbf{X}}_{{\rm G}_{*,j}}
\emph{\textbf{M}}_{{\rm G}_{j,i}},
\quad {\rm for}\ i = 1, \cdots n,
\end{equation}
where ``$*$'' denotes any dimension in $d$ and the upper bound $i$ of cumulative sum blocks the information from future tokens. In the premise of avoiding information leakage, global mixing ${\rm Mix}_{ \rm G}(\cdot)$ reserves the global reception field for the long-range sequential dependency.

\subsubsection{Local Mixing}
Based on the aforementioned global triangular design, local mixing ${\rm Mix}_{ \rm L}(\cdot)$ further selectively drops specific upper-triangle neurons to capture the local patterns, calling for the short-term preferences. In principle, it treats the input sequence as $s$ non-overlapping sessions of length $l$ where $n=s\times l$ and forbids the cross-session communications. In practice, the shape of active neurons in $\emph{\textbf{M}}_{\rm L}$ converts into $s$ isosceles right triangles of equal side length $l$, which arrange along the main diagonal. The resulting sub-triangles are responsible for capturing the local patterns contained in corresponding sessions. As the example in Figure \ref{architecture} (c) where the session length $l=2$, the 4 th token attaches importance to the 3 rd token and itself, while the information from the 1 st and 2 nd token is ignored. Generally, the $i$ th token in the local mixing interact with each other as
\begin{equation}
\emph{\textbf{Y}}_{{\rm L}_{*,i}}=
\sum^i_{j=\lceil i/l\rceil}\emph{\textbf{X}}_{{\rm L}_{*,j}}\emph{\textbf{M}}_{{\rm L}_{j,i}},
\quad {\rm for}\ i = 1, \cdots n,
\end{equation}
where the lower bound of cumulative sum, i.e., the round-up operation $j=\lceil i/l\rceil \in [1, s]$, further cuts off the connections from previous sessions.

\begin{figure}[h]
\centering
\includegraphics[scale=0.15]{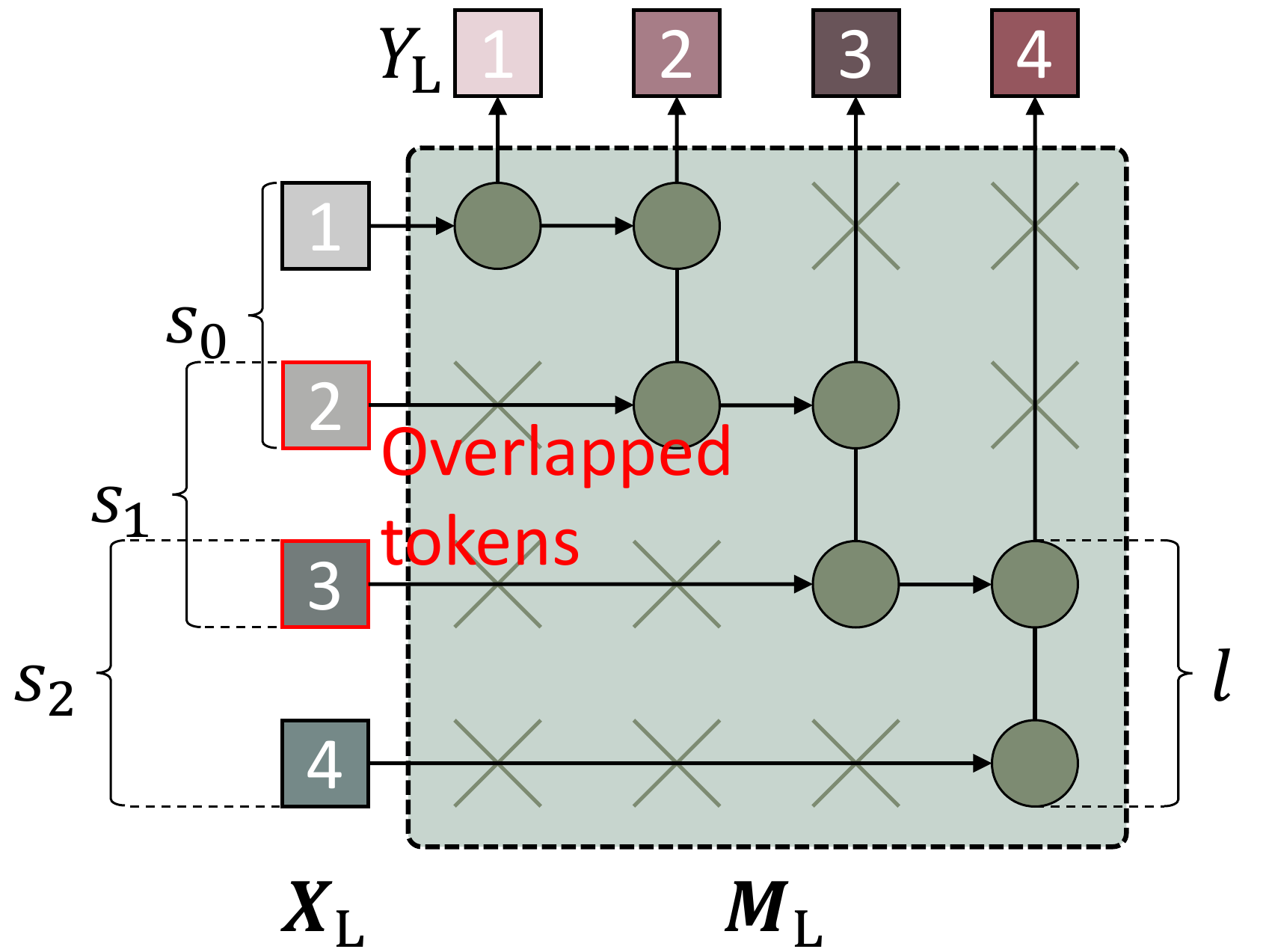}
\caption{Cross-token in another variant of local mixing that endows the cross-session connections. The input sequence length of $n=4$ are divided 3 overlapped sessions $s_0$, $s_1$ and $s_2$. Session $s_1$ extracts the information from session $s_0$ based on the overlapped 2 nd token and so on.}
\label{local mixing variant}
\end{figure}

We exploit another variant of local mixing which endows the cross-session communications\footnote{We block the cross-session interactions in TriMLP and Section \ref{cross session variant} compares the performances of these two local mixing layers.}. It treats the input historical sequence of length $n$ into $s=n-l+1$ sessions with length $l$. Accordingly, the active neurons in local mixing are reshaped as an isosceles trapezoid with the waist length of $l$. The cross-session connections are built upon the overlapped tokens, i.e., for adjacent sessions, there are $l-1$ over lapped tokens. Along with the example in Figure \ref{local mixing variant}, the input sequence of length $n=4$ is split into $s=3$ sessions $s_{\{ 0, 1, 2\}}$ of length $l=2$, and the session $s_0$, $s_1$ are connected by the overlapped 2 nd token, et cetera. This variant interacts tokens as
\begin{equation}
\label{cross session local mixing}
\emph{\textbf{Y}}_{{\rm L}_{*,i}}=
\sum^i_{j={\rm max}(1, i-l+1)}\emph{\textbf{X}}_{{\rm L}_{*,j}}\emph{\textbf{M}}_{{\rm L}_{j,i}},
\quad {\rm for}\ i = 1, \cdots n.
\end{equation}

The PyTorch-like pseudo-code of Triangular Mixer is presented in Algorithm \ref{code}. Since we convert the weights in $\emph{\textbf{M}}_{\{\rm G, L\}}$ to the probabilities over tokens by ${\rm Softmax}(\cdot)$, the dropping operation can be easily implemented by padding these neurons with ``$-\infty$''. Moreover, we find that initializing the active neurons with 1 which enforces each token contributes equally to the target during the early stage of training procedure\footnote{Section \ref{micro design} analyses the performances with different initialization.}.

\subsubsection{Discussion}
Triangular Mixer shares some similarities with the self-attention mechanism \cite{transformer}, including the global reception field and parallel processing capability. 

Notably, our mixer departs from the self-attention mechanism with the following peculiarities:
\begin{itemize}
\item\textbf{Positional Sensitive}: Since Triangular Mixer compels the cross-token interactions in strict line with the chronological order, the extra positional information (e.g. Positional Encoding \cite{transformer} or Embedding \cite{bert}) is no more necessary.

\item\textbf{Independent and Static Weights}: Triangular Mixer reserves the sequential dependency as static weights in MLP which is agnostic to the input, rather than the attention map which is dynamically generated by the scale-dot product of query and key matrices.

\item\textbf{Fewer Parameters and Higher Efficiency}: Triangular Mixer shrinks the parameter-scale by removing the query, key, value matrices mapping and Feed-Forward Network. Accordingly, our method is less computationally intensive than the self-attention mechanism.
\end{itemize}

\subsection{Classifier}
Recall that the output of Triangular Mixer is denoted as $\emph{\textbf{Y}}$, the Classifier, i.e., implemented with the plain linear layer and Softmax function, converts the $d$-dimension representation vector of each token to the probabilities over all candidate items at each time step. As follows,
\begin{equation}
\emph{\textbf{P}}={\rm Softmax}(\emph{\textbf{Y}}\cdot\emph{\textbf{W}}+\emph{\textbf{b}}),
\end{equation}
where $\emph{\textbf{W}}\in\mathbb{R}^{d \times \mathcal{|I|}}$ and $\emph{\textbf{b}}\in\mathbb{R}^{\mathcal{|I|}}$ are learnable parameters. $\emph{\textbf{P}}\in\mathbb{R}^{n \times \mathcal{|I|}}$ is the calculated probability matrix where $p_{i,c}\in[0, 1]$ is the probability over candidate item $c$ at time step $i$.

\subsection{Model Training and Recommendation}
During the training processing, we apply the standard auto-regressive fashion. Specifically, TriMLP takes the historical sequence excluded the last token $X=x_1 \rightarrow x_2 \rightarrow \cdots \rightarrow x_{n-1}$  as source, and the sequence excluded the first token $X=x_2 \rightarrow x_3 \rightarrow \cdots \rightarrow x_n$ as target. At each time step $i$, TriMLP aims at predicting the $i+1$ th token, i.e., maximizing the probability of the $i+1$ th interacted item. We use the following cross entropy loss to optimize TriMLP,
\begin{equation}
\mathcal{L}=-\sum_{X^u\in X^\mathcal{U}}\sum_{i=1}^n{\rm log}(p_{i, t_i}),
\end{equation}
where $X^\mathcal{U}$ is a training set of all users' historical sequences , $t_i$ is the target item at step $i$ and $p_{i, t_i}$ is the probability.

During the recommendation stage, TriMLP first extracts the last row $\emph{\textbf{p}}_n\in\mathbb{R}^{|\mathcal{I}|}$ from \emph{\textbf{P}} which contains the information of all interacted items in the historical sequence. Then, it ranks all candidate items according to the probabilities and retrieves $K$ items as the top-K recommendation list.

\section{Experiment and Discussion}
In this section, we start from introducing the datasets, metrics, baselines and the implement details. Then, we analyze the experimental results, including the overall recommendation performance and ablation study. Take a step further, we explore various characteristics of our Triangular Mixer. In summary,we conduct a large amount of experiments to answer the following four research questions:
\begin{itemize}
\item\textbf{RQ 1} How is the recommendation performance and inference cost of TriMLP compared to other neural network-based state-of-the-art sequential recommenders?
\item\textbf{RQ 2} How is the effectiveness of Triangular Mixer under the TriMLP architecture? Can global and local mixing both contribute to model the sequential dependency?
\item\textbf{RQ 3} How is the property of Triangular Mixer with respect to various internal structures and micro designs?
\item\textbf{RQ 4} How do the different settings of session number $s$ (or session length $l$) influence the performance of TriMLP and token-mixing manner in Triangular Mixer?
\end{itemize}

\subsection{Datasets}
\begin{table}[t]
  \centering
  \caption{Data Statistics (after pre-processed)}
  \resizebox{0.485\textwidth}{!}{
    \begin{tabular}{lccccccc}
    \toprule
    \textbf{Dataset} & \textbf{Scale} & \textbf{\# Users} & \textbf{\# Items} & \textbf{\# Interactions} & \textbf{Avg. Seq. Length} &\textbf{Max Seq. Length $n$} & \textbf{Sparsity} \\

    \midrule
    \textbf{Beauty} & \multirow{3}[2]{*}{Tiny} & 1,664  & 36,938  & 56,558  & 33.99 &32 & 99.91\% \\
    \textbf{Sports} &       & 1,958  & 55,688  & 58,844  & 30.05 &32 & 99.95\% \\
    \textbf{ML-100K} &       & 932   & 1,152  & 97,746  & 104.88 &128 & 90.90\% \\
    \midrule
    \textbf{NYC} & \multirow{3}[2]{*}{Small} & 1,031 & 5,135 & 142,237 & 137.96 &128 & 97.31\% \\
    \textbf{QB-Article} &       & 4,671  & 1,844  & 164,939  & 35.31 &32 & 98.09\% \\
    \textbf{TKY} &       & 2,267  & 7,873  & 444,183  & 195.93 &128 & 97.51\% \\
    \midrule
    \textbf{ML-1M} & \multirow{3}[2]{*}{Base} & 6,034  & 3,260  & 998,428  & 165.47 &128 & 94.92\% \\
    \textbf{QB-Video} &       & 19,047  & 15,608  & 1,370,577  & 71.96 &64 & 99.54\% \\
    \textbf{Brightkite} &       & 5,714  & 48,181  & 1,765,247  & 308.93 &256 & 99.36\% \\
    \midrule
    \textbf{Yelp} & \multirow{3}[2]{*}{Large} & 42,461  & 101,269  & 2,199,786  & 51.81 &64 & 99.95\% \\
    \textbf{Gowalla} &       & 32,439  & 131,329  & 2,990,783  & 92.20 &64 & 99.93\% \\
    \textbf{ML-10M} &       & 69,865  & 9,708  & 9,995,230  & 143.06 &128 & 98.53\% \\
    \bottomrule
    \end{tabular}}%
  \label{data statistics}%
\end{table}%

We evaluate our method on 12 publicly available datasets from 4 benchmarks. Specifically, we select 
\texttt{Beauty}, 
\texttt{Sports} from Amazon\footnote{\url{http://snap.stanford.edu/data/web-Amazon-links.html}} \cite{amazon}, 
\texttt{ML-100K}, 
\texttt{ML-1M}, 
\texttt{ML-10M} from MovieLens\footnote{\url{https://grouplens.org/datasets/movielens/}} \cite{mvln}, 
\texttt{QB-Article}, 
\texttt{QB-Video} from Tenrec\footnote{\url{https://static.qblv.qq.com/qblv/h5/algo-frontend/tenrec_dataset.html}} \cite{tenrec}, and 
\texttt{NYC}\footnote{\url{https://sites.google.com/site/yangdingqi/home/foursquare-dataset}\label{nyc}}, 
\texttt{TKY}\textsuperscript{\ref{nyc}}, 
\texttt{Brightkite}\footnote{\url{https://snap.stanford.edu/data/loc-brightkite.html}}, 
\texttt{Yelp}\footnote{\url{https://www.yelp.com/dataset}}, 
\texttt{Gowalla}\footnote{\url{https://snap.stanford.edu/data/loc-Gowalla.html}}
from the scenario of Location-based Social Network (LBSN). In accordance with \cite{stisan}, we remove the ``inactive'' users who interact with fewer than 20 items and the ``unpopular'' items which are interacted by less than 10 times. According to the number of interactions, we categorize the 12 datasets into 4 different scales: Tiny, Small, Base and Large which separately contain 50K\textasciitilde100K, 150K\textasciitilde500K, 1M\textasciitilde2M and 2M\textasciitilde10M interactions. Table \ref{data statistics} summarizes the statistics.

We set the maximum sequence length $n$ of each dataset according to the average sequence length. For each user, we take the last previously un-interacted item as the target and utilize all prior items for training during the data partition. 

\subsection{Metric}
We introduce the following 3 metrics to measure the efficiency and accuracy of sequential recommenders.
\begin{itemize}
\item\textbf{Inference Time (Infer. Time)} calculates the average time cost of finishing 100 rounds recommendation.

\item\textbf{Hit Rate (HR@K)} \cite{metric} counts the fraction of times that target item is among the top-K recommendation list.

\item\textbf{Normalized Discounted Cumulative Gain (NDCG@K)} \cite{metric} rewards the method that ranks the positive items in first few positions of the top-K recommendation list.
\end{itemize}

The smaller Infer. Time stands for the better efficiency, and the recommendation performance is positively correlated with the values of HR and NDCG. We report $K=\{5, 10\}$ in our experiments. To avoid the bias brought by different negative sampling strategies \cite{recall}, we compare the probability of the target item with all other items in the dataset, and compute the HR, NDCG based on the ranking of all items.

\subsection{Baselines}
Since TriMLP concentrates on improving the primary ability of MLP in encoding cross-token interactions, we compare it with the following four representative sequential recommenders developed from different neural networks:
\begin{itemize}
\item \textbf{GRU4Rec} \cite{gru4rec} utilizes RNN to model historical sequences and dynamic preferences for sequential recommendation.
\item \textbf{NextItNet} \cite{nextitnet} is a state-of-the-art CNN-based generative model for recommendation, which learns high-level representation from both short and long-range dependencies.

\item \textbf{SASRec} \cite{sasrec} employs Transformer-based encoder for recommendation where the self-attention mechanism dynamically models the sequential dependency.

\item \textbf{FMLP4Rec} \cite{f-mlp4rec} is a state-of-the-art all-MLP sequential recommender, which follows the denoising manner and establishes cross-token contact by Fourier Transform.
\end{itemize}

\begin{table*}[t]
  \centering
  \caption{Overall Recommendation Performance. The arrow ``$\uparrow$'' (or ``$\downarrow$'') denotes that the higher (or lower) value, the better metric. We use boldface and underline to indicate the best and second results in each column, respectively. The ``Comparison'' row reports the relative improvement or decline of TriMLP against the strongest baseline.}
  \resizebox{\textwidth}{!}{
    \begin{tabular}{lc|ccccc|ccccc|ccccc}
    \toprule
    \multicolumn{2}{l|}{\textbf{Dataset-Tiny}} & \multicolumn{5}{c|}{\textbf{Beauty}}  & \multicolumn{5}{c|}{\textbf{Sports}}  & \multicolumn{5}{c}{\textbf{ML-100K}} \\
    \midrule
    \textbf{Model} & \textbf{Mixer} 
    & \textbf{Infer. Time (s) $\downarrow$} & \textbf{HR@5 $\uparrow$} & \textbf{NDCG@5 $\uparrow$} & \textbf{HR@10 $\uparrow$} & \textbf{NDCG@10 $\uparrow$} 
    & \textbf{Infer. Time (s) $\downarrow$} & \textbf{HR@5 $\uparrow$} & \textbf{NDCG@5 $\uparrow$} & \textbf{HR@10 $\uparrow$} & \textbf{NDCG@10 $\uparrow$} 
    & \textbf{Infer. Time (s) $\downarrow$} & \textbf{HR@5 $\uparrow$} & \textbf{NDCG@5 $\uparrow$} & \textbf{HR@10 $\uparrow$} & \textbf{NDCG@10 $\uparrow$} \\
    \midrule
    \textbf{GRU4Rec} & \textbf{RNN} 
    & 0.3176 & 0.07752 & 0.05130 & 0.11839 & 0.06440 
    & 0.3361 & 0.01124 & 0.00665 & 0.01839 & 0.00900 
    & \underline{0.3778} & 0.05472 & 0.03557 & \underline{0.12124} & 0.05682 \\
    
    \textbf{NextItNet} & \textbf{CNN} 
    & \underline{0.3196} & \underline{0.08534} & \underline{0.06376} & 0.10998 & \underline{0.07164} 
    & 0.3339 & \underline{0.01532} & \underline{0.01177} & 0.02043 & 0.01334 
    & 0.3861 & \underline{0.06974} & \underline{0.04463} & 0.12017 & \underline{0.06083} \\
    
    \textbf{SASRec} & \textbf{Trans.} 
    & 0.3185 & 0.08113 & 0.06106 & 0.11358 & 0.07160 
    & \underline{0.3293} & 0.00409 & 0.00191 & 0.00817 & 0.00322 
    & 0.3953 & 0.02682 & 0.01444 & 0.05365 & 0.02300 \\
    
    \textbf{FMLP4Rec} & \textbf{Four.} 
    & 0.3246 & 0.08353 & 0.05962 & \underline{0.12019} & 0.07121 
    & 0.3323 & 0.01481 & 0.01167 & \underline{0.02298} & \underline{0.01420} 
    & 0.3948 & 0.06760 & 0.04144 & 0.11373 & 0.05646 \\
    
    \textbf{TriMLP} & \textbf{MLP} 
    & \textbf{0.3103} & \textbf{0.09615} & \textbf{0.07070} & \textbf{0.12560} & \textbf{0.08003} 
    & \textbf{0.3176} & \textbf{0.01839} & \textbf{0.01258} & \textbf{0.02451} & \textbf{0.01442} 
    & \textbf{0.3763} & \textbf{0.08691} & \textbf{0.05848} & \textbf{0.15451} & \textbf{0.07988} \\
    \midrule
    \multicolumn{2}{l|}{\textbf{Comparison}} 
    & \textcolor{NavyBlue}{-2.30\%} & \textcolor{purple}{+11.24\%} & \textcolor{purple}{+9.82\%} & \textcolor{purple}{+4.50\%} & \textcolor{purple}{+11.71\%}
    & \textcolor{NavyBlue}{-3.55\%} & \textcolor{purple}{+20.04\%} & \textcolor{purple}{+6.88\%} & \textcolor{purple}{+6.66\%} & \textcolor{purple}{+1.55\%} 
    & \textcolor{NavyBlue}{-0.40\%} & \textcolor{purple}{+24.62\%} & \textcolor{purple}{+31.03\%} & \textcolor{purple}{+27.44\%} & \textcolor{purple}{+31.32\%} \\
    \midrule
    \midrule
    \multicolumn{2}{l|}{\textbf{Dataset-Small}} & \multicolumn{5}{c|}{\textbf{NYC}}     & \multicolumn{5}{c|}{\textbf{QB-Article}} & \multicolumn{5}{c}{\textbf{TKY}} \\
    \midrule
    \textbf{Model} & \textbf{Mixer} & \textbf{Infer. Time (s) $\downarrow$} & \textbf{HR@5 $\uparrow$} & \textbf{NDCG@5 $\uparrow$} & \textbf{HR@10 $\uparrow$} & \textbf{NDCG@10 $\uparrow$} & \textbf{Infer. Time (s) $\downarrow$} & \textbf{HR@5 $\uparrow$} & \textbf{NDCG@5 $\uparrow$} & \textbf{HR@10 $\uparrow$} & \textbf{NDCG@10 $\uparrow$} & \textbf{Infer. Time (s) $\downarrow$} & \textbf{HR@5 $\uparrow$} & \textbf{NDCG@5 $\uparrow$} & \textbf{HR@10 $\uparrow$} & \textbf{NDCG@10 $\uparrow$} \\
    \midrule
    \textbf{GRU4Rec} & \textbf{RNN} 
    & \underline{0.3364} & 0.02037 & 0.01268 & 0.03589 & 0.01783 
    & \underline{0.3315} & \underline{0.06487} & 0.03811 & \underline{0.13466} & \underline{0.06052} 
    & \underline{0.3540} & 0.02603 & 0.01715 & 0.04279 & 0.234 \\
    \textbf{NextItNet} & \textbf{CNN} 
    & 0.3482 & 0.03686 & 0.02272 & 0.05141 & 0.02748 
    & 0.3389 & 0.06101 & 0.03565 & 0.11732 & 0.05373 
    & 0.3716 & 0.03264 & 0.01836 & 0.06308 & 0.02802 \\
    \textbf{SASRec} & \textbf{Trans.} 
    & 0.3568 & \underline{0.04074} & \underline{0.02370} & \underline{0.05723} & \underline{0.02889} 
    & 0.3461 & 0.05973 & \underline{0.03820} & 0.10662 & 0.05324 
    & 0.3909 & 0.02955 & 0.01821 & 0.05073 & 0.02498 \\
    \textbf{FMLP4Rec} & \textbf{Four.} 
    & 0.3551 & 0.03395 & 0.02077 & 0.05432 & 0.02722 
    & 0.3465 & 0.05331 & 0.03158 & 0.10026 & 0.47290 
    & 0.3854 & \underline{0.04499} & \underline{0.02566} & \underline{0.08293} & \underline{0.03794} \\
    
    \textbf{TriMLP} & \textbf{MLP} 
    & \textbf{0.3350} & \textbf{0.04365} & \textbf{0.02574} & \textbf{0.06111} & \textbf{0.03121} 
    & \textbf{0.3225} & \textbf{0.07621} & \textbf{0.04751} & \textbf{0.13552} & \textbf{0.06639} 
    & \textbf{0.3350} & \textbf{0.05293} & \textbf{0.03175} & \textbf{0.09043} & \textbf{0.04384} \\
    \midrule
    \multicolumn{2}{l|}{\textbf{Comparison}} 
    & \textcolor{NavyBlue}{-0.42\%} & \textcolor{purple}{+7.14\%} & \textcolor{purple}{+8.61\%} & \textcolor{purple}{+6.78\%} & \textcolor{purple}{+8.03\% }
    & \textcolor{NavyBlue}{-2.79\%} & \textcolor{purple}{+17.48\%} & \textcolor{purple}{+24.37\%} & \textcolor{purple}{+0.64\%} & \textcolor{purple}{+9.70\% }
    & \textcolor{NavyBlue}{-5.76\%} & \textcolor{purple}{+17.65\%} & \textcolor{purple}{+23.73\%} & \textcolor{purple}{+9.04\%} & \textcolor{purple}{+15.55\%} \\
    \midrule
    \midrule
    \multicolumn{2}{l|}{\textbf{Dataset-Base}} & \multicolumn{5}{c|}{\textbf{ML-1M}}   & \multicolumn{5}{c|}{\textbf{QB-Video}} & \multicolumn{5}{c}{\textbf{Brightkite}} \\
    \midrule
    \textbf{Model} & \textbf{Mixer} & \textbf{Infer. Time (s) $\downarrow$} & \textbf{HR@5 $\uparrow$} & \textbf{NDCG@5 $\uparrow$} & \textbf{HR@10 $\uparrow$} & \textbf{NDCG@10 $\uparrow$} & \textbf{Infer. Time (s) $\downarrow$} & \textbf{HR@5 $\uparrow$} & \textbf{NDCG@5 $\uparrow$} & \textbf{HR@10 $\uparrow$} & \textbf{NDCG@10 $\uparrow$} & \textbf{Infer. Time (s) $\downarrow$} & \textbf{HR@5 $\uparrow$} & \textbf{NDCG@5 $\uparrow$} & \textbf{HR@10 $\uparrow$} & \textbf{NDCG@10 $\uparrow$} \\
    \midrule
    \textbf{GRU4Rec} & \textbf{RNN} 
    & \underline{0.4204} & 0.14700 & 0.09583 & 0.20655 & 0.10151 
    & \underline{0.4656} & 0.03560 & 0.02135 & 0.06762 & 0.03159 
    & \underline{0.5385} & 0.01540 & 0.00948 & 0.02678 & 0.01314 \\
    \textbf{NextItNet} & \textbf{CNN} 
    & 0.4600 & \underline{0.15291} & \underline{0.09969} & \underline{0.21113} & \underline{0.12480} 
    & 0.5227 & \underline{0.03602} & \underline{0.02149} & \underline{0.06872} & \underline{0.03200} 
    & 0.6272 & 0.02468 & 0.01232 & 0.04705 & 0.02266 \\
    \textbf{SASRec} & \textbf{Trans.} 
    & 0.5279 & 0.15214 & 0.09286 & 0.20968 & 0.10877 
    & 0.6173 & 0.03418 & 0.02051 & 0.06321 & 0.02977 
    & 0.8505 & \underline{0.03220} & \underline{0.02054} & \underline{0.04953} & \underline{0.02615} \\
    \textbf{FMLP4Rec} & \textbf{Four.} 
    & 0.5110 & 0.08336 & 0.05189 & 0.12728 & 0.06602 
    & 0.6287 & 0.02426 & 0.01332 & 0.04415 & 0.01969 
    & 0.7262 & 0.02520 & 0.01330 & 0.04428 & 0.01951 \\
    \textbf{TriMLP} & \textbf{MLP} 
    & \textbf{0.3924} & \textbf{0.16390} & \textbf{0.11196} & \textbf{0.23434} & \textbf{0.13454} 
    & \textbf{0.4176} & \textbf{0.04284} & \textbf{0.02550} & \textbf{0.07445} & \textbf{0.03563} 
    & \textbf{0.4878} & \textbf{0.04848} & \textbf{0.03016} & \textbf{0.05863} & \textbf{0.03335} \\
    \midrule
    \multicolumn{2}{l|}{\textbf{Comparison}} 
    & \textcolor{NavyBlue}{-6.66\%} & \textcolor{purple}{+7.19\%} & \textcolor{purple}{+12.31\%} & \textcolor{purple}{+10.99\%} & \textcolor{purple}{+7.80\%} 
    & \textcolor{NavyBlue}{-10.31\%} & \textcolor{purple}{+18.93\%} & \textcolor{purple}{+18.66\%} & \textcolor{purple}{+8.34\%} & \textcolor{purple}{+11.34\%} 
    & \textcolor{NavyBlue}{-9.42\%} & \textcolor{purple}{+50.56\%} & \textcolor{purple}{+46.84\%} & \textcolor{purple}{+18.37\%} & \textcolor{purple}{+27.53\%} \\
    \midrule
    \midrule
    \multicolumn{2}{l|}{\textbf{Dataset-Large}} & \multicolumn{5}{c|}{\textbf{Yelp}}    & \multicolumn{5}{c|}{\textbf{Gowalla}} & \multicolumn{5}{c}{\textbf{ML-10M}} \\
    \midrule
    \textbf{Model} & \textbf{Mixer} & \textbf{Infer. Time (s) $\downarrow$} & \textbf{HR@5 $\uparrow$} & \textbf{NDCG@5 $\uparrow$} & \textbf{HR@10 $\uparrow$} & \textbf{NDCG@10 $\uparrow$} & \textbf{Infer. Time (s) $\downarrow$} & \textbf{HR@5 $\uparrow$} & \textbf{NDCG@5 $\uparrow$} & \textbf{HR@10 $\uparrow$} & \textbf{NDCG@10 $\uparrow$} & \textbf{Infer. Time (s) $\downarrow$} & \textbf{HR@5 $\uparrow$} & \textbf{NDCG@5 $\uparrow$} & \textbf{HR@10 $\uparrow$} & \textbf{NDCG@10 $\uparrow$} \\
    \midrule
    \textbf{GRU4Rec} & \textbf{RNN} 
    & \underline{0.6634} & 0.01347 & 0.00820 & 0.02508 & 0.01190 
    & \underline{0.7060} & 0.02022 & 0.01075 & 0.05012 & 0.02025 
    & \underline{1.1730} & 0.10844 & 0.07214 & 0.17025 & 0.09523 \\
    \textbf{NextItNet} & \textbf{CNN} 
    & 0.8083 & \underline{0.01503} & \underline{0.00901} & \underline{0.02737} & \underline{0.01295} 
    & 0.8100 & \underline{0.05012} & \underline{0.03022} & \underline{0.08752} & \underline{0.04216} 
    & 1.6706 & \underline{0.12769} & \underline{0.08992} & \underline{0.19066} & \underline{0.10748} \\
    \textbf{SASRec} & \textbf{Trans.} 
    & 0.9810 & 0.01006 & 0.00602 & 0.01948 & 0.00904 
    & 0.9352 & 0.04254 & 0.02595 & 0.07904 & 0.03761 
    & 2.5037 & 0.11720 & 0.08830 & 0.17869 & 0.09905 \\
    \textbf{FMLP4Rec} & \textbf{Four.} 
    & 1.0085 & 0.00794 & 0.00411 & 0.01557 & 0.00657 
    & 0.9492 & 0.02799 & 0.0144 & 0.04834 & 0.02095 
    & 2.2933 & 0.06319 & 0.03677 & 0.09812 & 0.04799 \\
    \textbf{TriMLP} & \textbf{MLP} 
    & \textbf{0.5241} & \textbf{0.01554} & \textbf{0.00963} & \textbf{0.02784} & \textbf{0.01357} 
    & \textbf{0.5815} & \textbf{0.06175} & \textbf{0.03968} & \textbf{0.09997} & \textbf{0.05194} 
    & \textbf{0.8946} & \textbf{0.13900} & \textbf{0.09636} & \textbf{0.20273} & \textbf{0.11685} \\
    \midrule
    \multicolumn{2}{l|}{\textbf{Comparison}} 
    & \textcolor{NavyBlue}{-20.80\%} & \textcolor{purple}{+3.39\%} & \textcolor{purple}{+6.88\%} & \textcolor{purple}{+1.72\%} & \textcolor{purple}{+4.79\%} 
    & \textcolor{NavyBlue}{-17.63\%} & \textcolor{purple}{+23.20\%} & \textcolor{purple}{+31.30\%} & \textcolor{purple}{+14.23\%} & \textcolor{purple}{+23.20\%} 
    & \textcolor{NavyBlue}{-23.73\%} & \textcolor{purple}{+8.86\%} & \textcolor{purple}{+7.16\%} & \textcolor{purple}{+6.33\%} & \textcolor{purple}{+8.72\%} \\
    \bottomrule
    \end{tabular}}%
  \label{overall recommendation performance}%
  \vspace{-10pt}
\end{table*}%

\subsection{Implementation Details}
For the rigorous comparison, we uniform the width and depth of TriMLP and baselines, and ensure that all the compared method differentiate only in the neural network architecture. Specifically, we set the embedding dimension $d$ to 128, and the intermediate dimension in Triangular Mixer to $n$. Since TriMLP contains 2 MLP layers for cross-token interactions, we stack 2 token-mixing encoders in each baseline.

During the training stage, we perform the standard auto-regressive manner, and adopt the identical gradient-updating strategy for all compared methods where dropout rate is 0.5 and the optimizer is Adam \cite{adam} with the learning rate of 0.001. The parameters keep updating until the performance no longer increases for consecutive 10 epochs. All experiments are conducted on a single server with 64GB RAM, AMD Ryzen 5900X CPU and NVIDIA RTX 3090 GPU.

\subsection{Overall Recommendation Performance (RQ 1)}
The experimental results of all compared methods on 12 datasets are summarized in Table \ref{overall recommendation performance}. From the table, we have the following observations.

\textit{Observation 1: Consistent superior recommendation performance.} NextItNet with temporal CNN architecture is the strongest baseline with decent scores on most datasets, while the performances of GRU4Rec, SASRec and FMLP4Rec drifts sharply. Notably, TriMLP achieves the state-of-the-art performances on all validated 12 datasets. Specifically, TriMLP is substantial ahead of the strongest baseline averagely by 15.57\%, 18.23\%, 18.35\% and 11.66\% cross 4 different scales of datasets in terms of HR and NDCG. It demonstrates that our method equips MLP with the ample sequential modeling ability under the same training manner, which is competitive to RNN, CNN, Transformer and Fourier transform.

\textit{Observation 2: Incremental ascendancy in efficiency.} According to the metric of Infer. Time, TriMLP saves the inference cost by 2.08\%, 2.99\%, 8.80\% and 20.72\% respectively on Tiny, Small, Base and Large datasets compared to the fastest competitor. The reduction increases with the scale of datasets which is in line with our expectation. Since the number of interactions are minor in Tiny and Small datasets, the computation is more intensive on the Embedding part and Classifier. Along with the number of interactions increases, i.e., mixing token occupies the main part of the computation, the efficiency advantages of TriMLP show up. It is credited to the plain structure in Triangular Mixer, which only contains 2 matrix transposition and multiplications. We provide a more detailed case study on the largest \texttt{ML-10M} in Section  \ref{case study} to reveal the advantage of TriMLP in computational complexity.

\textit{Observation 3: Surprisingly good accuracy/efficiency trade-off.} Throughout all 12 validated datasets, TriMLP can averagely provide 14.88\% higher recommendation performance against SOTA and reduce 8.64\% inference time. The proposed TriMLP architecture reveals the promising potential to be served as an alternative for sequential recommenders.

\begin{table}[t]
  \centering
  \caption{Computational Complexity Comparison on ML-10M.}
  \resizebox{0.485\textwidth}{!}{
    \begin{tabular}{l|cccccc}
    \toprule
    \textbf{Dataset-Large} & \multicolumn{5}{c}{\textbf{ML-10M}}   &  \\
    \midrule
    \textbf{Settings} & \multicolumn{5}{c}{batch size: 512, sequence length $n=128$, dimension $d=128$ and kernel size $k=3$}   &  \\
    \midrule
    \textbf{Model} 
    & \textbf{Complexity} & \textbf{MACs$\downarrow$} & \textbf{Par. Scale$\downarrow$} & \textbf{GPU Mem.$\downarrow$} 
    & \textbf{Infer. Time$\downarrow$} & \textbf{HR@5 $\uparrow$} \\
    \midrule
    \textbf{GRU4Rec} & $O(nd^2)$ & \underline{13.04} G & \underline{0.19} M & 2,536 MB & \underline{1.1730} s & 0.10844 \\
    \textbf{NextItNet} & $O(k\cdot nd^2)$ & 26.03 G & 0.40 M & \underline{1,810} MB & 1.6706 s & \underline{0.12769} \\
    \textbf{SASRec} & $O(n^2d)$ & 25.94 G & 0.40 M & 1,819 MB & 2.5037 s & 0.11720 \\
    \textbf{FMLP4Rec} & $O(nd\cdot{\rm log}(nd))$ & 17.18 G & 0.26 M & 2,093 MB & 2.2933 s & 0.06319 \\
    \textbf{TriMLP} & $O(n^2d)$ & \textbf{2.15 G} & \textbf{0.03 M} & \textbf{1,391} MB & \textbf{0.8946} s & \textbf{0.13900} \\
    \midrule
    \textbf{Comparison}  & - & \textcolor{NavyBlue}{-83.51\%} & \textcolor{NavyBlue}{-84.21\%} & \textcolor{NavyBlue}{-23.15\%} & \textcolor{NavyBlue}{-23.73\%} & \textcolor{purple}{+8.86\%} \\
    \bottomrule
    \end{tabular}}%
  \label{Computational Complexity Comparison on ML-10M}%
  \vspace{-10pt}
\end{table}%

\subsection{Computational Complexity Analysis on ML-10M}\label{case study}
We count the Multiply–Accumulate Operations (MACs), Parameter Scale (Para. Scale) and GPU Memory Occupation (GPU Mem.) on the largest dataset \texttt{ML-10M}. Since all the compared methods share the common Embedding layer and Classifier, we only calculate MACs and Par. Scale of the encoder in each model, i.e., the RNN (or CNN) layer in GRU4Rec (or NextItNet) and the self-attention (or denoised Fourier) layer in SASRec (or FMLP4Rec).

The experimental results are summarized in Table \ref{Computational Complexity Comparison on ML-10M}. Albeit the computational complexity is quadratic correlated to the input sequence length $n$, TriMLP possesses the higher efficiency due to parallel and minimalist matrix multiplication. Notably, TriMLP shrinks 83.51\% MACs, 84.21\% Para. Scale, 23.15\% GPU Mem. and 23.73\% Infer. Time.

\begin{table*}[htbp]
  \centering
  \caption{Ablation Study. The best performance is boldfaced. The metric of ``Avg. Impv.'' stands for the average recommendation performance improvement of the variant against the baseline EyeMLP.}
    \resizebox{\textwidth}{!}{
    \begin{tabular}{l|ccccc|ccccc|ccccc}
    \toprule
    \textbf{Dataset-Tiny} & \multicolumn{5}{c|}{\textbf{Beauty}}  & \multicolumn{5}{c|}{\textbf{Sports}}  & \multicolumn{5}{c}{\textbf{ML-100K}} \\
    \midrule
    \textbf{Model} & \textbf{HR@5$\uparrow$} & \textbf{NDCG@5$\uparrow$} & \textbf{HR@10$\uparrow$} & \textbf{NDCG@10$\uparrow$} & \textbf{Avg. Impv.$\uparrow$} & \textbf{HR@5$\uparrow$} & \textbf{NDCG@5$\uparrow$} & \textbf{HR@10$\uparrow$} & \textbf{NDCG@10$\uparrow$} & \textbf{Avg. Impv.$\uparrow$} & \textbf{HR@5$\uparrow$} & \textbf{NDCG@5$\uparrow$} & \textbf{HR@10$\uparrow$} & \textbf{NDCG@10$\uparrow$} & \textbf{Avg. Impv.$\uparrow$} \\
    \midrule
    \textbf{EyeMLP} & 0.08053 & 0.05942 & 0.09976 & 0.06561 & -     & 0.01685 & 0.01118 & 0.02043 & 0.01229 & -     & 0.05579 & 0.03624 & 0.09227 & 0.04768 & - \\
    \textbf{SqrMLP} 
    & 0.00421 & 0.00362 & 0.00781 & 0.00473 & \textcolor{NavyBlue}{-93.41}\% 
    & 0.00051 & 0.00022 & 0.00153 & 0.00053 & \textcolor{NavyBlue}{-95.80}\% 
    & 0.01180 & 0.00759 & 0.02253 & 0.01100 & \textcolor{NavyBlue}{-77.60}\% \\
    
    \textbf{TriMLP\_G} 
    & 0.08233 & 0.06177 & 0.11599 & 0.07261 & \textcolor{purple}{+8.28}\% 
    & 0.01583 & 0.01063 & 0.02298 & 0.01287 & \textcolor{purple}{+1.56}\% 
    & 0.06545 & 0.03819 & 0.11803 & 0.05517 & \textcolor{purple}{+16.58}\% \\
    
    \textbf{TriMLP\_L} 
    & 0.08173 & 0.06218 & 0.09916 & 0.06754 & \textcolor{purple}{+2.12}\% 
    & 0.01733 & 0.01176 & 0.02331 & 0.01399 & \textcolor{purple}{+9.73}\% 
    & 0.06009 & 0.03973 & 0.10622 & 0.05451 & \textcolor{purple}{+11.70}\% \\
    
    \textbf{TriMLP} 
    & \textbf{0.09615} & \textbf{0.07070} & \textbf{0.12560} & \textbf{0.08003} & \textbf{\textcolor{purple}{+21.57\%}} 
    & \textbf{0.01839} & \textbf{0.01258} & \textbf{0.02451} & \textbf{0.01442} & \textbf{\textcolor{purple}{+14.74\%}} 
    & \textbf{0.08691} & \textbf{0.05848} & \textbf{0.15451} & \textbf{0.07988} & \textbf{\textcolor{purple}{+63.03\%}} \\
    \midrule
    \midrule
    \textbf{Dataset-Small} & \multicolumn{5}{c|}{\textbf{NYC}}     & \multicolumn{5}{c|}{\textbf{QB-Article}} & \multicolumn{5}{c}{\textbf{TKY}} \\
    \midrule
    \textbf{Model} & \textbf{HR@5$\uparrow$} & \textbf{NDCG@5$\uparrow$} & \textbf{HR@10$\uparrow$} & \textbf{NDCG@10$\uparrow$} & \textbf{Avg. Impv.$\uparrow$} & \textbf{HR@5$\uparrow$} & \textbf{NDCG@5$\uparrow$} & \textbf{HR@10$\uparrow$} & \textbf{NDCG@10$\uparrow$} & \textbf{Avg. Impv.$\uparrow$} & \textbf{HR@5$\uparrow$} & \textbf{NDCG@5$\uparrow$} & \textbf{HR@10$\uparrow$} & \textbf{NDCG@10$\uparrow$} & \textbf{Avg. Impv.$\uparrow$} \\
    \midrule
    \textbf{EyeMLP} 
    & 0.01940 & 0.01297 & 0.02813 & 0.01570 & -     
    & 0.04389 & 0.02650 & 0.08885 & 0.04074 & -     
    & 0.04801 & 0.02715 & 0.08255 & 0.03318 & - \\
    
    \textbf{SqrMLP} 
    & 0.00291 & 0.00128 & 0.00582 & 0.00221 & \textcolor{NavyBlue}{-85.09\%} 
    & 0.01841 & 0.00885 & 0.05523 & 0.02058 & \textcolor{NavyBlue}{-53.00\%} 
    & 0.00706 & 0.00431 & 0.01279 & 0.00612 & \textcolor{NavyBlue}{-83.87\%} \\
    \textbf{TriMLP\_G} 
    & 0.02813 & 0.02030 & \textbf{0.06111} & 0.03065 & \textcolor{purple}{+78.49}\% 
    & 0.07193 & 0.04421 & 0.13273 & 0.06356 & \textcolor{purple}{+59.03}\% 
    & 0.04955 & 0.02722 & 0.08640 & 0.03364 & \textcolor{purple}{+2.38\%} \\
    
    \textbf{TriMLP\_L} 
    & 0.04171 & 0.02337 & 0.05626 & 0.02797 & \textcolor{purple}{+93.33\%} 
    & 0.07129 & 0.04363 & 0.13702 & 0.06456 & \textcolor{purple}{+59.94\%} 
    & 0.05026 & 0.02924 & 0.08907 & 0.04163 & \textcolor{purple}{+11.48\%} \\
    
    \textbf{TriMLP} 
    & \textbf{0.04365} & \textbf{0.02574} & \textbf{0.06111} & \textbf{0.03121} & \textbf{\textcolor{purple}{+109.87\%}} 
    & \textbf{0.07621} & \textbf{0.04751} & \textbf{0.13552} & \textbf{0.06639} & \textbf{\textcolor{purple}{+64.10\%}} 
    & \textbf{0.05293} & \textbf{0.03175} & \textbf{0.09043} & \textbf{0.04384} & \textbf{\textcolor{purple}{+17.22\%}} \\
    \midrule
    \midrule
    
    \textbf{Dataset-Base} & \multicolumn{5}{c|}{\textbf{ML-1M}}   & \multicolumn{5}{c|}{\textbf{QB-Video}} & \multicolumn{5}{c}{\textbf{Brightkite}} \\
    \midrule
    \textbf{Model} & \textbf{HR@5$\uparrow$} & \textbf{NDCG@5$\uparrow$} & \textbf{HR@10$\uparrow$} & \textbf{NDCG@10$\uparrow$} & \textbf{Avg. Impv.$\uparrow$} & \textbf{HR@5$\uparrow$} & \textbf{NDCG@5$\uparrow$} & \textbf{HR@10$\uparrow$} & \textbf{NDCG@10$\uparrow$} & \textbf{Avg. Impv.$\uparrow$} & \textbf{HR@5$\uparrow$} & \textbf{NDCG@5$\uparrow$} & \textbf{HR@10$\uparrow$} & \textbf{NDCG@10$\uparrow$} & \textbf{Avg. Impv.$\uparrow$} \\
    \midrule
    \textbf{EyeMLP} & 0.12562 & 0.08582 & 0.18097 & 0.10360 & -     & 0.02426 & 0.01491 & 0.04274 & 0.02079 & -     & 0.04025 & 0.02106 & 0.06493 & 0.02901 & - \\
    \textbf{SqrMLP} 
    & 0.00530 & 0.00316 & 0.01210 & 0.00529 & \textcolor{NavyBlue}{-95.08\%} 
    & 0.00546 & 0.00274 & 0.01402 & 0.00545 & \textcolor{NavyBlue}{-75.02\%} 
    & 0.00665 & 0.00360 & 0.01120 & 0.00508 & \textcolor{NavyBlue}{-82.91\%} \\
    
    \textbf{TriMLP\_G} 
    & 0.14418 & 0.09651 & 0.21064 & 0.11792 & \textcolor{purple}{+14.36\%} 
    & 0.04127 & 0.02479 & 0.07397 & 0.03527 & \textcolor{purple}{+69.77\%} 
    & 0.04498 & 0.02688 & 0.05828 & 0.03118 & \textcolor{purple}{+9.16\%} \\
    
    \textbf{TriMLP\_L} 
    & 0.13656 & 0.09207 & 0.20865 & 0.11520 & \textcolor{purple}{+10.62\%} 
    & 0.03670 & 0.02255 & 0.06673 & 0.03215 & \textcolor{purple}{+53.32\%} 
    & 0.04760 & 0.02872 & 0.05775 & 0.03194 & \textcolor{purple}{+13.42\%} \\
    
    \textbf{TriMLP} 
    & \textbf{0.16390} & \textbf{0.11196} & \textbf{0.23434} & \textbf{0.13454} & \textbf{\textcolor{purple}{+30.07\%}} 
    & \textbf{0.04284} & \textbf{0.02550} & \textbf{0.07445} & \textbf{0.03563} & \textbf{\textcolor{purple}{+73.30\%}} 
    & \textbf{0.04848} & \textbf{0.03016} & \textbf{0.05863} & \textbf{0.03335} & \textbf{\textcolor{purple}{+17.23\%}} \\
    \midrule
    \midrule
    \textbf{Dataset-Large} & \multicolumn{5}{c|}{\textbf{Yelp}}    & \multicolumn{5}{c|}{\textbf{Gowalla}} & \multicolumn{5}{c}{\textbf{ML-10M}} \\
    \midrule
    \textbf{Model} & \textbf{HR@5$\uparrow$} & \textbf{NDCG@5$\uparrow$} & \textbf{HR@10$\uparrow$} & \textbf{NDCG@10$\uparrow$} & \textbf{Avg. Impv.$\uparrow$} & \textbf{HR@5$\uparrow$} & \textbf{NDCG@5$\uparrow$} & \textbf{HR@10$\uparrow$} & \textbf{NDCG@10$\uparrow$} & \textbf{Avg. Impv.$\uparrow$} & \textbf{HR@5$\uparrow$} & \textbf{NDCG@5$\uparrow$} & \textbf{HR@10$\uparrow$} & \textbf{NDCG@10$\uparrow$} & \textbf{Avg. Impv.$\uparrow$} \\
    \midrule
    \textbf{EyeMLP} & 0.01116 & 0.00687 & 0.01934 & 0.00948 & -     & 0.04749 & 0.02822 & 0.08632 & 0.04303 & -     & 0.10652 & 0.07313 & 0.15292 & 0.08804 & - \\
    \textbf{SqrMLP} 
    & 0.00210 & 0.00112 & 0.00469 & 0.00194 & \textcolor{NavyBlue}{-80.04\%} 
    & 0.00487 & 0.00240 & 0.00909 & 0.00376 & \textcolor{NavyBlue}{-90.49\%} 
    & 0.08097 & 0.05532 & 0.11866 & 0.06746 & \textcolor{NavyBlue}{-23.53\%} \\
    
    \textbf{TriMLP\_G} 
    & 0.01491 & 0.00903 & \textbf{0.02812} & 0.01325 & \textcolor{purple}{+37.55\%} 
    & 0.05629 & 0.03489 & 0.09726 & 0.04801 & \textcolor{purple}{+16.60\%} 
    & 0.12317 & 0.08480 & 0.18006 & 0.10312 & \textcolor{purple}{+16.62\%} \\
    
    \textbf{TriMLP\_L} 
    & 0.01241 & 0.00724 & 0.02053 & 0.01019 & \textcolor{purple}{+7.56\%} 
    & 0.05809 & 0.03624 & 0.09822 & 0.04982 & \textcolor{purple}{+20.08\%} 
    & 0.12609 & 0.08699 & 0.18347 & 0.10539 & \textcolor{purple}{+19.25\%} \\
    
    \textbf{TriMLP} 
    & \textbf{0.01554} & \textbf{0.00963} & 0.02784 & \textbf{0.01357} & \textbf{\textcolor{purple}{+41.63\%}} 
    & \textbf{0.06175} & \textbf{0.03968} & \textbf{0.09997} & \textbf{0.05194} & \textbf{\textcolor{purple}{+26.79\%}} 
    & \textbf{0.13900} & \textbf{0.09636} & \textbf{0.20273} & \textbf{0.11685} & \textbf{\textcolor{purple}{+31.89\%}} \\
    \bottomrule
    \end{tabular}}%
  \label{ablation study}%
  \vspace{-10pt}
\end{table*}%

\subsection{Macro-Design of Triangular Mixer (RQ 2)}
Recall our vanilla implementation of TriMLP contains the complete Triangular Mixer, including both global and local mixing (without cross-session interactions), we derive the following 4 variants to carry on the ablation study:

\begin{itemize}
\item \textbf{EyeMLP} replaces Triangular Mixer with the identity matrix where tokens no longer interact with each other.

\item \textbf{SqrMLP} replaces Triangular Mixer with two standard MLPs where cross-token interactions are no-limited.

\item \textbf{TriMLP\_G} only reserves the global mixing in Triangular Mixer to model the long-range dependencies.

\item \textbf{TriMLP\_L} only reserves the local mixing in Triangular Mixer to capture the local patterns.
\end{itemize}

We consider EyeMLP as baseline, and measure the effectiveness of different variants with the corresponding Average Improvement (Avg. Impv.) against baseline. According to Table \ref{ablation study}, we have the following findings:

\textit{Finding 1: Fully-connection profoundly impairs performance.} Compared to EyeMLP, SqrMLP erodes performances on all validated 12 datasets averagely by 77.99\%. It unveils that the incompatibility between the fully-connection structure and the auto-regressive training fashion. The resulting information leakage is a serious and non-negligible issue which originally motivates this paper. We also explore the feasibility of adopting the auto-encoding manner to ingratiate the bidirectional particularity of MLP in section \ref{atrg vs atec}.

\textit{Finding 2: Triangular design does matter.} Both TriMLP\_G and TriMLP\_L remarkably boost the performance against EyeMLP, where the leading margins achieve up to 27.55\% and 26.05\%, singly. It demonstrates that our triangular design sufficiently evokes the sequential modeling potential in MLP under the auto-regressive training mode.

\textit{Finding 3: Two mixing layers complement each other.} Triangular Mixer constantly attains superior performance than solely employing either global or local mixing. It shows that jointly utilizing these two mixing layers is productive to the fine-grained modeling of long and short-term preferences. We compare the performance of serial-connected mixing layers with other internal structures in Triangular Mixer in Section \ref{different structures}, and visualize how is the mutual influence among these two mixing branches in Section \ref{reception field}.

\begin{figure*}[t]
\centering
	\subfigure[Beauty]{
	\includegraphics[scale=0.14725]{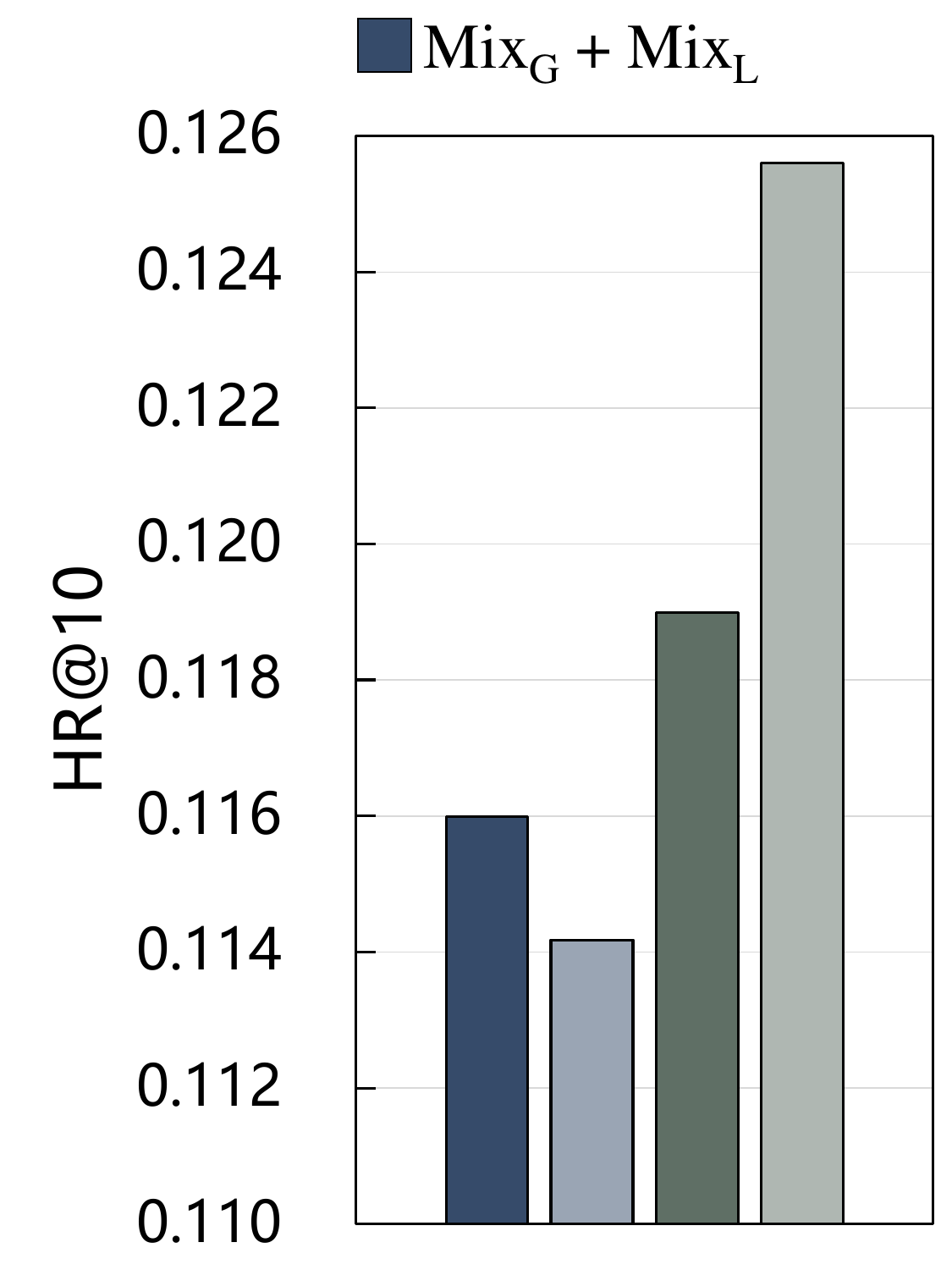}}
	\vspace{-10pt}
	\subfigure[Sports]{
	\includegraphics[scale=0.14725]{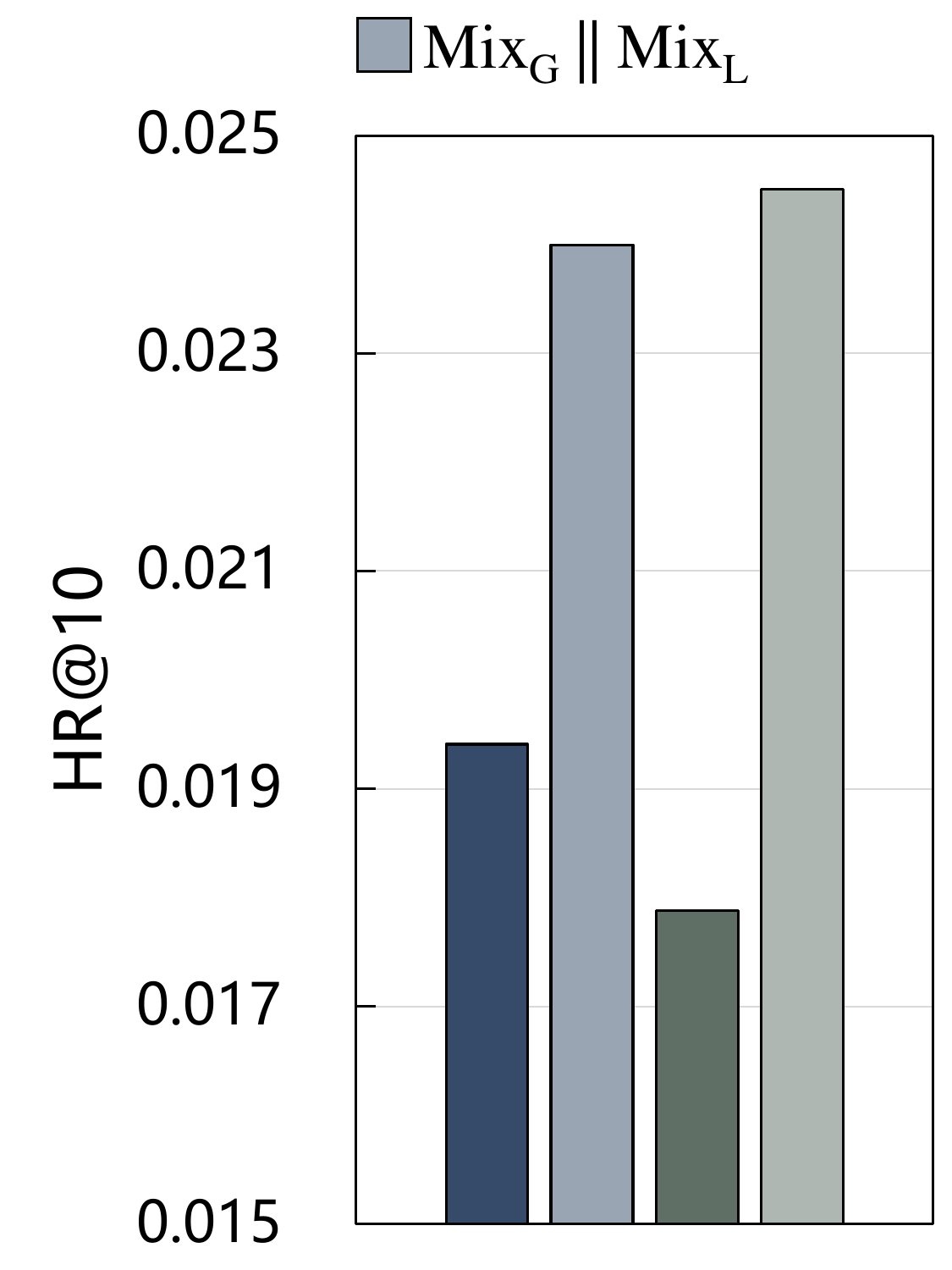}}
	\subfigure[ML-100K]{
	\includegraphics[scale=0.14725]{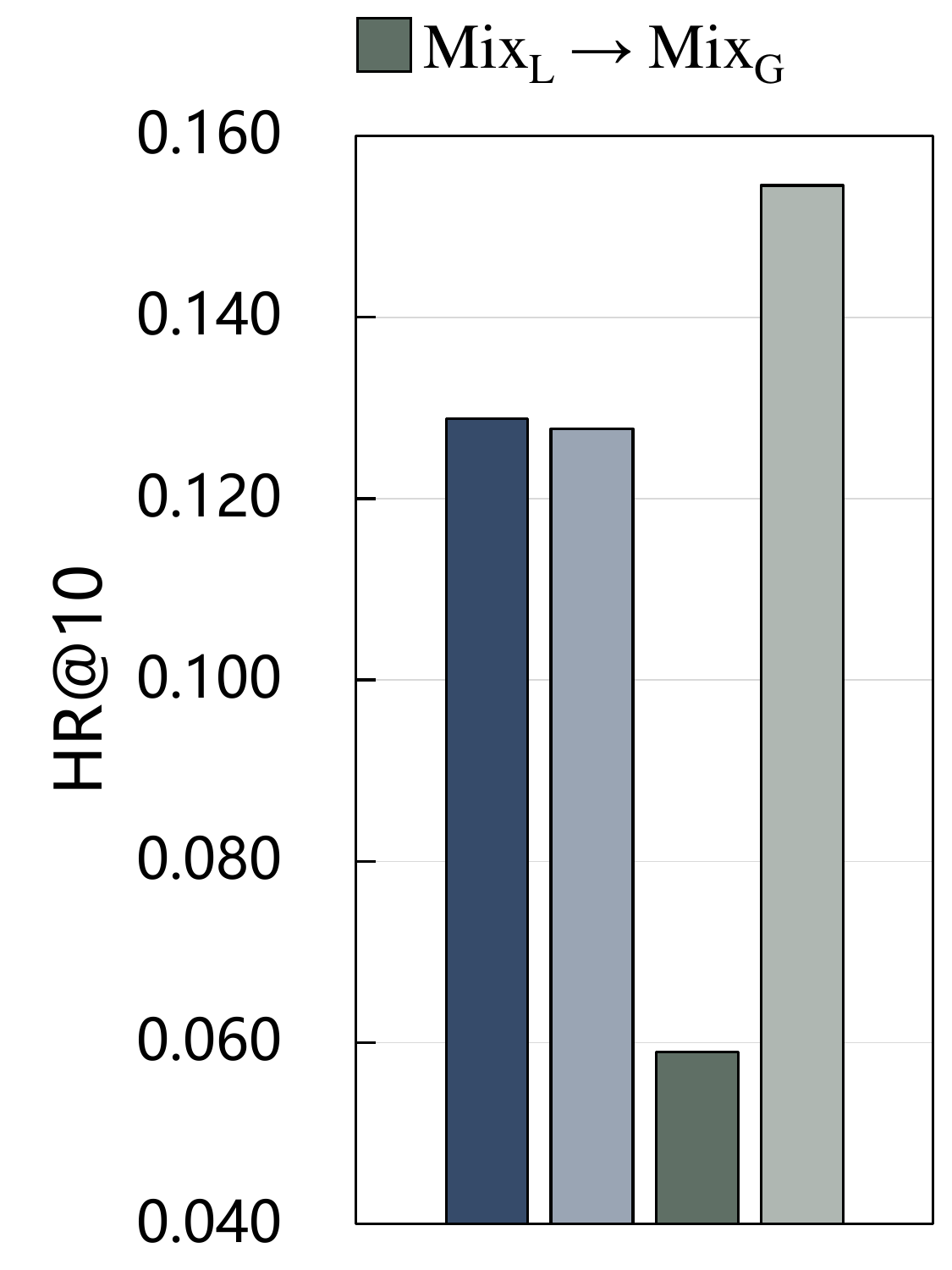}}
	\subfigure[NYC]{
	\includegraphics[scale=0.14725]{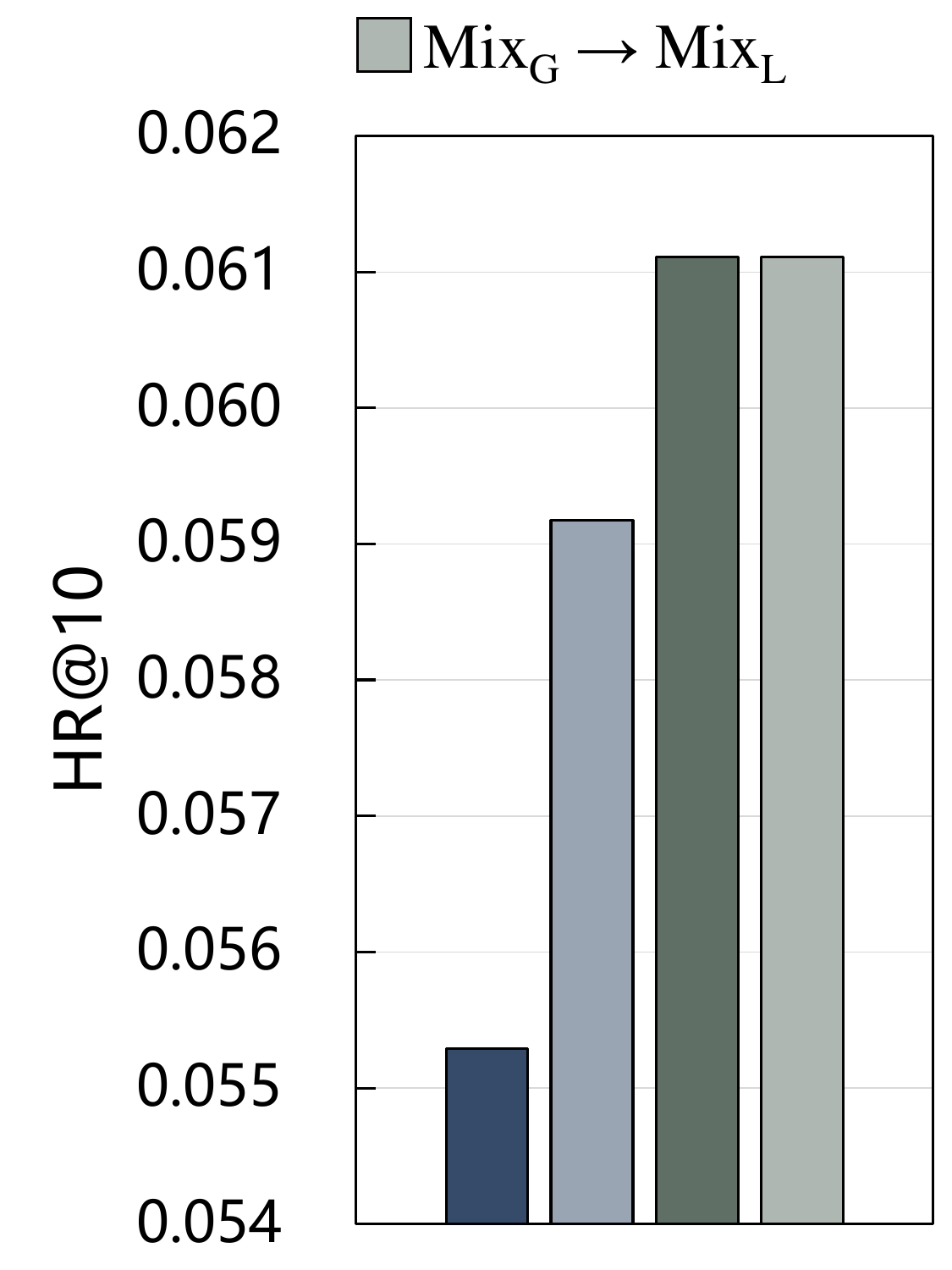}}
	\subfigure[QB-Article]{
	\includegraphics[scale=0.14725]{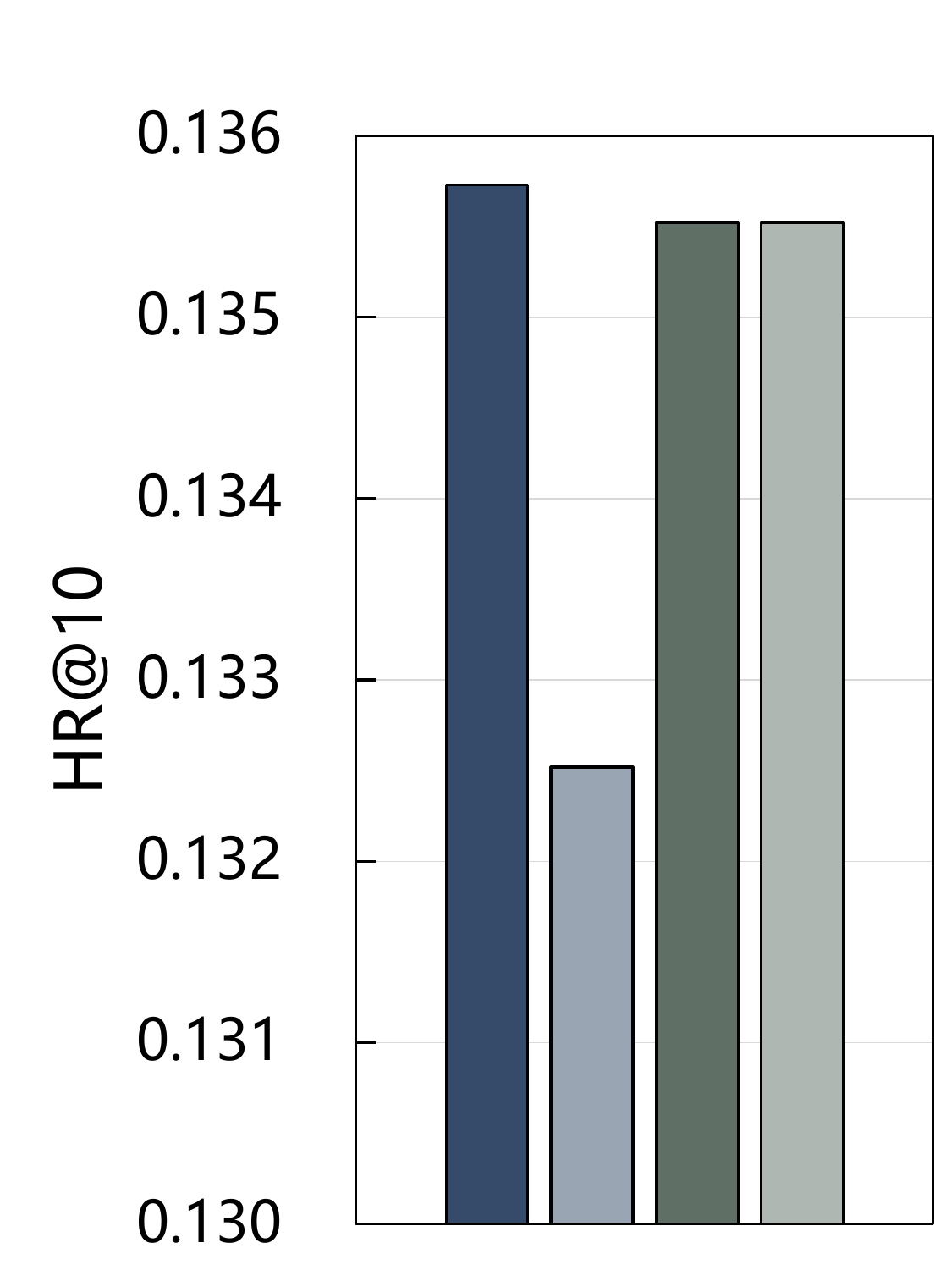}}
	\subfigure[TKY]{
	\includegraphics[scale=0.14725]{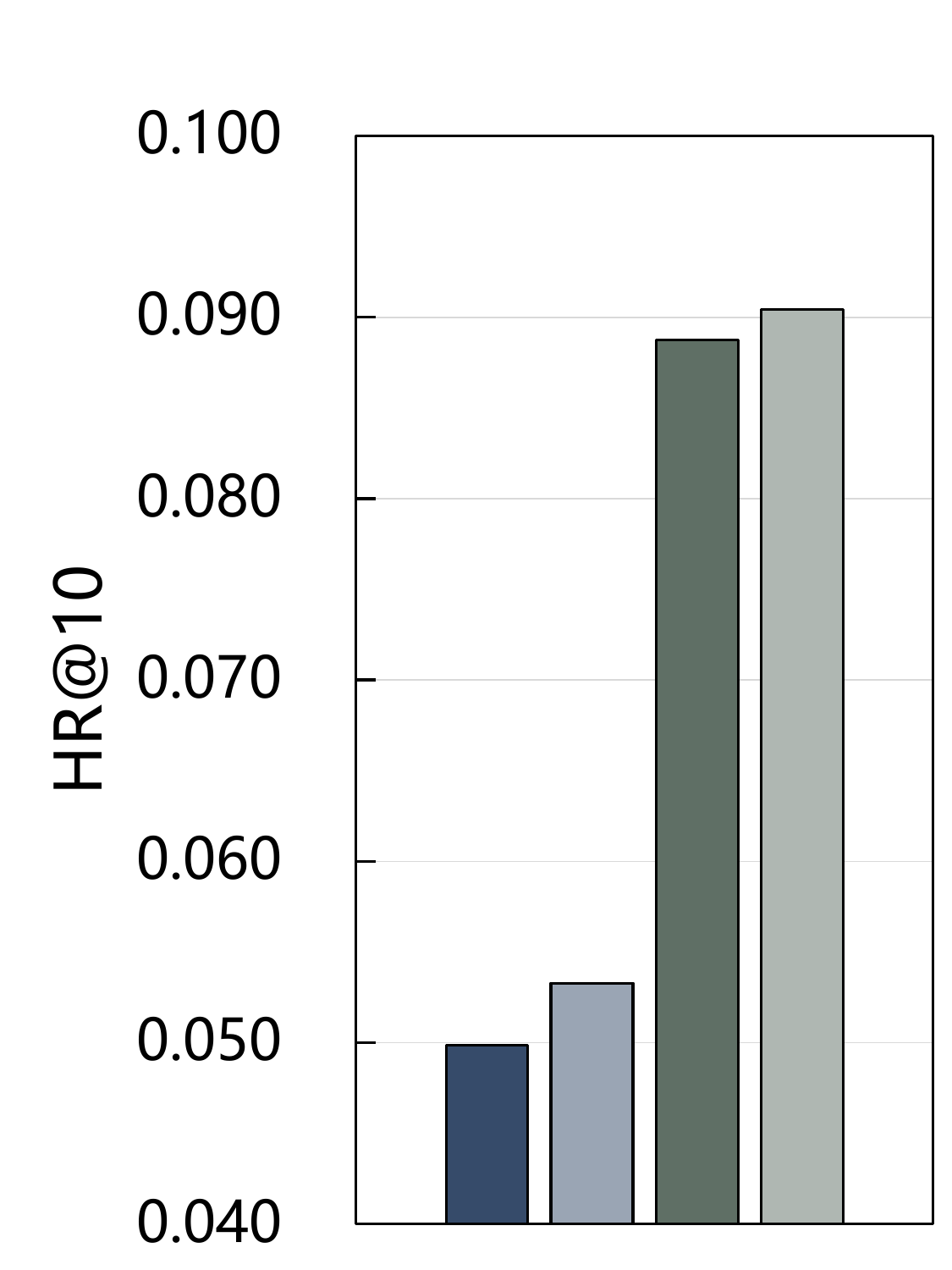}}
	\subfigure[ML-1M]{
	\includegraphics[scale=0.14725]{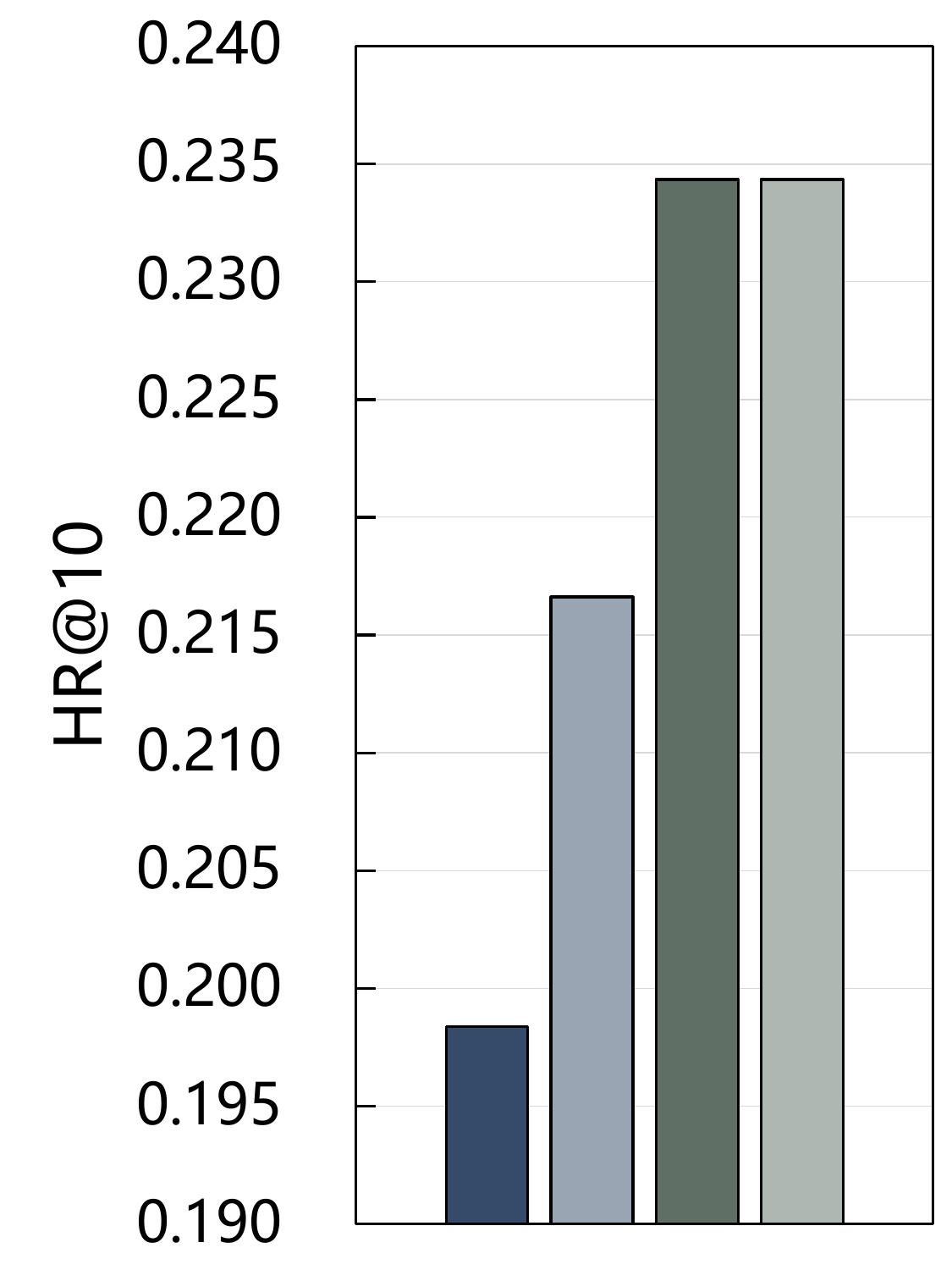}}
	\subfigure[QB-Video]{
	\includegraphics[scale=0.14725]{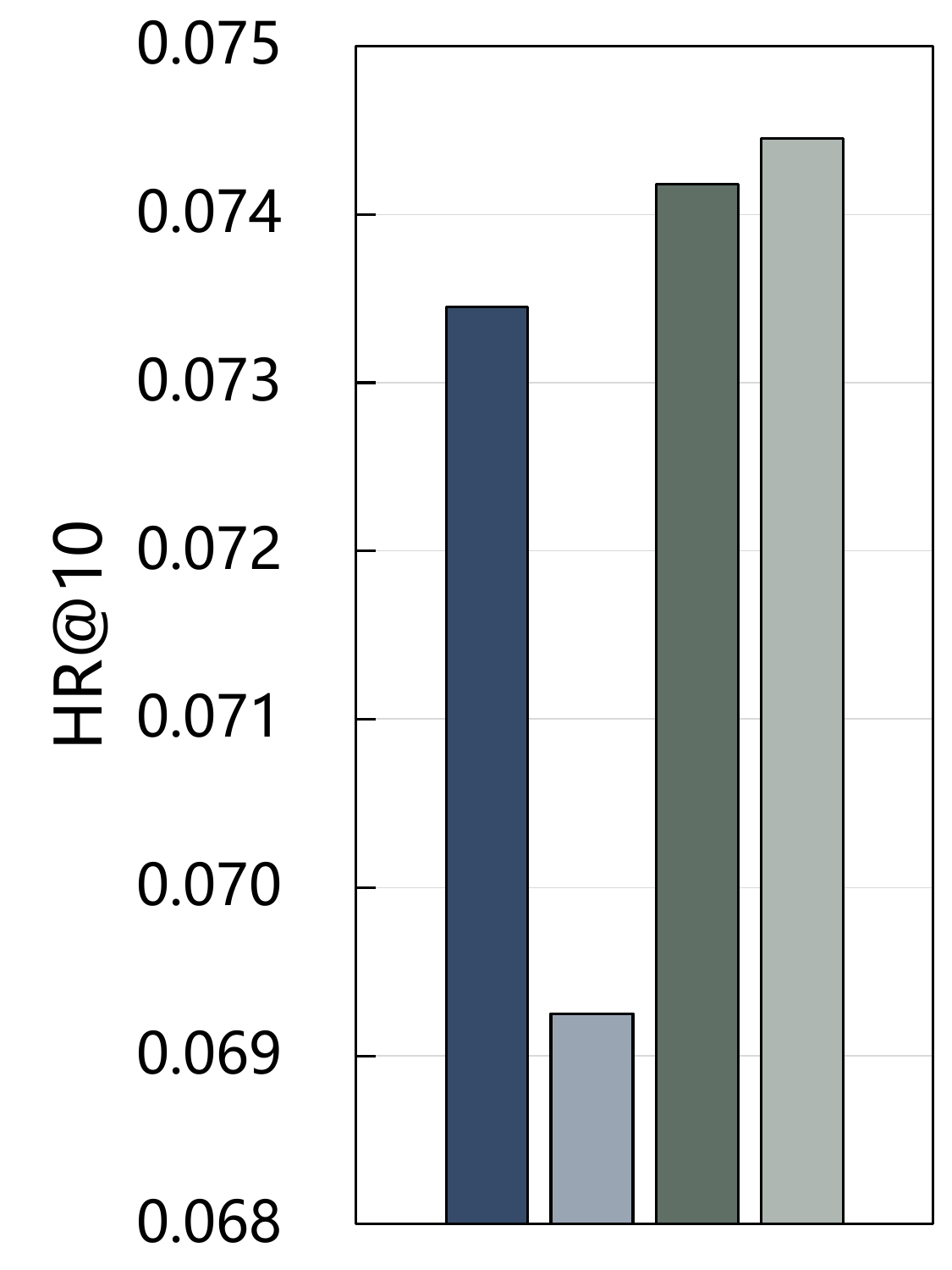}}
	\subfigure[Brightkite]{
	\includegraphics[scale=0.14725]{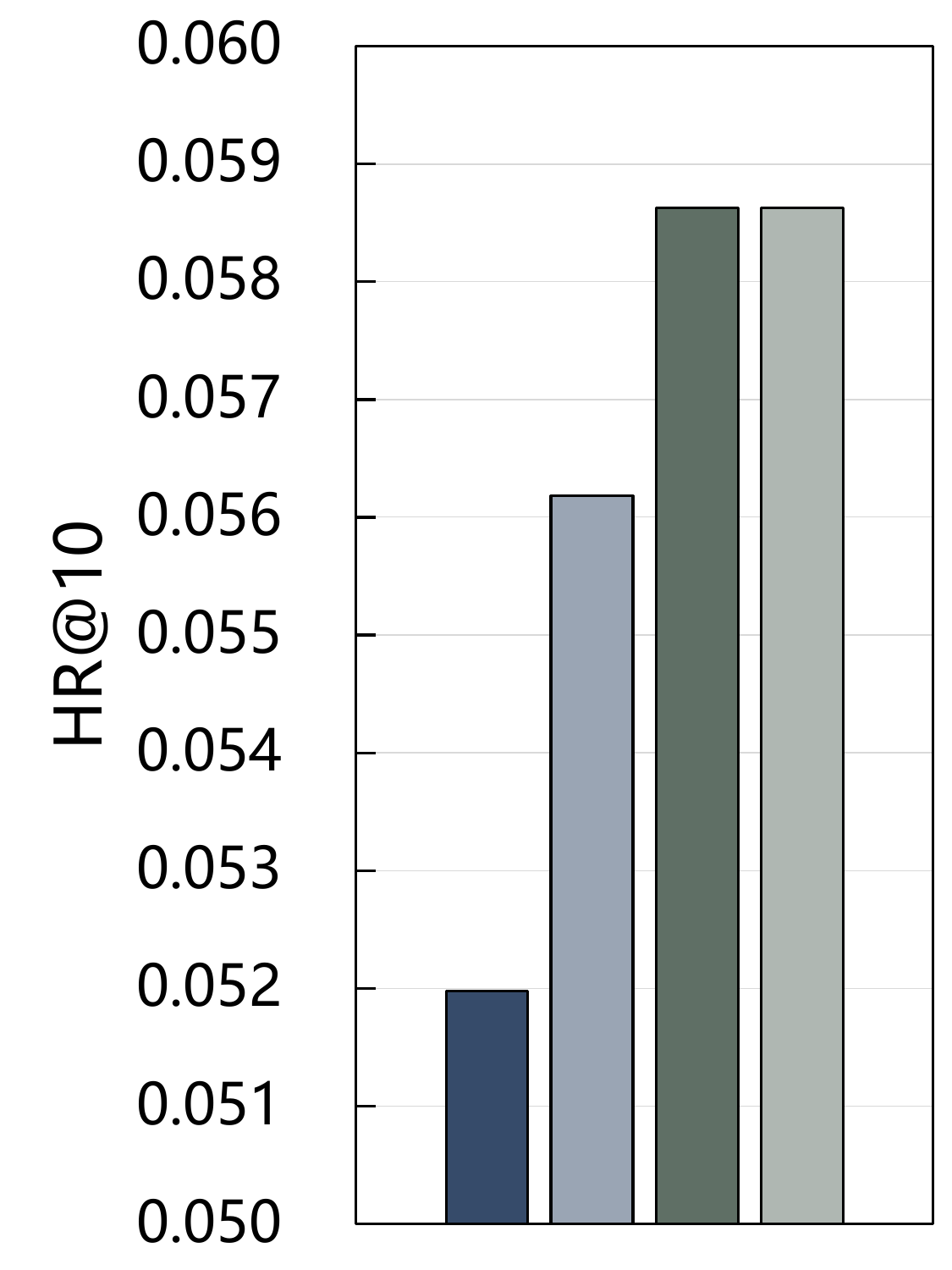}}
	\subfigure[Yelp]{
	\includegraphics[scale=0.14725]{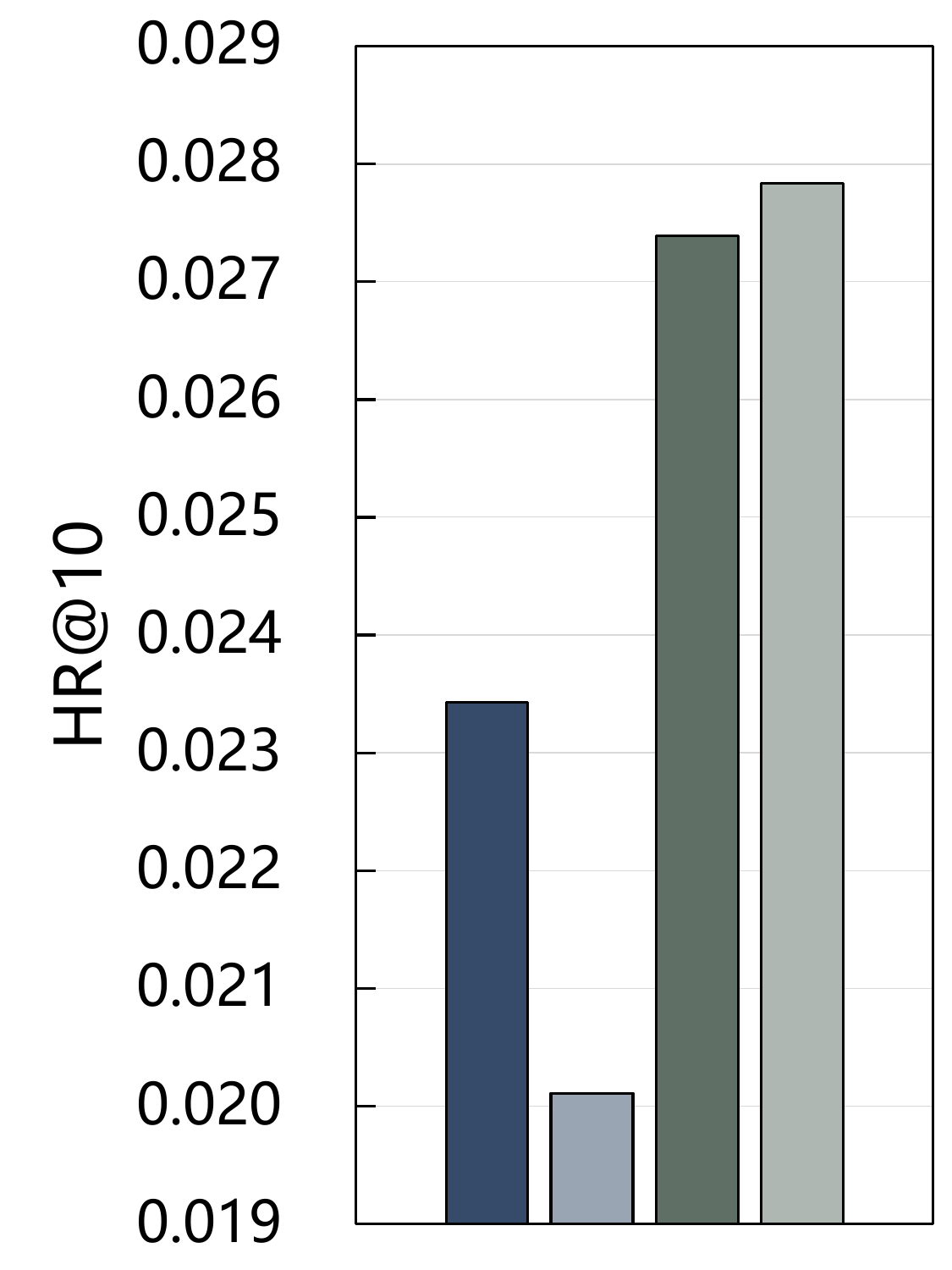}}
	\subfigure[Gowalla]{
	\includegraphics[scale=0.14725]{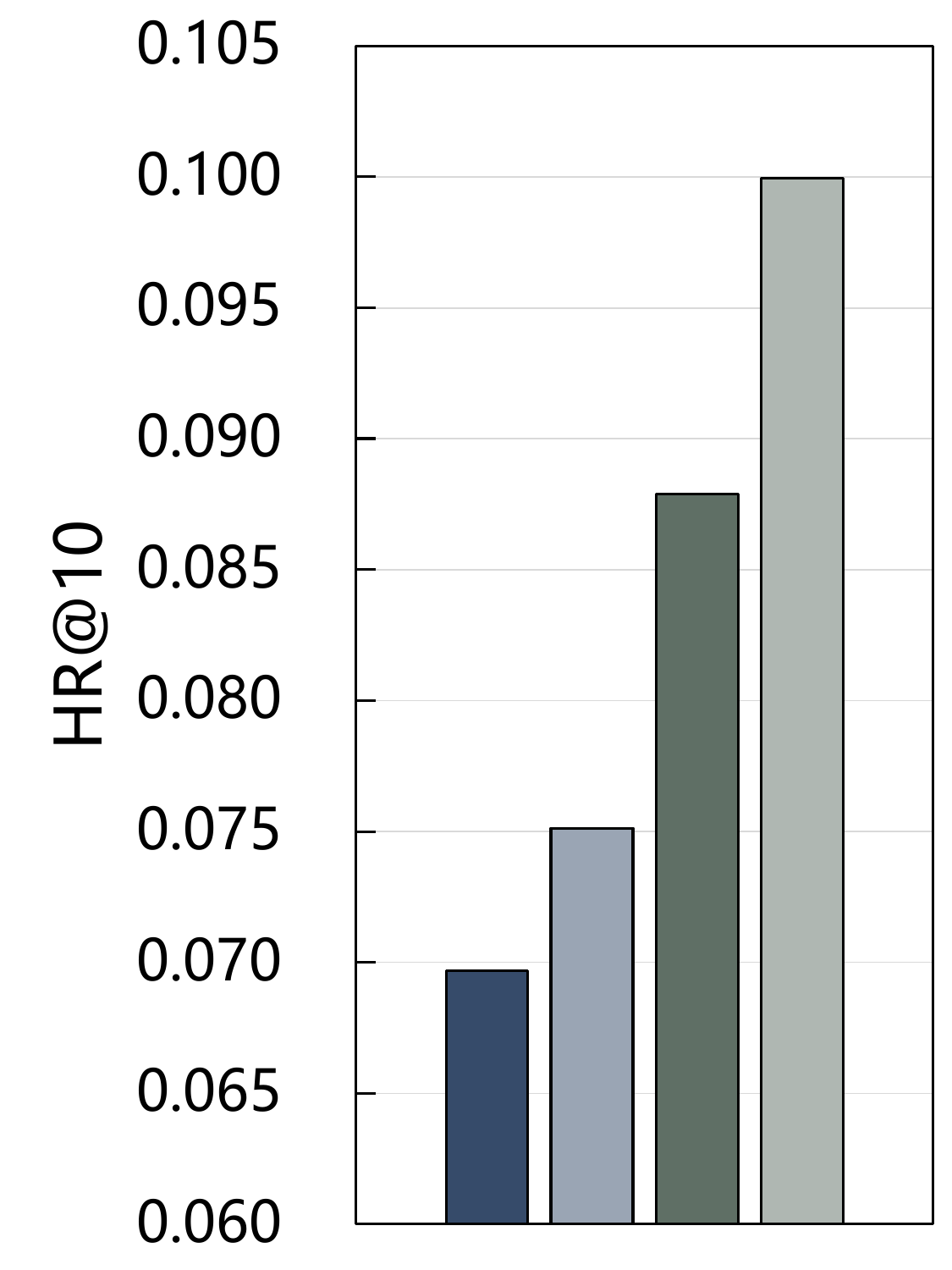}}
	\subfigure[ML-10M]{
	\includegraphics[scale=0.14725]{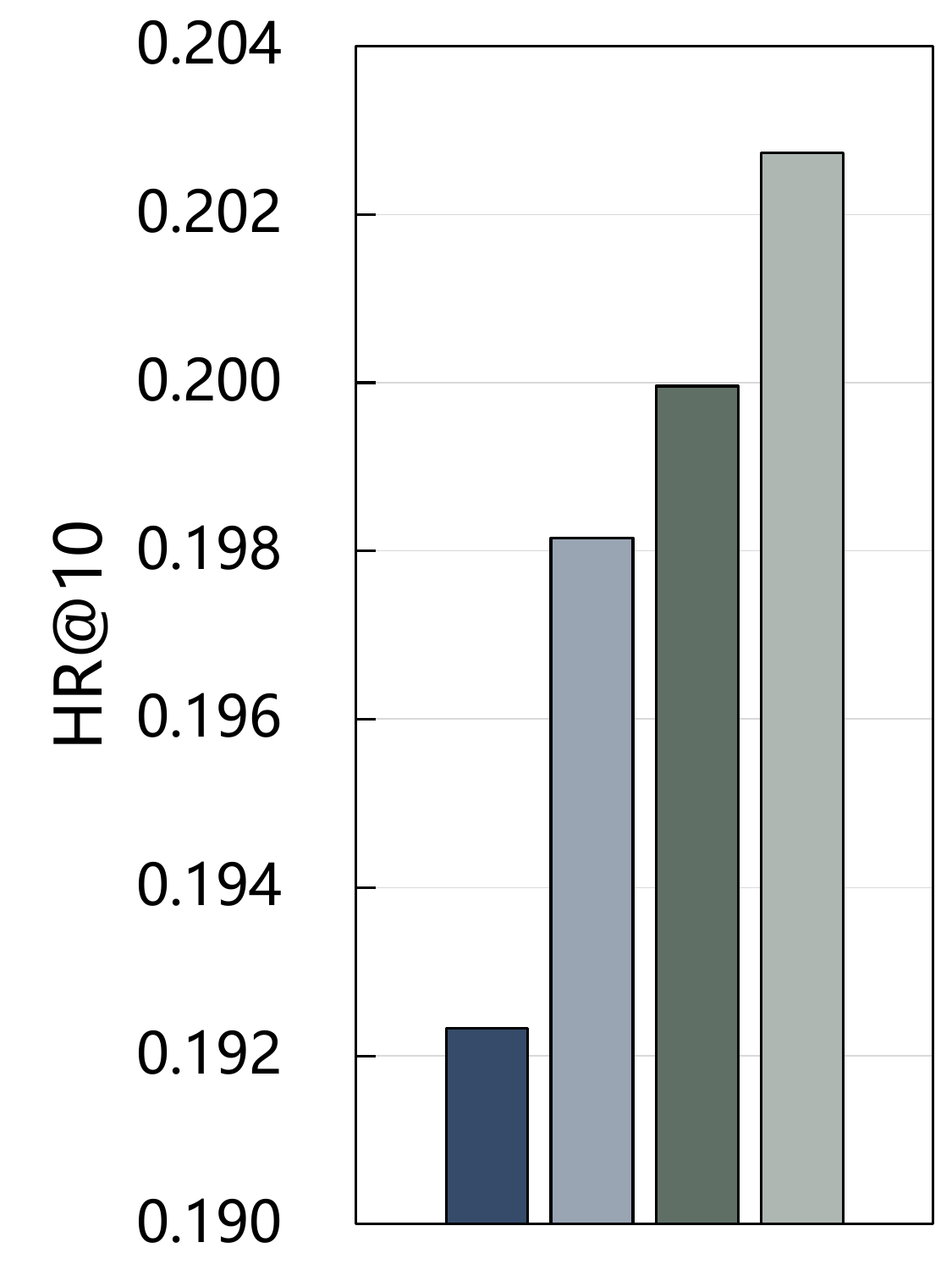}}
\caption{Recommendation performance comparison (HR@10) of Triangular Mixer with different internal structures.}
\label{Internal Structure}
\vspace{-10pt}
\end{figure*}

\subsection{Micro-Design of Triangular Mixer (RQ 3)}
This section analyzes the intrinsic properties of Triangular Mixer by decomposing it into various internal structures and operation components.

\subsubsection{Sensitivity w.r.t. Different Internal Structures}\label{different structures}
Recall that the vanilla Triangular Mixer is denoted as \textbf{Mix\_G $\rightarrow$ Mix\_L}, we consider the following 3 variants:
\begin{itemize}
\item\textbf{Mix\_L $\rightarrow$ Mix\_G} follows the serial-connection. It first encodes the local patterns by local mixing and then models the long-range dependency with global mixing.
\item\textbf{Mix\_G + Mix\_L} employs the parallel structure that combines the results of independent global and local mixing branches with element-wise addition.
\item\textbf{Mix\_G $||$ Mix\_L} employs the parallel structure that concatenates the results of independent global and local mixing branches and merges them with linear layer.
\end{itemize}

As depicted in Figure \ref{Internal Structure}, the serial combinations generally perform better than the parallel ones on most datasets, and Mix\_G $\rightarrow$ Mix\_L achieves more stable scores. Note that Mix\_L $\rightarrow$ Mix\_G has extremely poor performances on Tiny datasets. It might be caused by the split of historical sequences, where the inappropriate truncation leads to the non-uniform distribution of time intervals among user-item interactions. Thus, the short and long-term preferences might vary greatly, and the encoded local patterns would mislead the global one.

\subsubsection{Sensitivity w.r.t. Various Operation Components}\label{micro design}
Recall that the vanilla Triangular Mixer drops the positional information and Feed-Forward Network, and utilizes the 1-0 initialization and Softmax normalization, we consider the following 4 alternatives:
\begin{itemize}
\item \textbf{w. PE} injects the absolute order information into the sequence representation by employing the positional embedding \cite{transformer} after the Embedding layer.
\item \textbf{w. FFN} adds the Feed-Forward Network \cite{transformer} after Triangular Mixer, along with the pre-layer-normalization \cite{layernorm, prenorm} and residual connection \cite{resconnect}.
\item \textbf{w.o. 1-0 Init} initializes the mixing kernels with the default kaiming uniform distribution \cite{kaiuniform} in PyTorch.
\item \textbf{w.o. Softmax} removes the Softmax operation conducted on the mixing kernels.
\end{itemize}

\begin{table*}[t]
  \centering
  \caption{Sensitivity w.r.t. Various Operation Components in Global and Local Mixing. The best performance is boldfaced. \colorbox{purple!12.5}{The vanilla implementation is marked with purple shading}.}
  \resizebox{\textwidth}{!}{
    \begin{tabular}{ll|cccc|cccc|cccc}
    \toprule
    \multicolumn{2}{l|}{\textbf{Dataset-Tiny}} & \multicolumn{4}{c|}{\textbf{Beauty}} & \multicolumn{4}{c|}{\textbf{Sports}} & \multicolumn{4}{c}{\textbf{ML-100K}} \\
    \midrule
    \multicolumn{1}{l|}{\textbf{Variant}} & \multicolumn{1}{c|}{\textbf{Alternative}} & \textbf{HR@5$\uparrow$} & \textbf{NDCG@5$\uparrow$} & \textbf{HR@10$\uparrow$} & \textbf{NDCG@10$\uparrow$} & \textbf{HR@5$\uparrow$} & \textbf{NDCG@5$\uparrow$} & \textbf{HR@10$\uparrow$} & \textbf{NDCG@10$\uparrow$} & \textbf{HR@5$\uparrow$} & \textbf{NDCG@5$\uparrow$} & \textbf{HR@10$\uparrow$} & \textbf{NDCG@10$\uparrow$} \\
    \midrule
    \multicolumn{1}{l|}{\multirow{2}[2]{*}{\textbf{Layer}}} & \multicolumn{1}{c|}{\textbf{w. PE}} & 0.00060 & 0.00060 & 0.00601 & 0.00227 & 0.00102 & 0.00077 & 0.00664 & 0.00257 & 0.02253 & 0.01492 & 0.04614 & 0.02244 \\
    \multicolumn{1}{l|}{} & \multicolumn{1}{c|}{\textbf{w. FFN}} & 0.09195 & 0.06939 & 0.12260 & 0.07910 & 0.01634 & 0.01057 & 0.01839 & 0.01118 & 0.06009 & 0.03815 & 0.11695 & 0.05597 \\
    \midrule
    \multicolumn{1}{l|}{\multirow{2}[2]{*}{\textbf{Normalization}}} & \multicolumn{1}{c|}{\textbf{w.o. 1-0 Init.}} & 0.08534 & 0.06288 & 0.11599 & 0.07282 & 0.01532 & 0.01022 & 0.02145 & 0.01211 & 0.08369 & 0.05541 & 0.13627 & 0.07189 \\
    \multicolumn{1}{l|}{} & \multicolumn{1}{c|}{\textbf{w.o. Softmax}} & 0.00781 & 0.00444 & 0.01262 & 0.00603 & 0.00000 & 0.00000 & 0.00102 & 0.00030 & 0.01395 & 0.00726 & 0.03004 & 0.01245 \\
    \midrule
    \rowcolor{purple!12.5}
    \multicolumn{2}{l|}{\textbf{Triangular Mixer}} & \textbf{0.09615} & \textbf{0.07070} & \textbf{0.12560} & \textbf{0.08003} & \textbf{0.01839} & \textbf{0.01258} & \textbf{0.02451} & \textbf{0.01442} & \textbf{0.08691} & \textbf{0.05848} & \textbf{0.15451} & \textbf{0.07988} \\
    \midrule
    \midrule
    \multicolumn{2}{l|}{\textbf{Dataset-Small}} & \multicolumn{4}{c|}{\textbf{NYC}} & \multicolumn{4}{c|}{\textbf{QB-Article}} & \multicolumn{4}{c}{\textbf{TKY}} \\
    \midrule
    \multicolumn{1}{l|}{\textbf{Variant}} & \multicolumn{1}{c|}{\textbf{Alternative}} & \textbf{HR@5$\uparrow$} & \textbf{NDCG@5$\uparrow$} & \textbf{HR@10$\uparrow$} & \textbf{NDCG@10$\uparrow$} & \textbf{HR@5$\uparrow$} & \textbf{NDCG@5$\uparrow$} & \textbf{HR@10$\uparrow$} & \textbf{NDCG@10$\uparrow$} & \textbf{HR@5$\uparrow$} & \textbf{NDCG@5$\uparrow$} & \textbf{HR@10$\uparrow$} & \textbf{NDCG@10$\uparrow$} \\
    \midrule
    \multicolumn{1}{l|}{\multirow{2}[2]{*}{\textbf{Layer}}} & \multicolumn{1}{c|}{\textbf{w. PE}} & 0.02619 & 0.01469 & 0.05723 & 0.02445 & 0.03961 & 0.02387 & 0.08563 & 0.03858 & 0.02955 & 0.01819 & 0.04985 & 0.02464 \\
    \multicolumn{1}{l|}{} & \multicolumn{1}{c|}{\textbf{w. FFN}} & 0.03977 & 0.02226 & 0.05529 & 0.02750 & 0.06337 & 0.03899 & 0.12096 & 0.05734 & \textbf{0.05382} & 0.03088 & \textbf{0.09263} & 0.04309 \\
    \midrule
    \multicolumn{1}{l|}{\multirow{2}[2]{*}{\textbf{Normalization}}} & \multicolumn{1}{c|}{\textbf{w.o. 1-0 Init}} & 0.04268 & 0.02529 & 0.05723 & 0.02990 & 0.07193 & 0.04433 & \textbf{0.13766} & 0.06523 & 0.04014 & 0.02498 & 0.08690 & 0.04003 \\
    \multicolumn{1}{l|}{} & \multicolumn{1}{c|}{\textbf{w.o. Softmax}} & 0.03298 & 0.02056 & 0.04850 & 0.02548 & 0.01905 & 0.00989 & 0.04624 & 0.01854 & 0.02955 & 0.01865 & 0.04940 & 0.02501 \\
    \midrule
    \rowcolor{purple!12.5}
    \multicolumn{2}{l|}{\textbf{Triangular Mixer}} 
    & \textbf{0.04365} 
    & \textbf{0.02574} 
    & \textbf{0.06111} 
    & \textbf{0.03121} 
    & \textbf{0.07621} 
    & \textbf{0.04751} 
    & 0.13552 
    & \textbf{0.06639} 
    & 0.05293 
    & \textbf{0.03175} 
    & 0.09043 
    & \textbf{0.04384} \\
    \midrule
    \midrule
    \multicolumn{2}{l|}{\textbf{Dataset-Base}} & \multicolumn{4}{c|}{\textbf{ML-1M}} & \multicolumn{4}{c|}{\textbf{QB-Video}} & \multicolumn{4}{c}{\textbf{Brightkite}} \\
    \midrule
    \multicolumn{1}{l|}{\textbf{Variant}} & \multicolumn{1}{c|}{\textbf{Alternative}} & \textbf{HR@5$\uparrow$} & \textbf{NDCG@5$\uparrow$} & \textbf{HR@10$\uparrow$} & \textbf{NDCG@10$\uparrow$} & \textbf{HR@5$\uparrow$} & \textbf{NDCG@5$\uparrow$} & \textbf{HR@10$\uparrow$} & \textbf{NDCG@10$\uparrow$} & \textbf{HR@5$\uparrow$} & \textbf{NDCG@5$\uparrow$} & \textbf{HR@10$\uparrow$} & \textbf{NDCG@10$\uparrow$} \\
    \midrule
    \multicolumn{1}{l|}{\multirow{2}[2]{*}{\textbf{Layer}}} & \multicolumn{1}{c|}{\textbf{w. PE}} & 0.00829 & 0.00503 & 0.02121 & 0.00917 & 0.01885 & 0.01157 & 0.03171 & 0.01569 & 0.04550 & 0.02475 & 0.05880 & 0.02909 \\
    \multicolumn{1}{l|}{} & \multicolumn{1}{c|}{\textbf{w. FFN}} & 0.16042 & 0.10813 & \textbf{0.24130} & 0.13408 & 0.03381 & 0.02029 & 0.06284 & 0.02955 & 0.03903 & 0.02114 & \textbf{0.06860} & 0.03066 \\
    \midrule
    \multicolumn{1}{l|}{\multirow{2}[2]{*}{\textbf{Normalization}}} & \multicolumn{1}{c|}{\textbf{w.o. 1-0 Init}} & 0.16341 & 0.11194 & 0.23616 & \textbf{0.13538} & 0.04048 & 0.02427 & 0.07418 & 0.03505 & 0.04830 & 0.03009 & 0.05915 & 0.03349 \\
    \multicolumn{1}{l|}{} & \multicolumn{1}{c|}{\textbf{w.o. Softmax}} & 0.00795 & 0.00394 & 0.01972 & 0.00770 & 0.00635 & 0.00357 & 0.01449 & 0.00616 & 0.04235 & \textbf{0.03522} & 0.05250 & \textbf{0.03848} \\
    \midrule
    \rowcolor{purple!12.5}
    \multicolumn{2}{l|}{\textbf{Triangular Mixer}} & \textbf{0.16390} & \textbf{0.11196} & 0.23434 & 0.13454 & \textbf{0.04284} & \textbf{0.02550} & \textbf{0.07445} & \textbf{0.03563} & \textbf{0.04848} & 0.03016 & 0.05863 & 0.03335 \\
    \midrule
    \midrule
    \multicolumn{2}{c|}{\textbf{Dataset-Large}} & \multicolumn{4}{c|}{\textbf{Yelp}} & \multicolumn{4}{c|}{\textbf{Gowalla}} & \multicolumn{4}{c}{\textbf{ML-10M}} \\
    \midrule
    \multicolumn{1}{l|}{\textbf{Variant}} & \multicolumn{1}{c|}{\textbf{Alternative}} & \textbf{HR@5$\uparrow$} & \textbf{NDCG@5$\uparrow$} & \textbf{HR@10$\uparrow$} & \textbf{NDCG@10$\uparrow$} & \textbf{HR@5$\uparrow$} & \textbf{NDCG@5$\uparrow$} & \textbf{HR@10$\uparrow$} & \textbf{NDCG@10$\uparrow$} & \textbf{HR@5$\uparrow$} & \textbf{NDCG@5$\uparrow$} & \textbf{HR@10$\uparrow$} & \textbf{NDCG@10$\uparrow$} \\
    \midrule
    \multicolumn{1}{l|}{\multirow{2}[2]{*}{\textbf{Layer}}} & \multicolumn{1}{c|}{\textbf{w. PE}} & 0.00221 & 0.00133 & 0.00370 & 0.00182 & 0.00786 & 0.00466 & 0.01372 & 0.00651 & 0.01553 & 0.00904 & 0.02970 & 0.00136 \\
    \multicolumn{1}{l|}{} & \multicolumn{1}{c|}{\textbf{w. FFN}} & 0.01213 & 0.00765 & 0.02301 & 0.01112 & 0.06113 & 0.03838 & \textbf{0.10013} & 0.05090 & \textbf{0.14109} & \textbf{0.09749} & \textbf{0.20442} & \textbf{0.11786} \\
    \midrule
    \multicolumn{1}{l|}{\multirow{2}[2]{*}{\textbf{Normalization}}} & \multicolumn{1}{c|}{\textbf{w.o. 1-0 Init}} & 0.01512 & 0.00915 & 0.02779 & 0.01319 & 0.06051 & 0.03879 & 0.09902 & 0.05115 & 0.13782 & 0.09557 & 0.20112 & 0.11593 \\
    \multicolumn{1}{l|}{} & \multicolumn{1}{c|}{\textbf{w.o. Softmax}} & 0.00120 & 0.00076 & 0.00207 & 0.00104 & 0.00663 & 0.00393 & 0.01267 & 0.00588 & 0.00638 & 0.00346 & 0.01491 & 0.00618 \\
    \midrule
    \rowcolor{purple!12.5}
    \multicolumn{2}{l|}{\textbf{Triangular Mixer}} & \textbf{0.01554} & \textbf{0.00963} & \textbf{0.02784} & \textbf{0.01357} & \textbf{0.06175} & \textbf{0.03968} & 0.09997 & \textbf{0.05194} & 0.13900 & 0.09636 & 0.20273 & 0.11685 \\
    \bottomrule
    \end{tabular}}%
  \label{ablamixing}%
  \vspace{-15pt}
\end{table*}%

According to the experimental results listed in Table \ref{ablamixing}, we find the following properties of Triangular Mixer:

\textit{Property 1: Positional embedding has conflicts with Triangular Mixer.} Since the mixing layers bust the symmetry of mixing kernels and explicitly endow cross-token interactions in chronological order, the extra positional information becomes redundant and dramatically damages the performances.

\textit{Property 2: FFN brings limited profits in certain scenario.} Through all validated 12 datasets, adding FFN slightly works on \texttt{ML-1M}. Since FFN significantly increases the parameter-scale, it provokes all-MLP architectures to trap in the data-hungry issue on Tiny, Small and Base datasets.

\textit{Property 3: 1-0 initialization proves to be helpful.} Compared to the uniform distributed initialization, our method make all the tokens contribute equally to the targets during the early training stage. It is conducive to avoid the local optimal, especially when the static parameters in MLP are more likely to be troubled with the under-fitting issue.

\textit{Property 4: Softmax prominently promotes performances.} As reported in \cite{mlpmixer, resmlp}, the weights might be irregular, messy and disorganized in standard MLPs. Conducting Softmax on the global and local mixing kernels, i.e., transforms the learnable parameter into probabilities, is instrumental in evolving the weights towards exhuming the relations among tokens.

\begin{figure*}[htbp]
\centering
	\subfigure[Beauty]{
	\includegraphics[scale=0.14725]{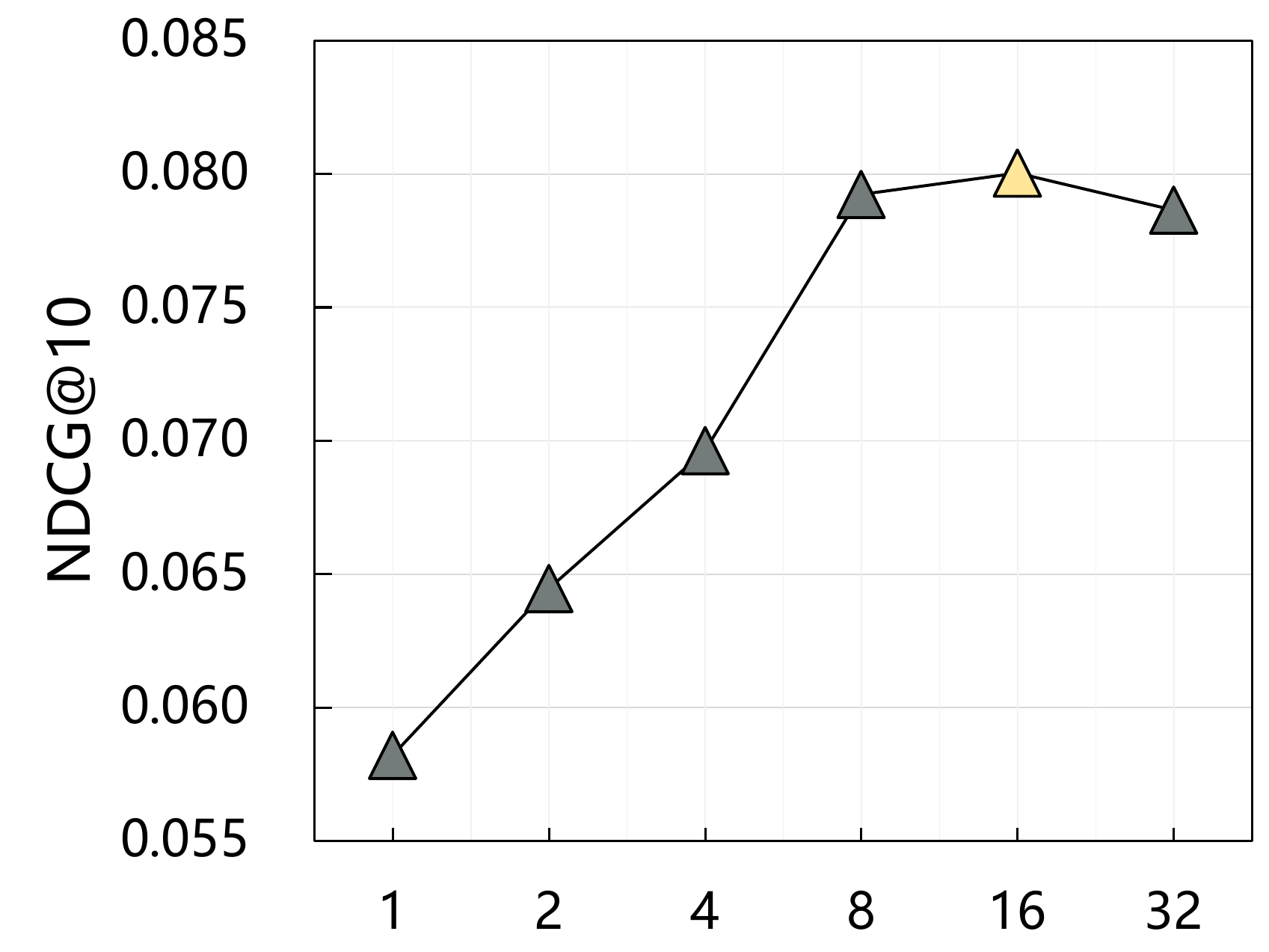}}
	\vspace{-10pt}
	\subfigure[Sports]{
	\includegraphics[scale=0.14725]{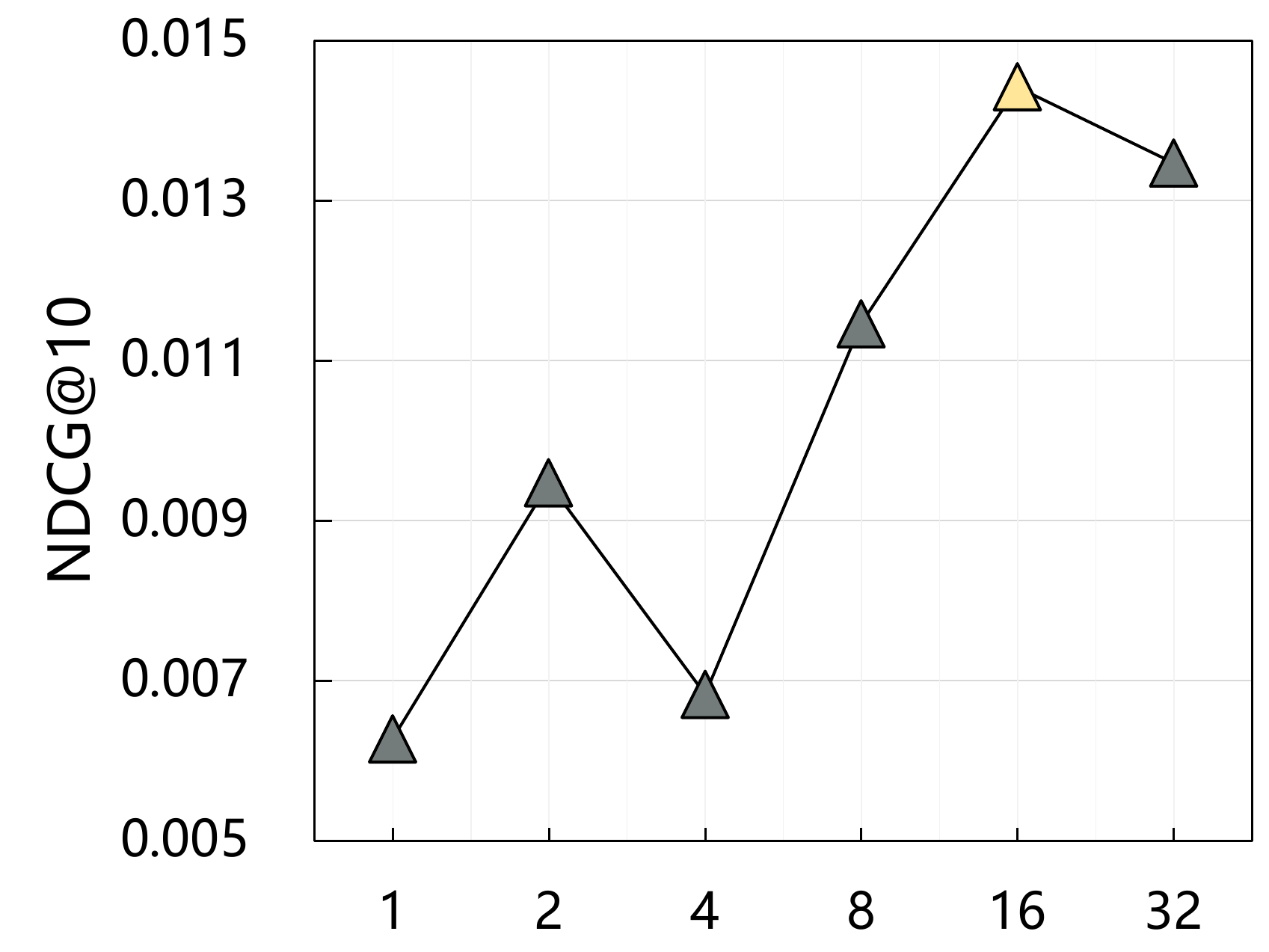}}
	\subfigure[ML-100K]{
	\includegraphics[scale=0.14725]{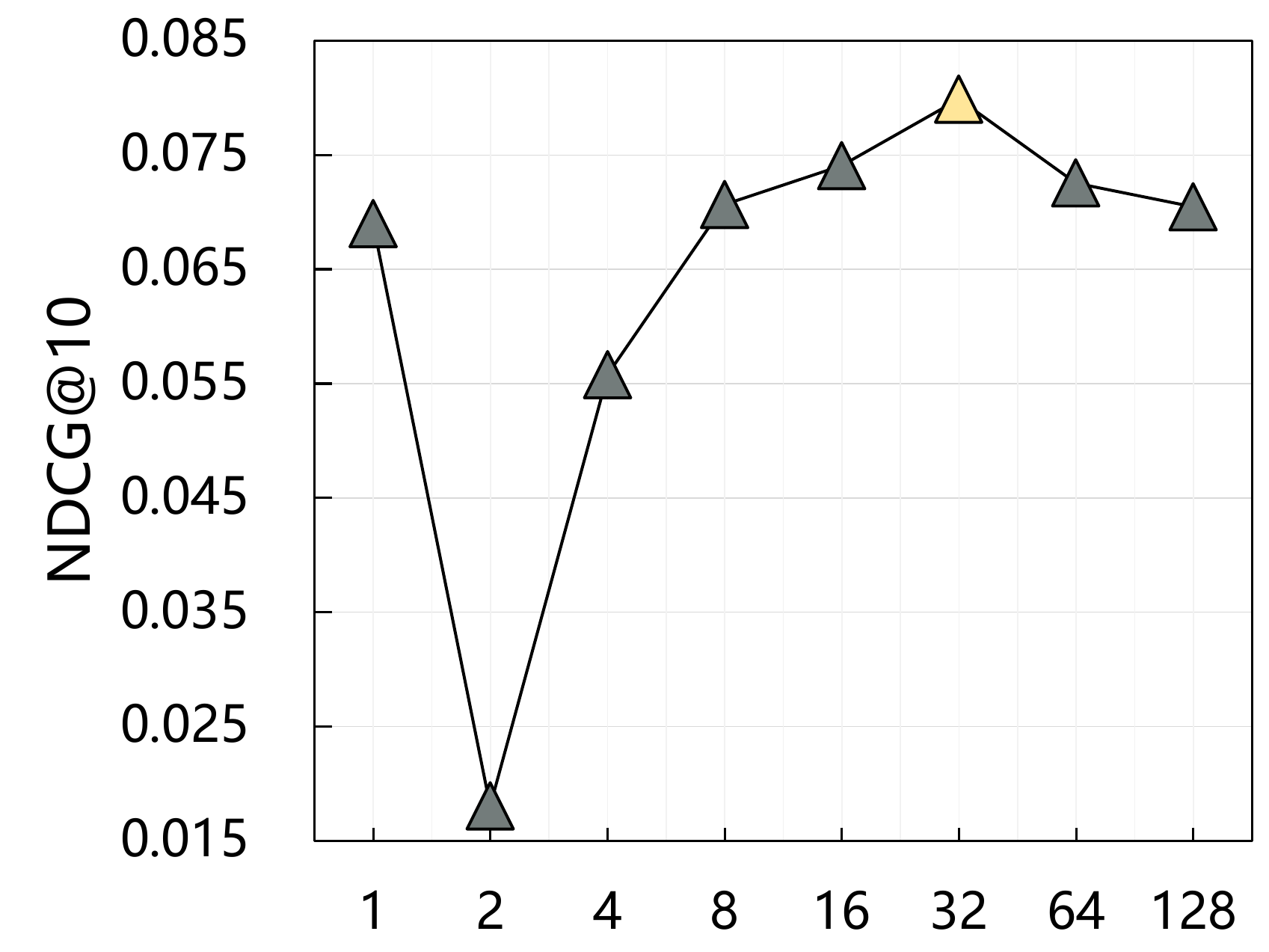}}
	\subfigure[NYC]{
	\includegraphics[scale=0.14725]{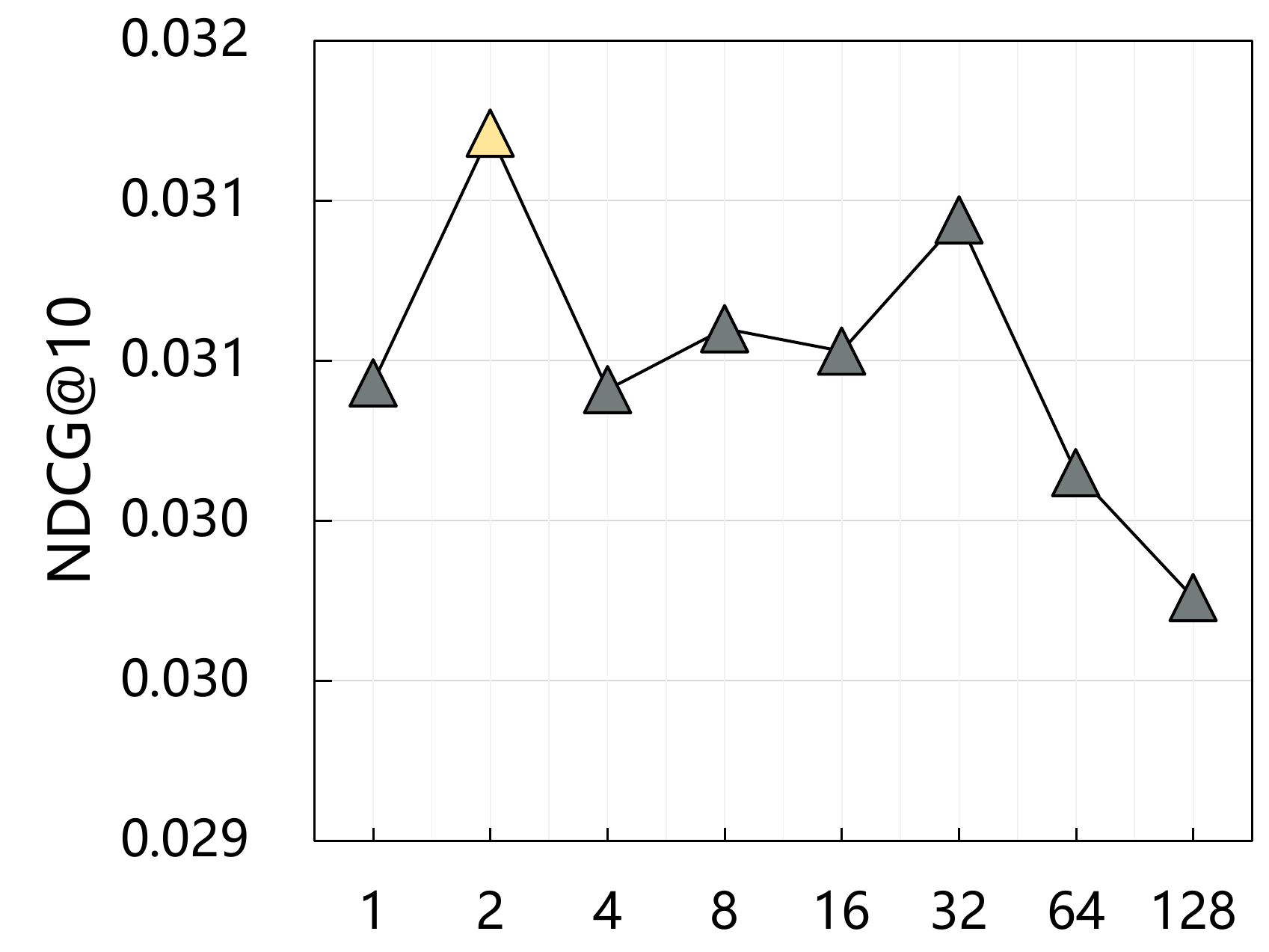}}
	\subfigure[QB-Article]{
	\includegraphics[scale=0.14725]{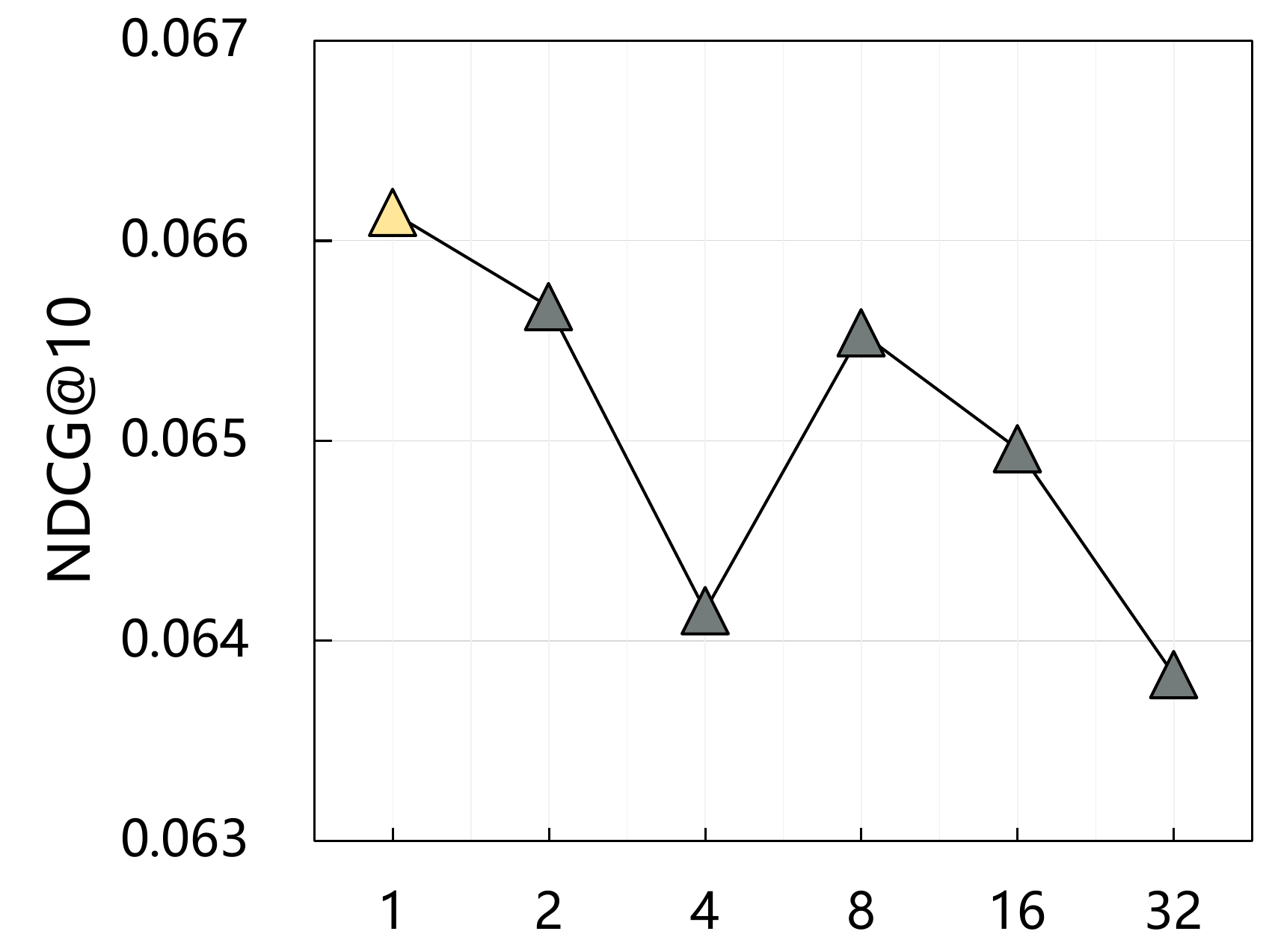}}
	\vspace{-10pt}
	\subfigure[TKY]{
	\includegraphics[scale=0.14725]{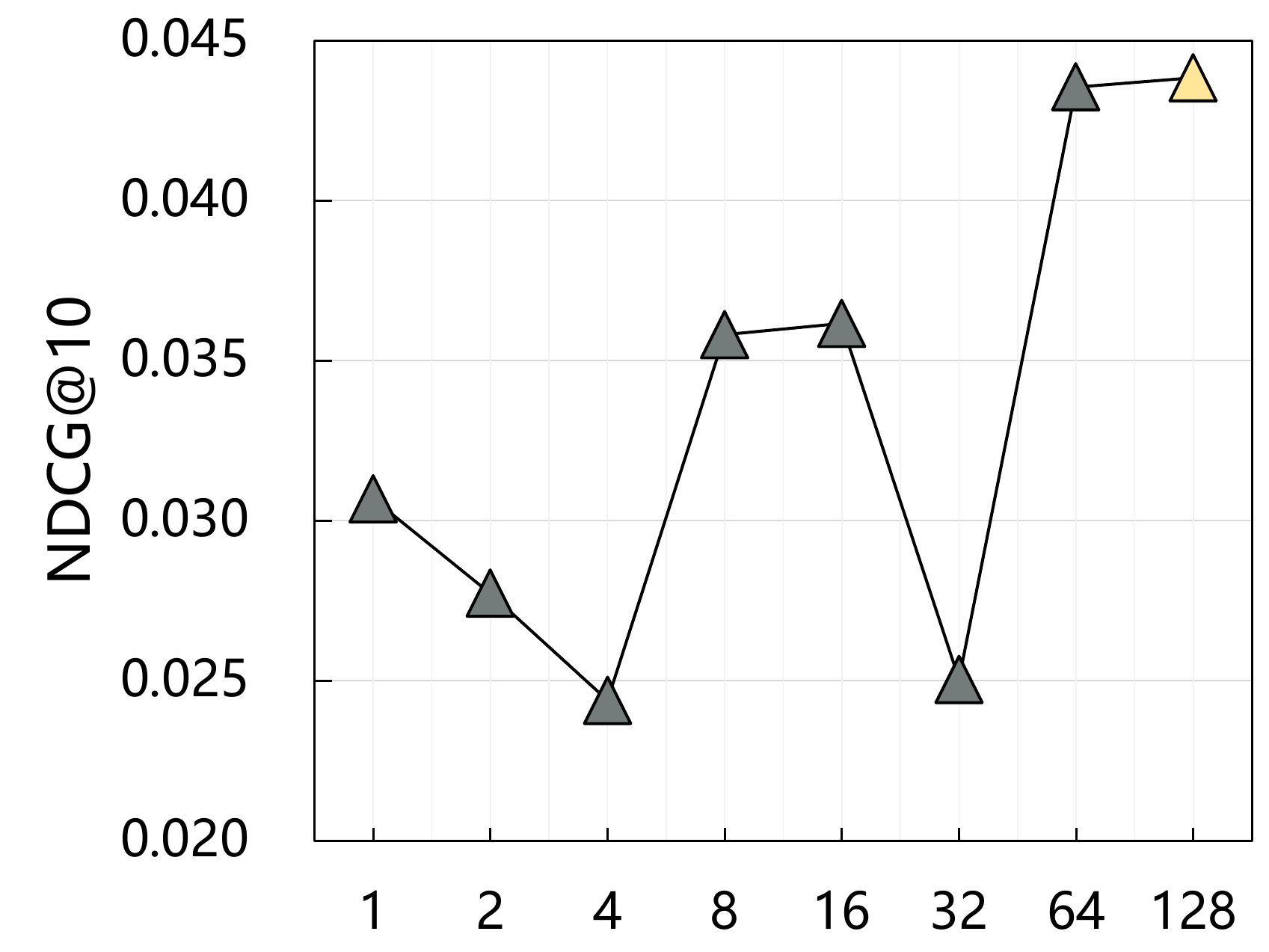}}
	\subfigure[ML-1M]{
	\includegraphics[scale=0.14725]{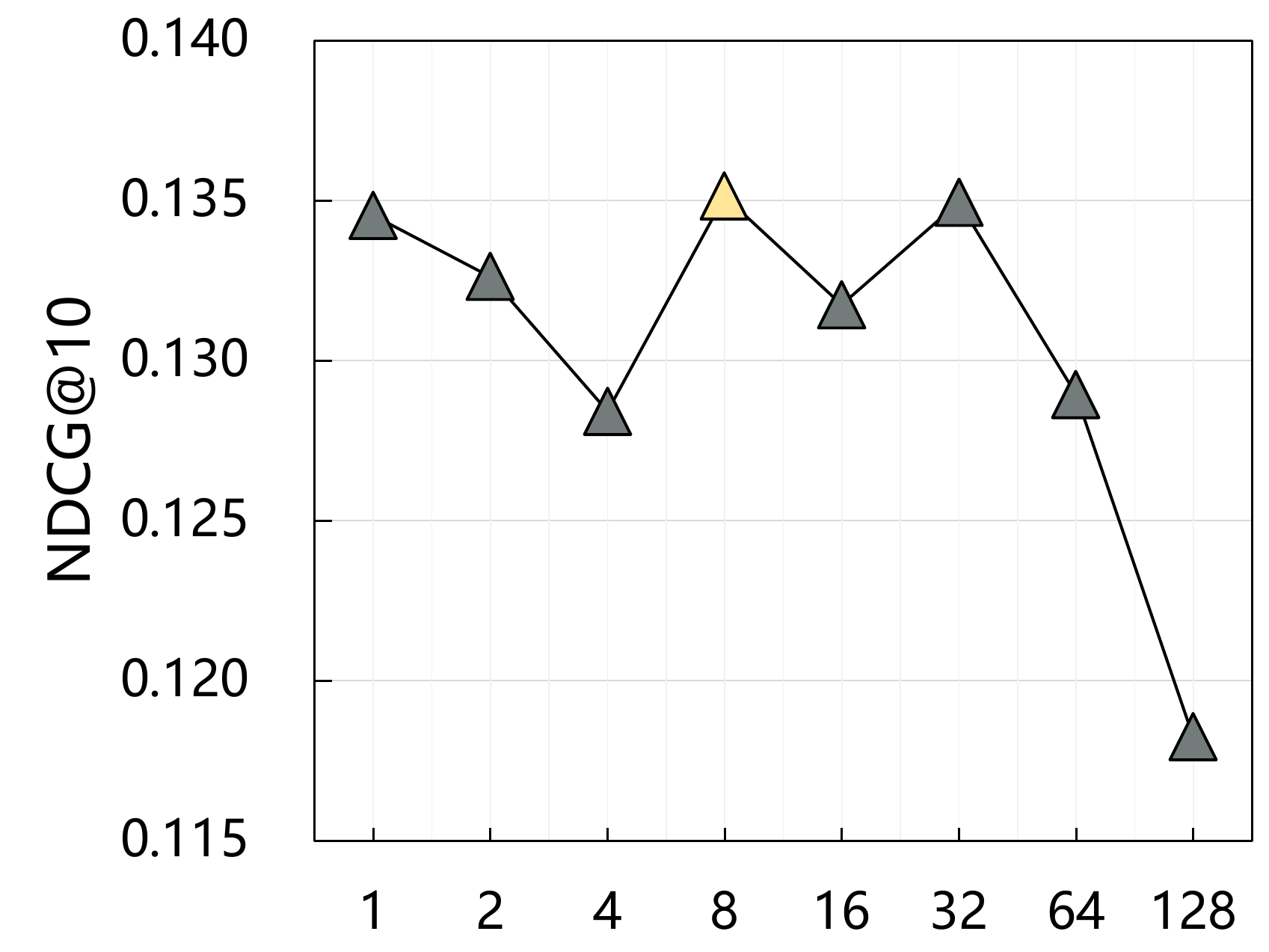}}
	\subfigure[QB-Video]{
	\includegraphics[scale=0.14725]{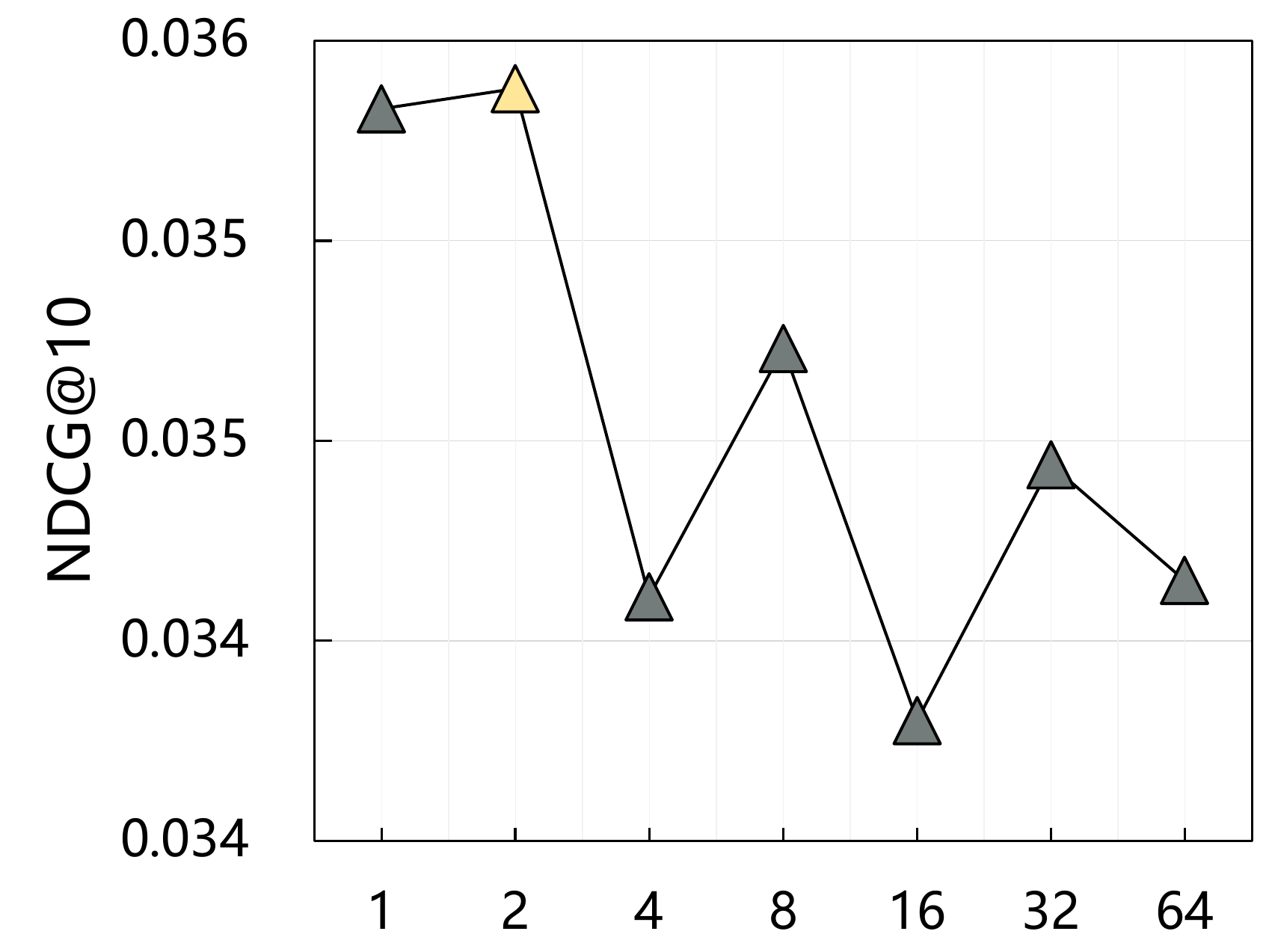}}
	\subfigure[Brightkite]{
	\includegraphics[scale=0.14725]{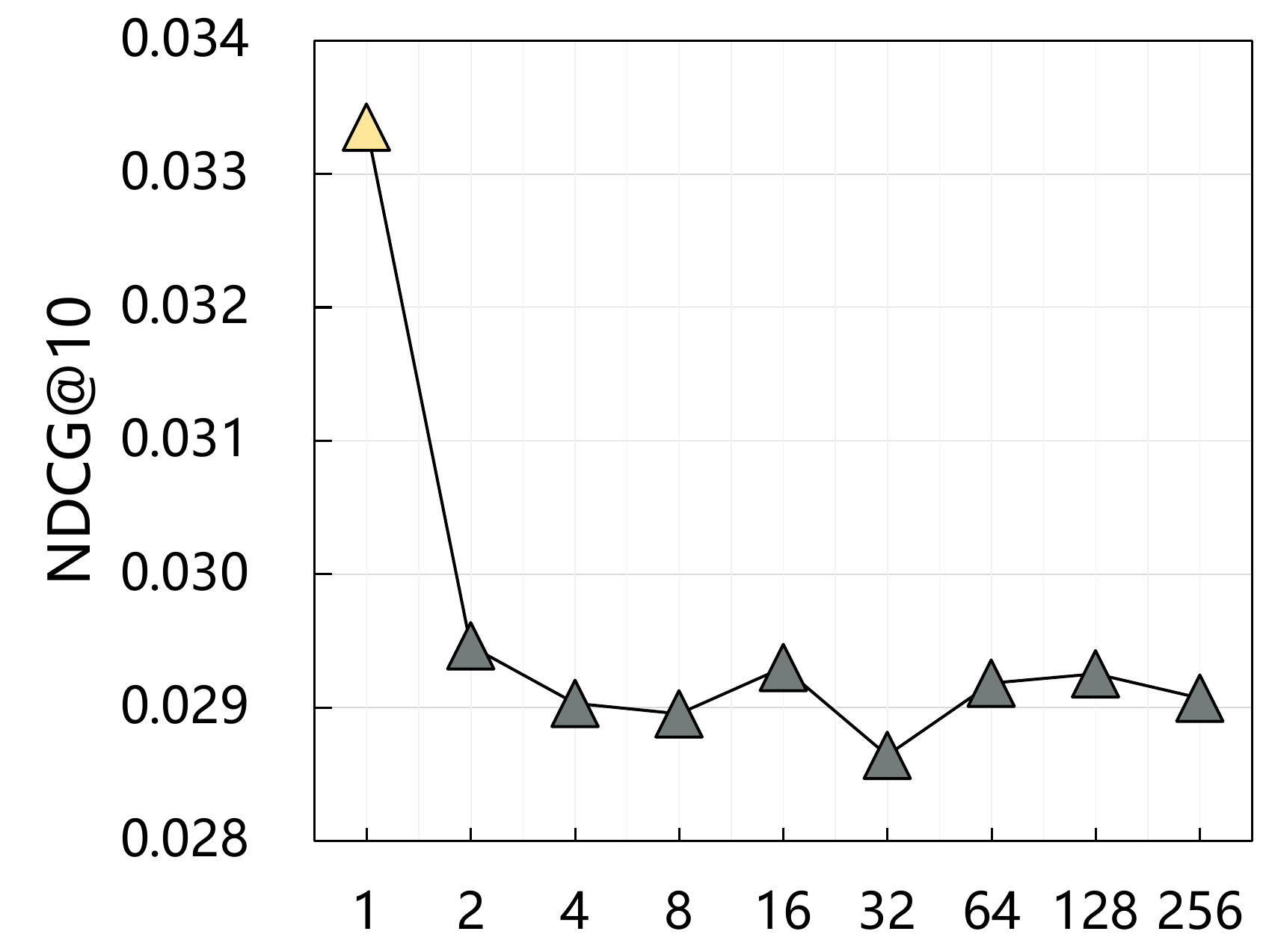}}
	\subfigure[Yelp]{
	\includegraphics[scale=0.14725]{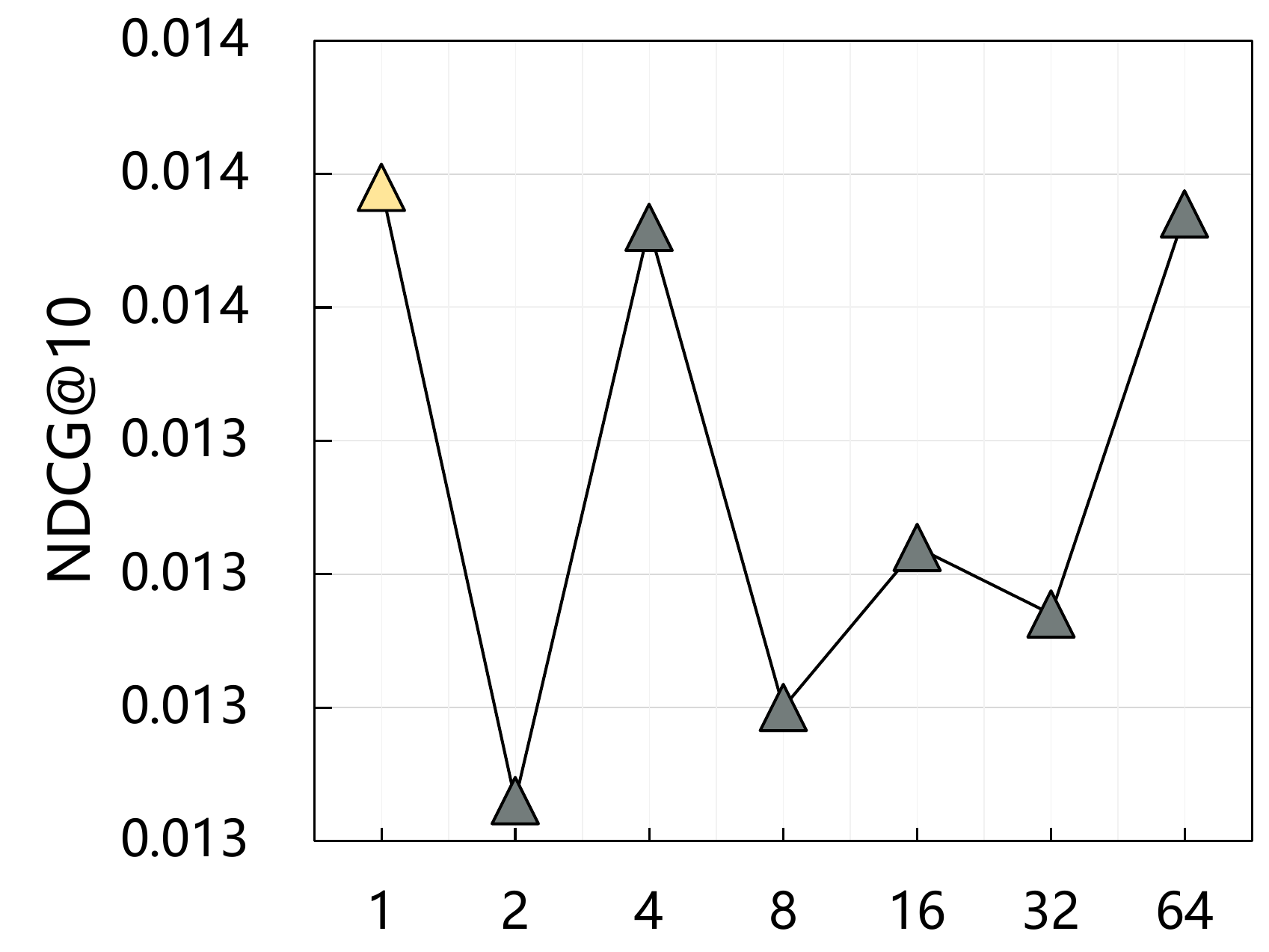}}
	\subfigure[Gowalla]{
	\includegraphics[scale=0.14725]{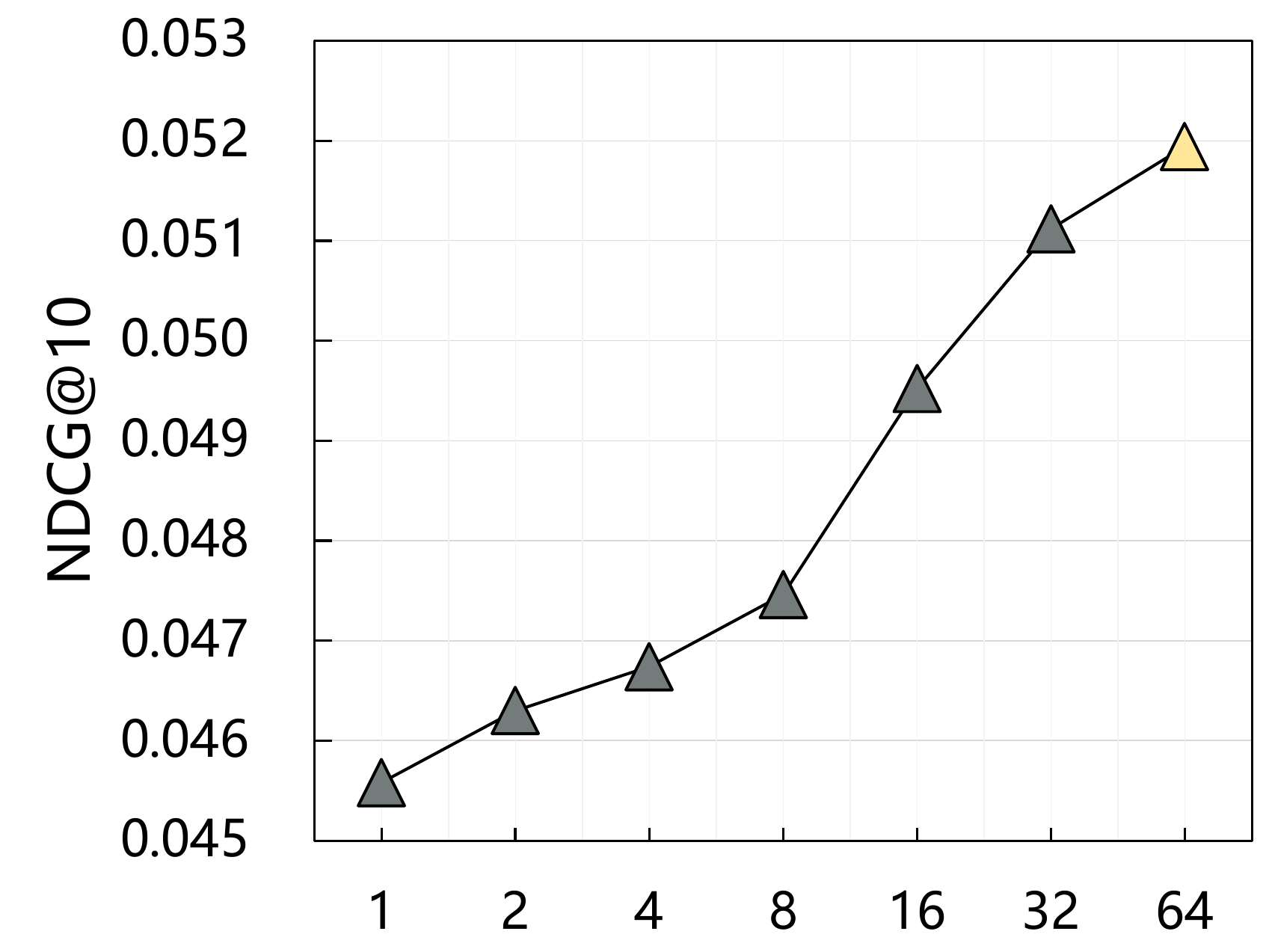}}
	\subfigure[ML-10M]{
	\includegraphics[scale=0.14725]{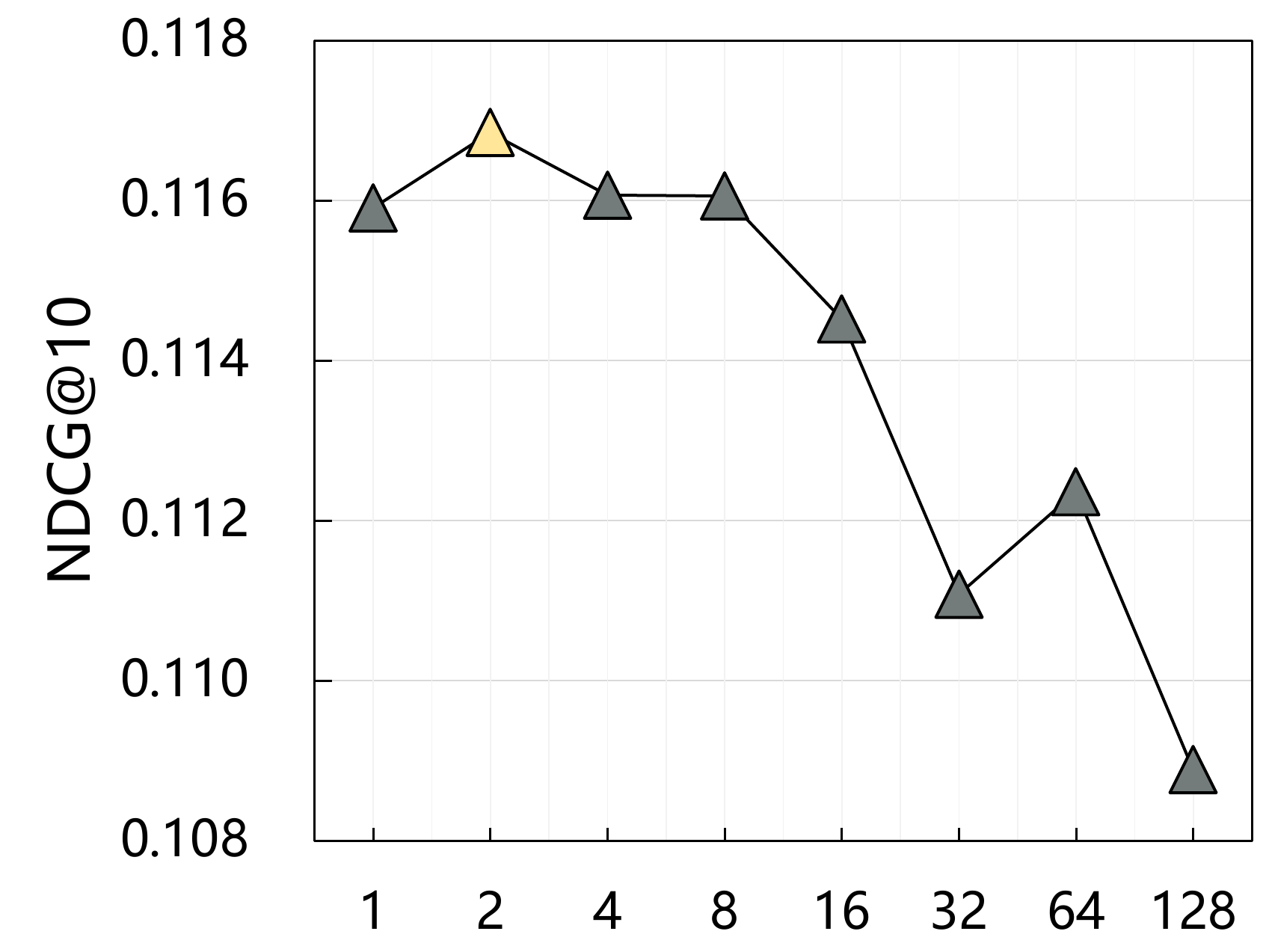}}
\caption{Recommendation performance comparison (NDCG@10) w.r.t. different session number $s$. The horizontal axes of all sub-figures are the variable $s$.}
\label{session number}
\vspace{-15pt}
\end{figure*}

\subsection{Sensitivity w.r.t. Hyper-parameter Setting}

We mainly verify the influence brought by setting different session number $s$ (or session  length $l$) in local mixing, which decides the short-term preference modeling.

\subsubsection{Influence on Performance}
Recall that the historical sequence of length $n$ should be divided into $s$ non-overlapped sessions of length $l$ that $n=l\times s$, we set $(s, l)=\{(1, n), (2, n/2), ... (n, 1)\}$. Note that the local mixing works in the global manner with $(s, l)=(1, n)$, and degrades as the identical mapping with $(s, l)=(n, 1)$.

As shown in Figure \ref{session number}, we find that independently modeling the short-term preferences in shorter sessions improves the performance on most datasets. On \texttt{QB-Article}, \texttt{Brightkite} and \texttt{Yelp}, we observe that modeling the dependency from the global perspective $(s=1)$ is more suitable, while \texttt{Gowalla} and \texttt{TKY} reveal the contrast situation $(s=n)$. The possible reason also lies in the split of historical sequences (as explained in Section \ref{different structures}). Generally, TriMLP achieves the best performance when setting $s=2$ on \texttt{NYC}, \texttt{QB-Video}, \texttt{ML-10M}, $s=8$ on \texttt{ML-1M}, $s=16$ on \texttt{Beauty}, \texttt{Sports} and $s=32$ on \texttt{ML-100K}.

\subsubsection{Influence on Reception Field}\label{reception field}
\begin{figure*}
\centering   
\begin{subfigure}[$s=1, l=128$]{   
\begin{minipage}[b]{0.23\textwidth}   
\includegraphics[width=1\textwidth]{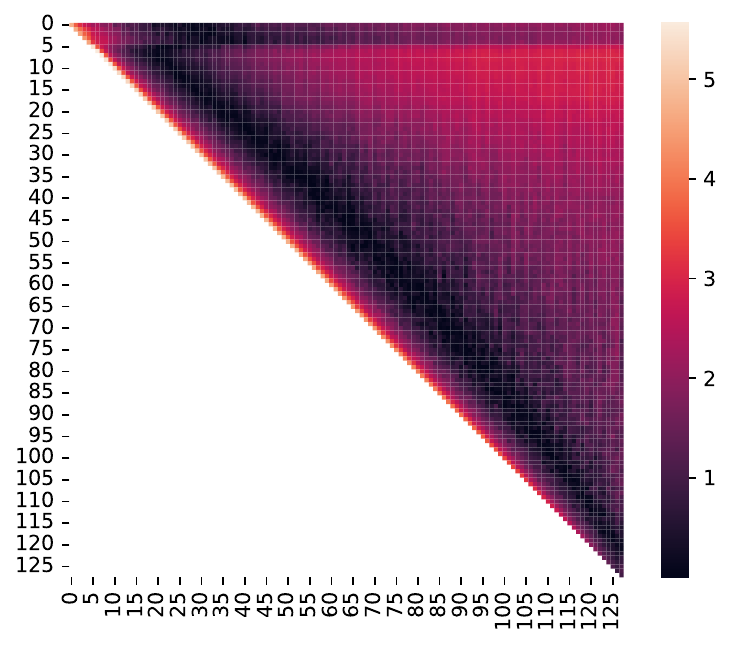} \\   
\includegraphics[width=1\textwidth]{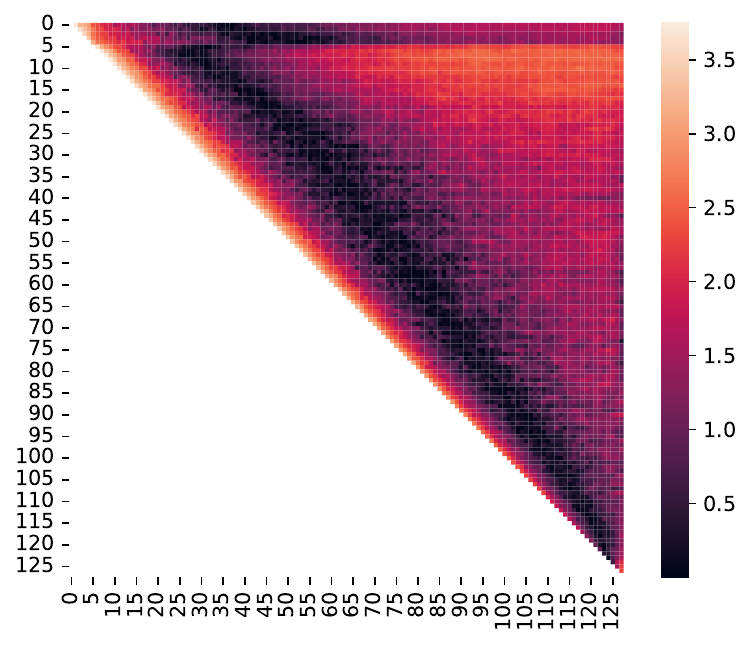}   
\end{minipage}}
\end{subfigure}
\vspace{-20pt}
\medskip
\begin{subfigure}[$s=2, l=64$]{   
\begin{minipage}[b]{0.23\textwidth}   
\includegraphics[width=1\textwidth]{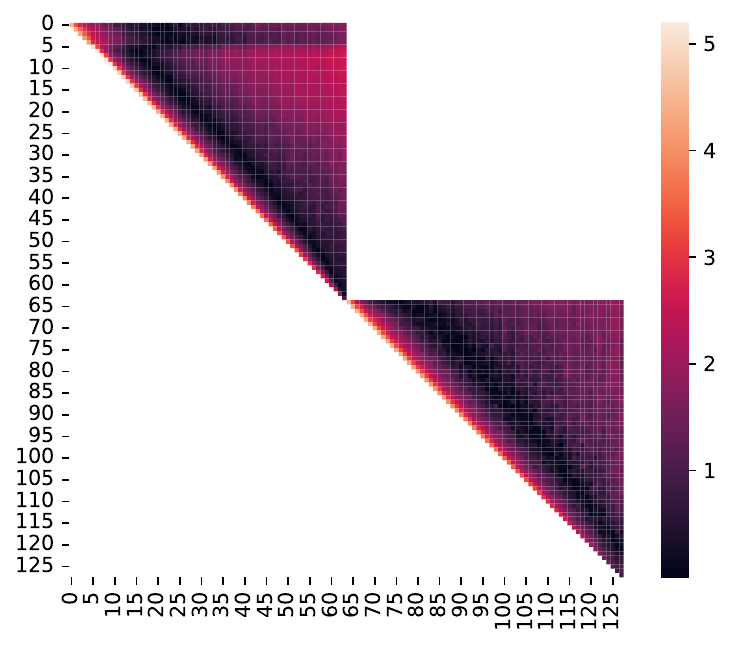} \\   
\includegraphics[width=1\textwidth]{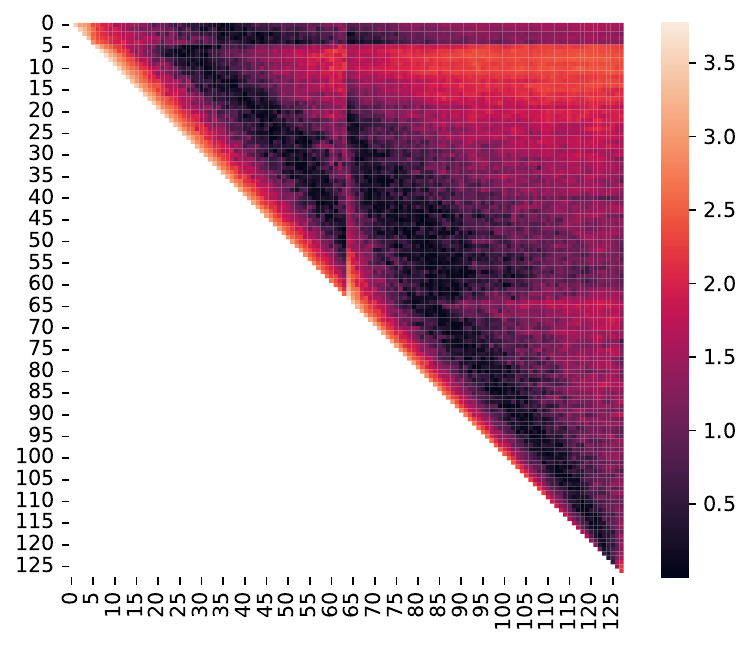}   
\end{minipage}}
\end{subfigure}
\medskip
\begin{subfigure}[$s=4, l=32$]{   
\begin{minipage}[b]{0.23\textwidth}   
\includegraphics[width=1\textwidth]{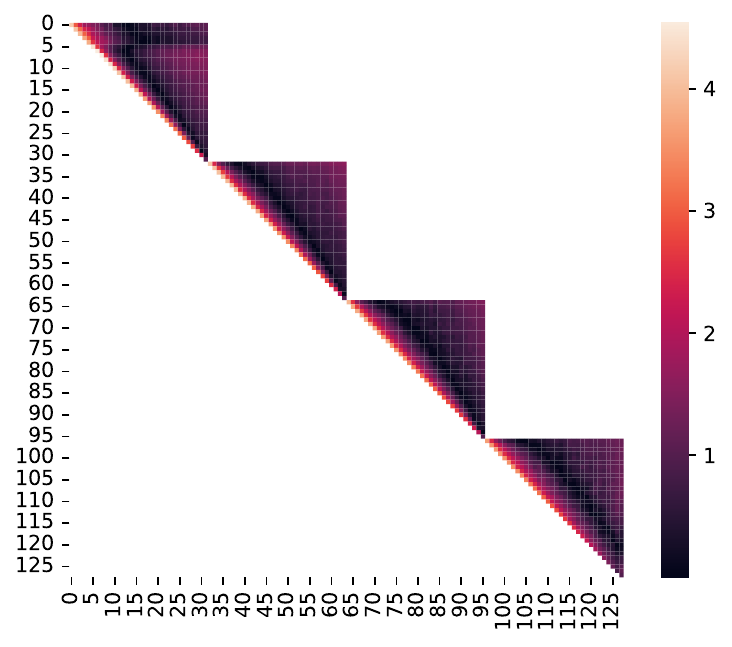} \\   
\includegraphics[width=1\textwidth]{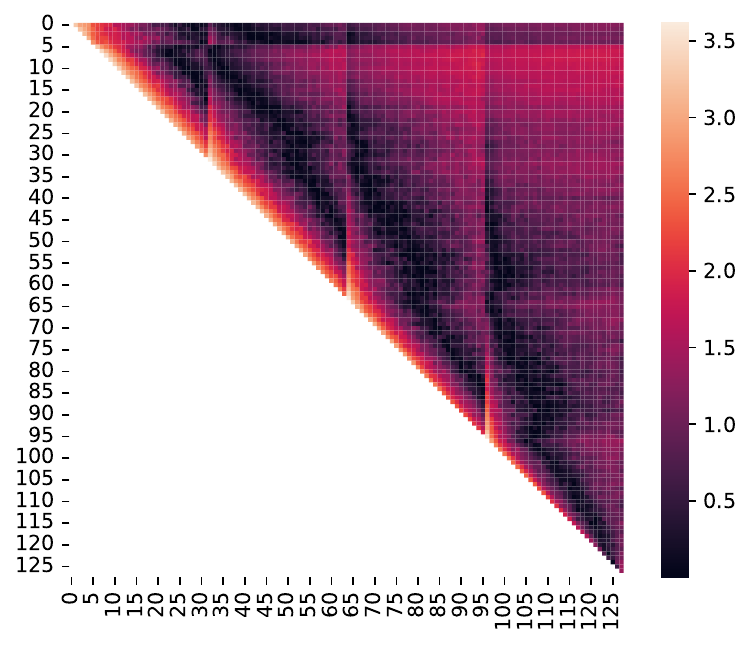}   
\end{minipage}}
\end{subfigure}
\medskip
\begin{subfigure}[$s=8, l=16$]{   
\begin{minipage}[b]{0.23\textwidth}   
\includegraphics[width=1\textwidth]{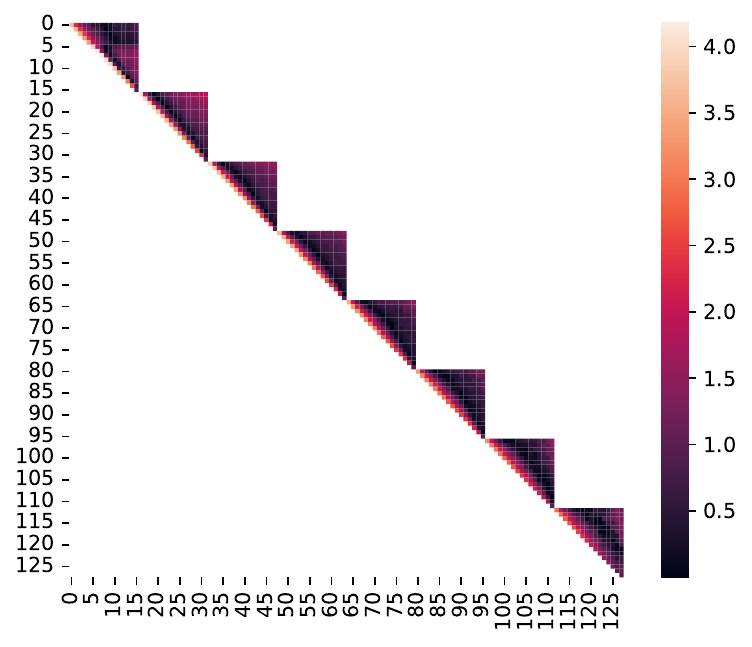} \\   
\includegraphics[width=1\textwidth]{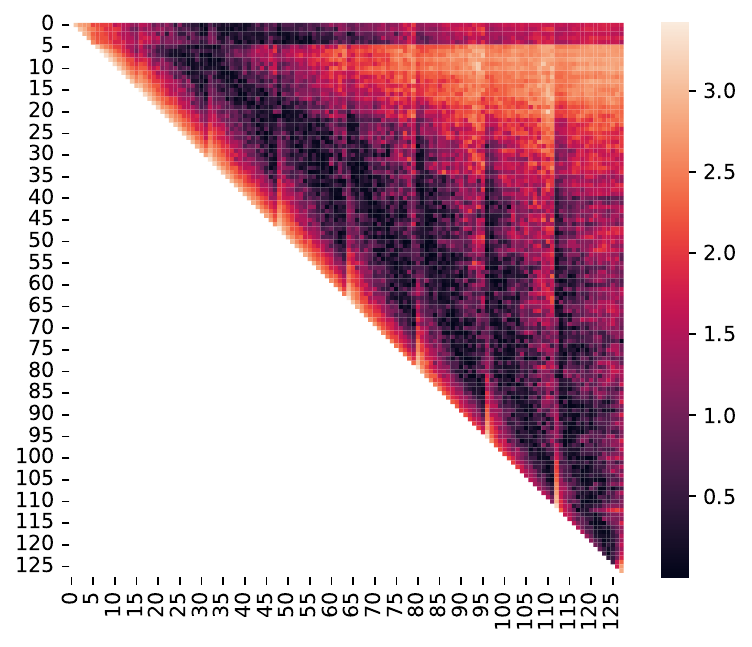}   
\end{minipage}}
\end{subfigure}

\begin{subfigure}[$s=16, l=8$]{   
\begin{minipage}[b]{0.23\textwidth}   
\includegraphics[width=1\textwidth]{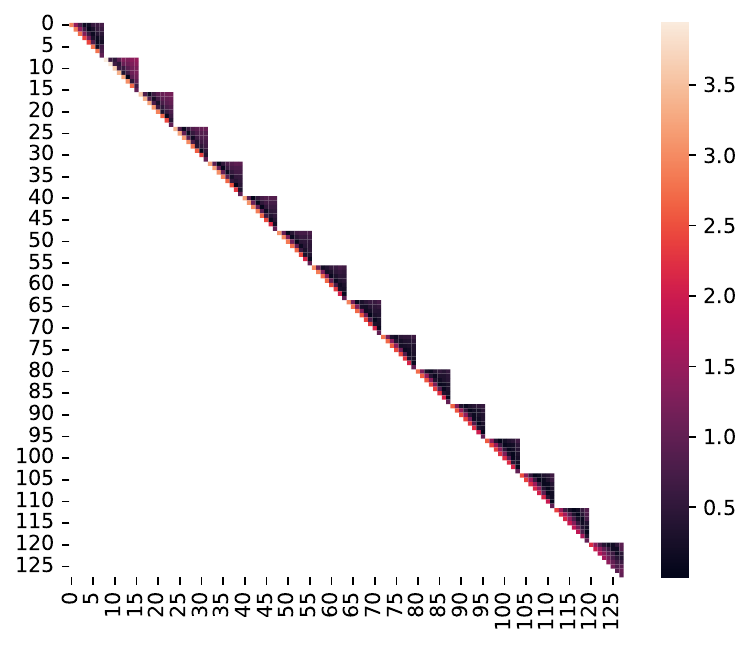} \\   
\includegraphics[width=1\textwidth]{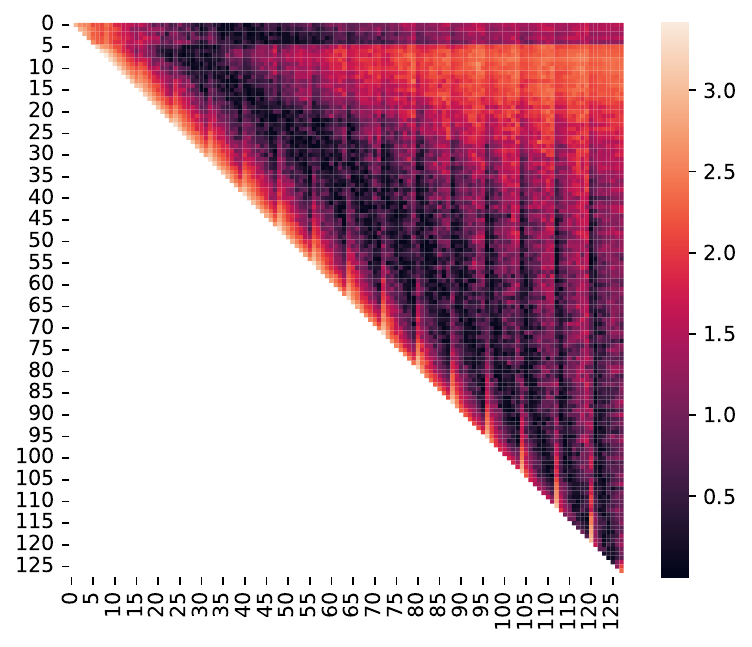}   
\end{minipage}}
\end{subfigure}
\medskip
\begin{subfigure}[$s=32, l=4$]{   
\begin{minipage}[b]{0.23\textwidth}   
\includegraphics[width=1\textwidth]{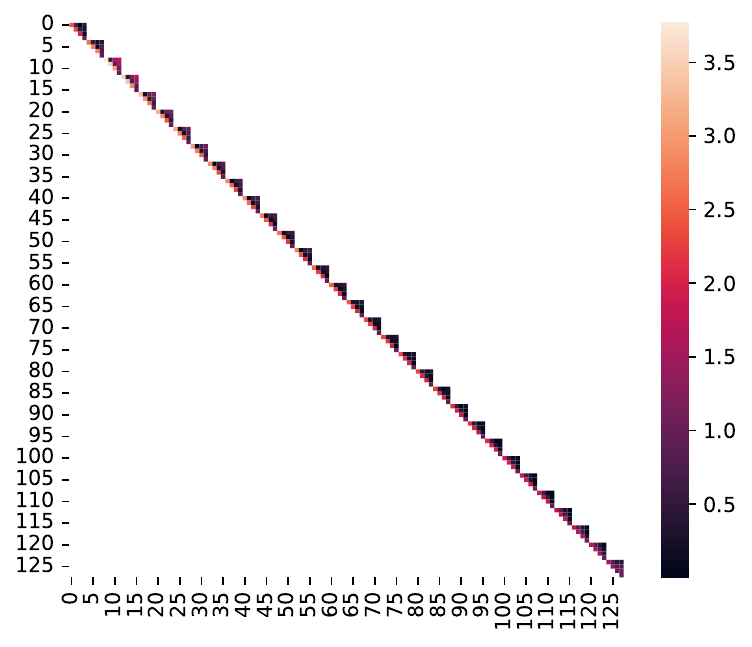} \\   
\includegraphics[width=1\textwidth]{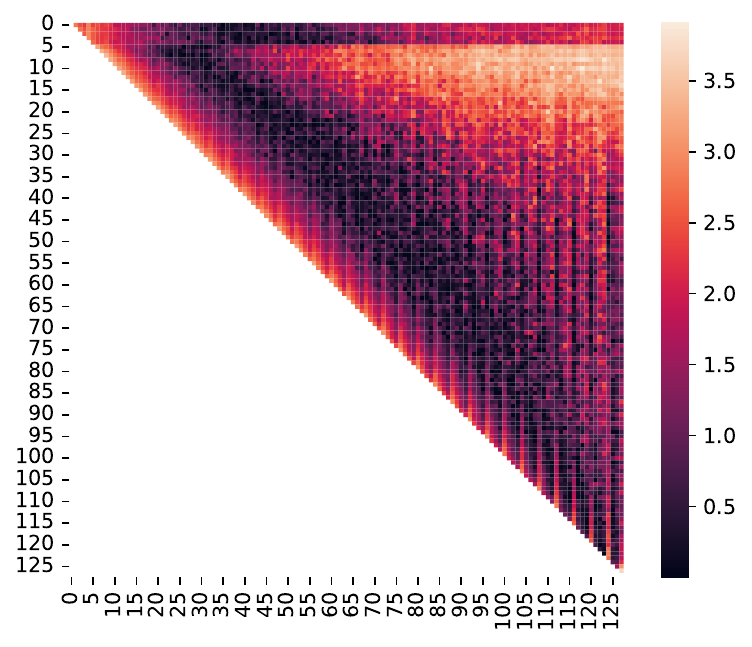}   
\end{minipage}}
\end{subfigure}
\medskip
\begin{subfigure}[$s=64, l=2$]{   
\begin{minipage}[b]{0.23\textwidth}   
\includegraphics[width=1\textwidth]{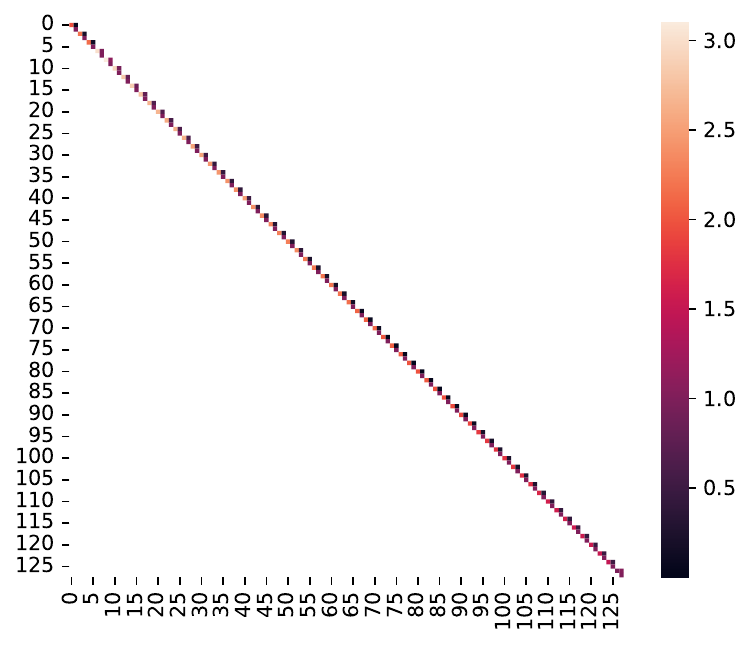} \\   
\includegraphics[width=1\textwidth]{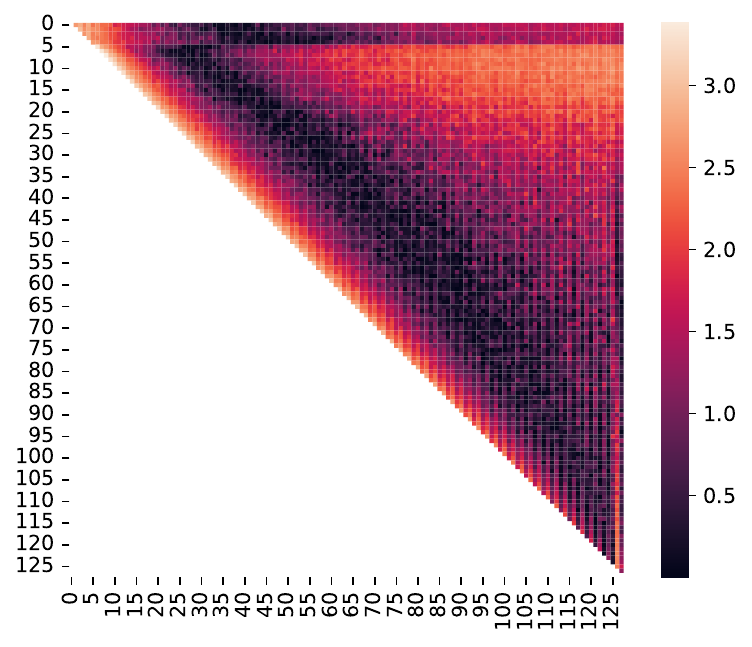}   
\end{minipage}}
\end{subfigure}
\medskip
\begin{subfigure}[$s=128, l=1$]{   
\begin{minipage}[b]{0.23\textwidth}   
\includegraphics[width=1\textwidth]{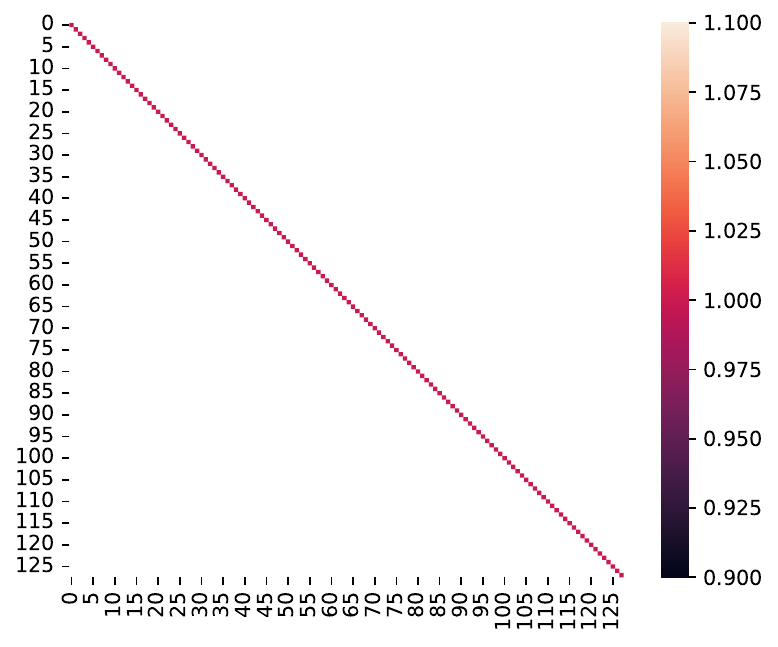} \\   
\includegraphics[width=1\textwidth]{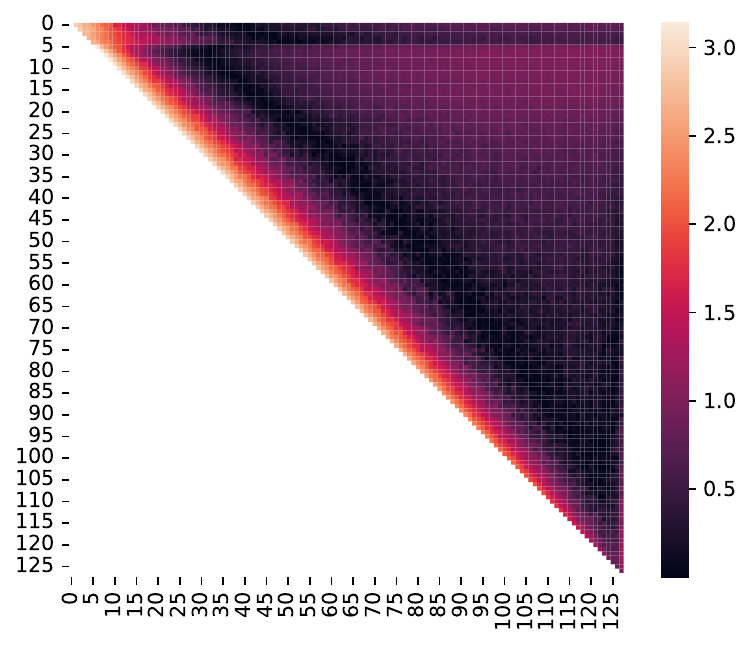}   
\end{minipage}}
\end{subfigure}
\vspace{-20pt}
\caption{Visualizing the weights of global and local mixing kernels with different session number $s$ on \texttt{ML-1M}. The global and local kernel are separately plotted in the upper and lower part of each sub-figure. Black indicates that the weight is 0, and the brighter, the greater the weight's absolute value.}
\vspace{-15pt}
\label{heatmap} 
\end{figure*}

We visualize the weights of global and local kernels on \texttt{ML-1M} $(n=128)$ to explore how these two mixing layers complement each other. The corresponding 8 heat maps with different session numbers $s$ (or session lengths $l$) are plotted in Figure \ref{heatmap}. Accordingly, we observe the following characteristics:

\textit{Characteristic 1: Local kernels sustain more attention on the tokens around the current time step.} We observe that the weights in all local kernels share the similar distribution, that the elements nearing the diagonal have greater absolute values than others, i.e., more active. When setting $s=n=128$ (Figure \ref{heatmap} (h)), the local kernel completely degrades into the identity matrix where all remaining diagonal elements are 1.

\begin{figure}[H]
\centering
	\subfigure[Sports]{
	\includegraphics[scale=0.14725]{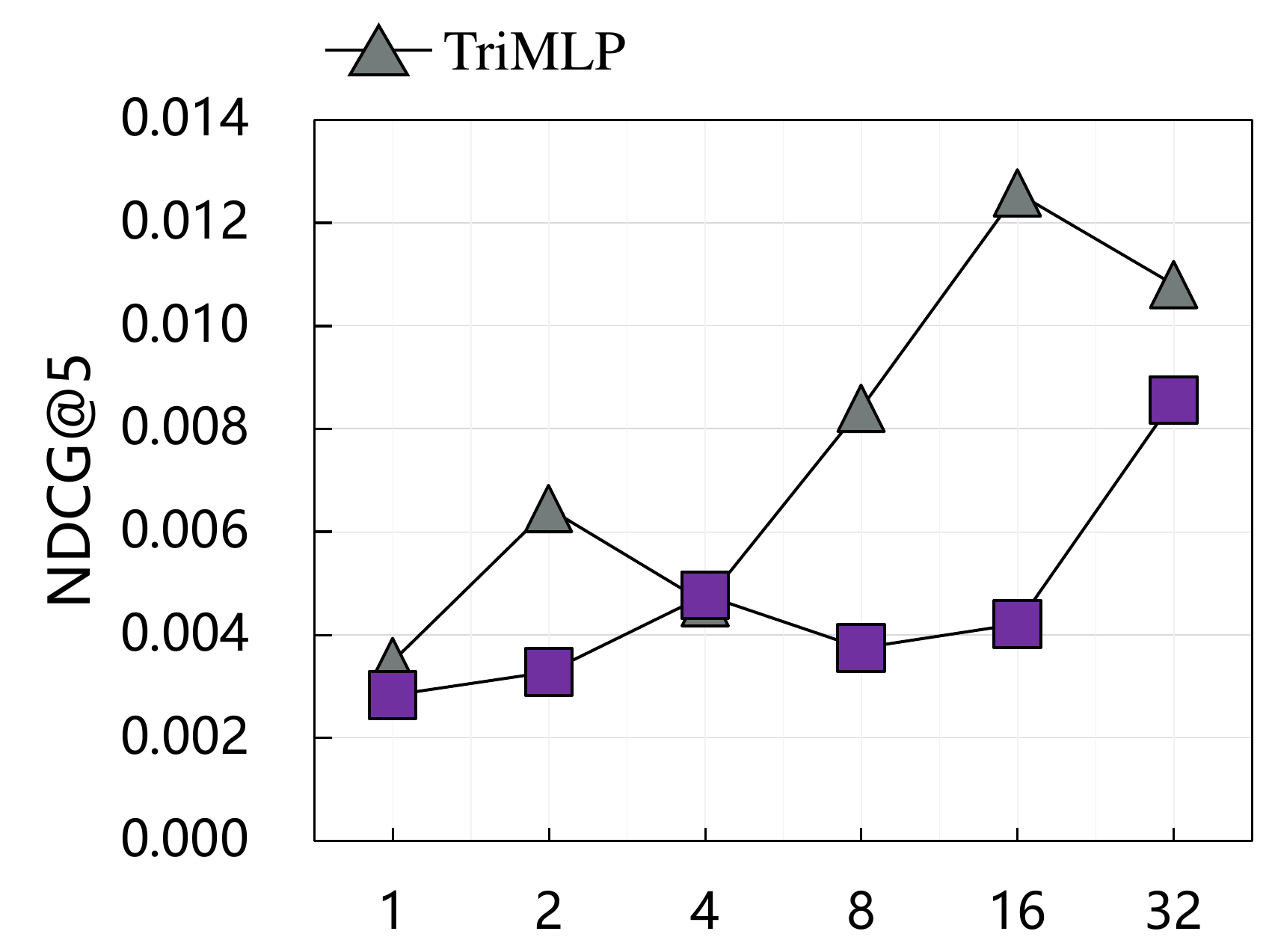}}
	\vspace{-10pt}
	\subfigure[QB-Video]{
	\includegraphics[scale=0.14725]{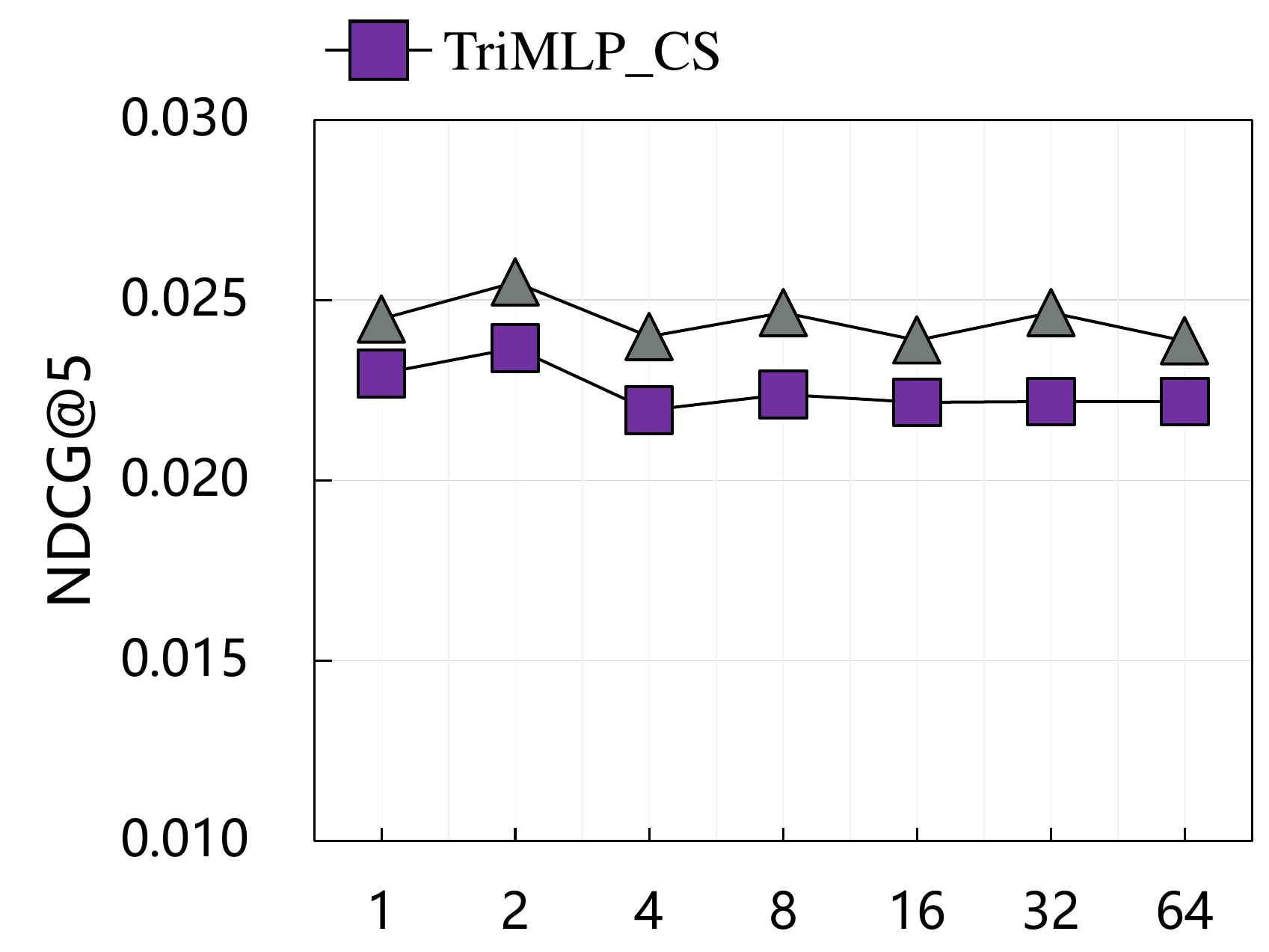}}
	\subfigure[ML-100K]{
	\includegraphics[scale=0.14725]{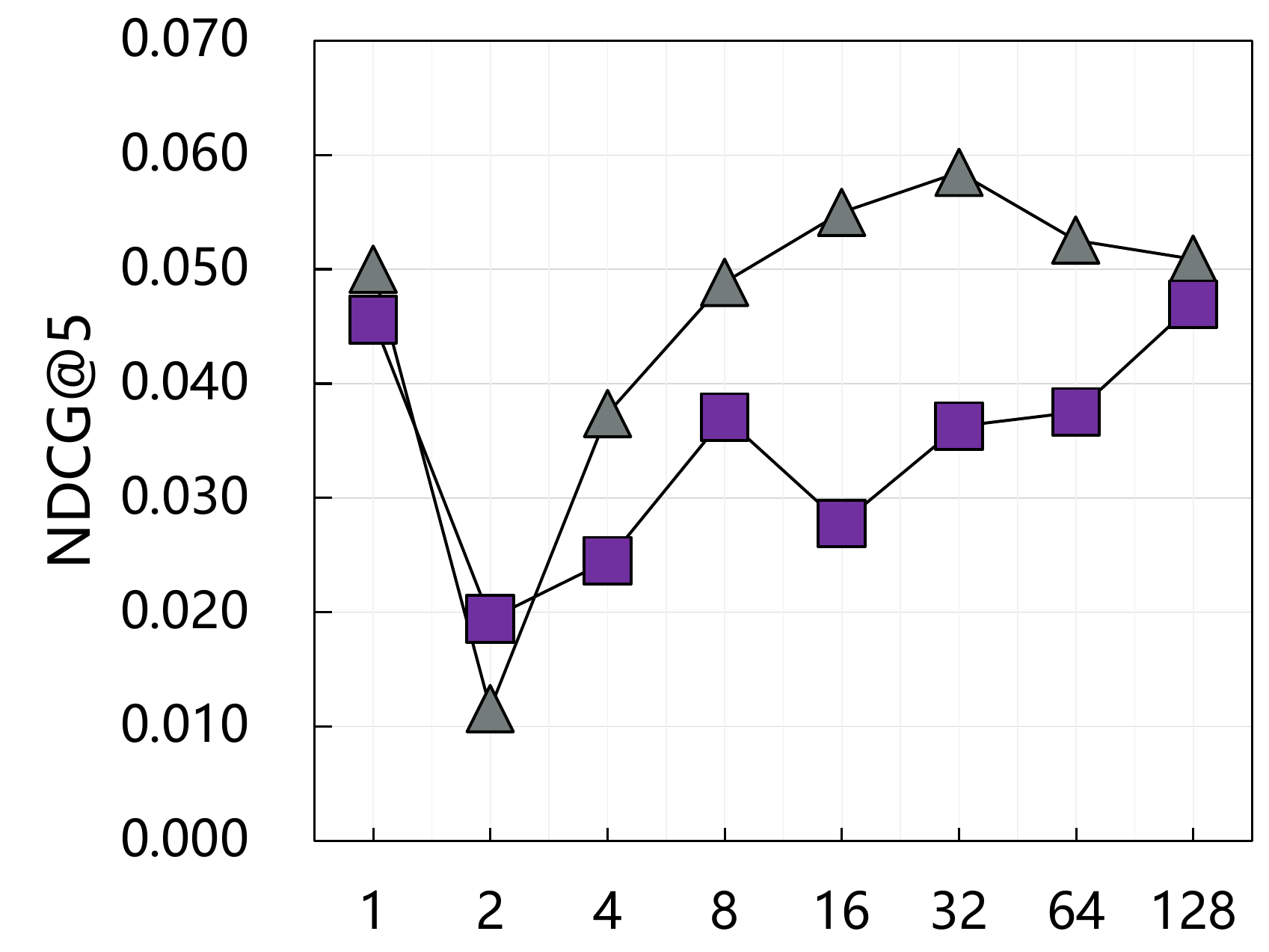}}
	\subfigure[Brightkite]{
	\includegraphics[scale=0.14725]{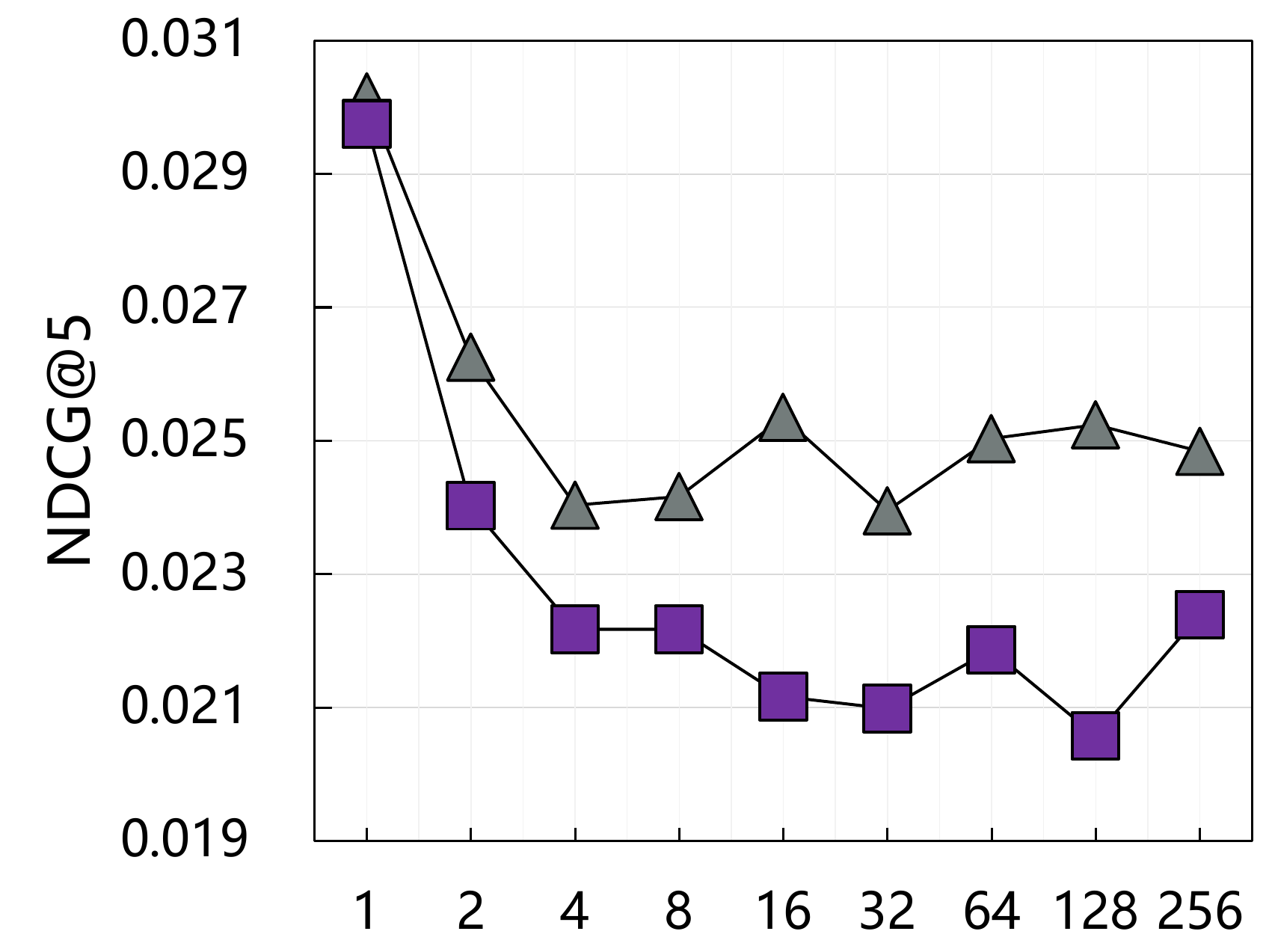}}
\caption{NDCG@5 comparison between w./w.o. cross-session interactions. The axes of all sub-figures stand for the variable session number $s$.}
\vspace{-10pt}
\label{cross}
\end{figure}

\textit{Characteristic 2: Shorter sessions encourage the global kernel attaching more importance to previous tokens.} Compared to Figure \ref{heatmap} (a), the upper-right elements in global kernels have greater absolute values (as Figure \ref{heatmap} (d)-(g)), which are responsible for the long-term user-item interactions. It indicates that the shorter sessions call for the larger reception field in global kernels to support the sufficient sequential dependency modeling. Unfortunately, such pattern seems to lose efficacy with $s=\{2, 4, 128\}$ (as Figure \ref{heatmap} (b, c, h)), i.e., global kernels are no longer sharp to long-range tokens, which accord to the inferior performances in Figure \ref{session number} (g).

\textit{Characteristic 3: Suitable session settings produce superior performances.} As the experimental results in Figure \ref{session number} (g), TriMLP achieves comparable and preferred scores when setting $s=\{8, 32\}$ on \texttt{ML-1M}. Combined with the corresponding heatmaps in Figure \ref{heatmap} (d, f), besides the adequate short-term patterns offered by the local kernels, both of the global kernels in these two cases are more perceptive to the previous tokens. It proves the effectiveness of the serial structure in Triangular Mixer, that the global mixing layer and local mixing layer mutually assit each other indeed to realize the fine-grained modeling of sequential dependency.

\subsection{Cross-session Communications in Local Mixing}\label{cross session variant}
Recall another variant of local mixing that endows the cross-session communications (Eq. \ref{cross session local mixing}), denoted as TriMLP\_CS, we verify the performance on 4 datasets \texttt{Sports}, \texttt{QB-Video}, \texttt{ML-100K} and \texttt{Brightkite}, where the maximum sequence length $n$ ranges from 32 to 256. TriMLP and TriMLP\_CS share the same local mixing layer when $s=\{1, n\}$.

Figure \ref{cross} reports the NDCG@5 scores correlated with different session number $s$. It shows that connection sessions decreases the performance. The possible reasons are two-fold. On the one hand, cross-session connections bring more previous information to the current overlapped token after multiple iterations, and degrade the local mixing as the global scheme, which disables the short-term preference modeling. On the other hand, since such connections are build solely upon the overlapped tokens, the corresponding neurons in the mixing kernel might accumulate more errors especially when the static weights are agnostic to the input sequence, which leads to the biased local pattern.

\subsection{Auto-regressive V.S. Auto-encoding}\label{atrg vs atec}
Auto-encoding is the other popular training fashion, represented by \cite{bert, bert4rec}, that utilizes the past and future tokens to predict the current one. To verify whether or not the auto-encoding is compatible with MLP, we derive another variant BiMLP which stacks 2 MLP layers as the basic architecture, and train it on 4 datasets of different scales \texttt{Beauty}, \texttt{NYC}, \texttt{ML-1M} and \texttt{Yelp} under the auto-encoding manner. Specifically, we randomly mask the tokens in historical sequences with setting mask ratio $r=\{0.1, 0.2, 0.3, 0.4, 0.5, 0.6, 0.7, 0.8, 0.9\}$. Among these masked tokens, 80\% are padded with \texttt{mask token}, 10\% are replaced with other tokens and 10\% reserve the original ones.

\begin{figure}[H]
\centering
	\subfigure[Beauty]{
	\includegraphics[scale=0.14725]{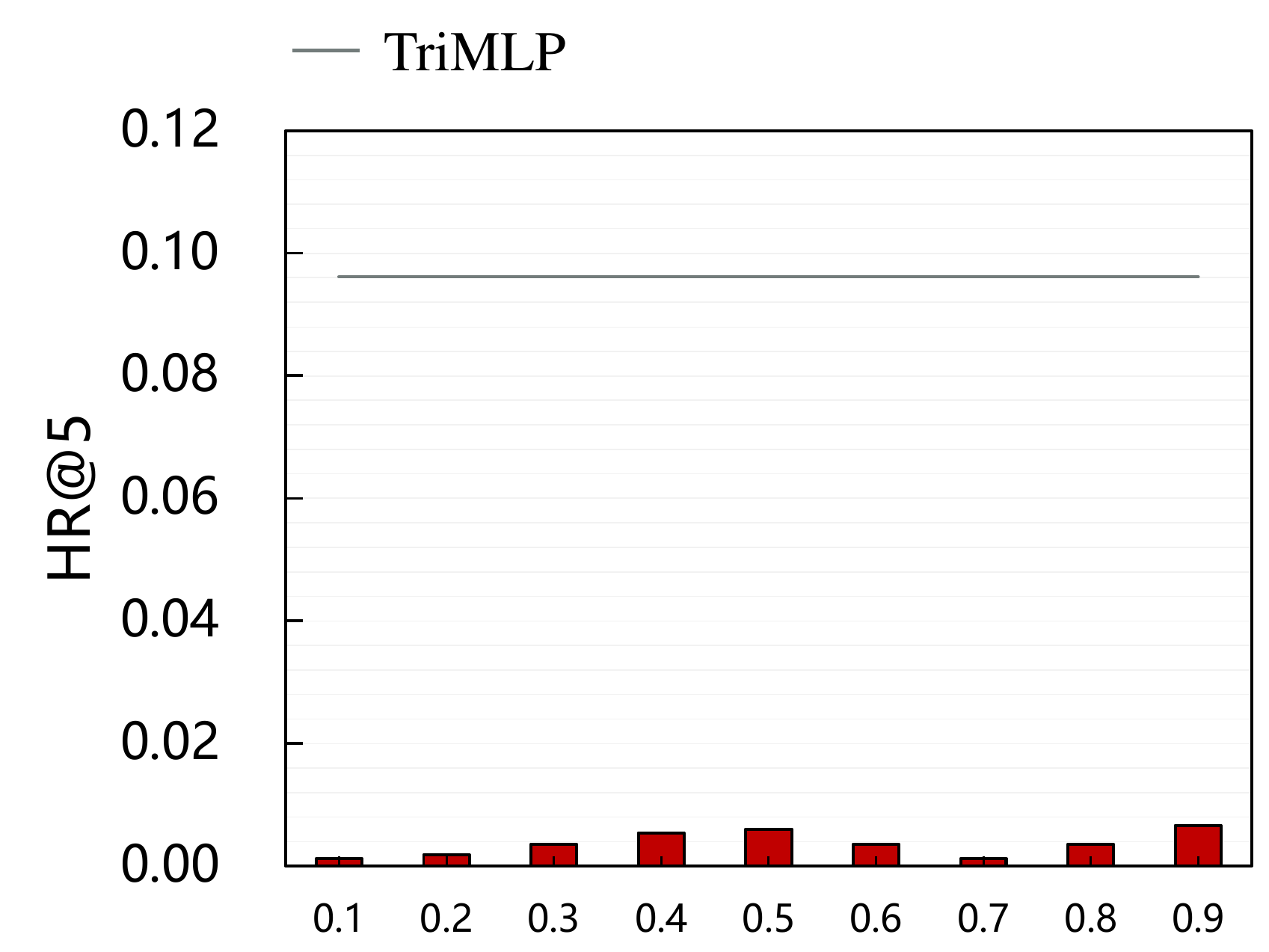}}
	\vspace{-10pt}
	\subfigure[NYC]{
	\includegraphics[scale=0.14725]{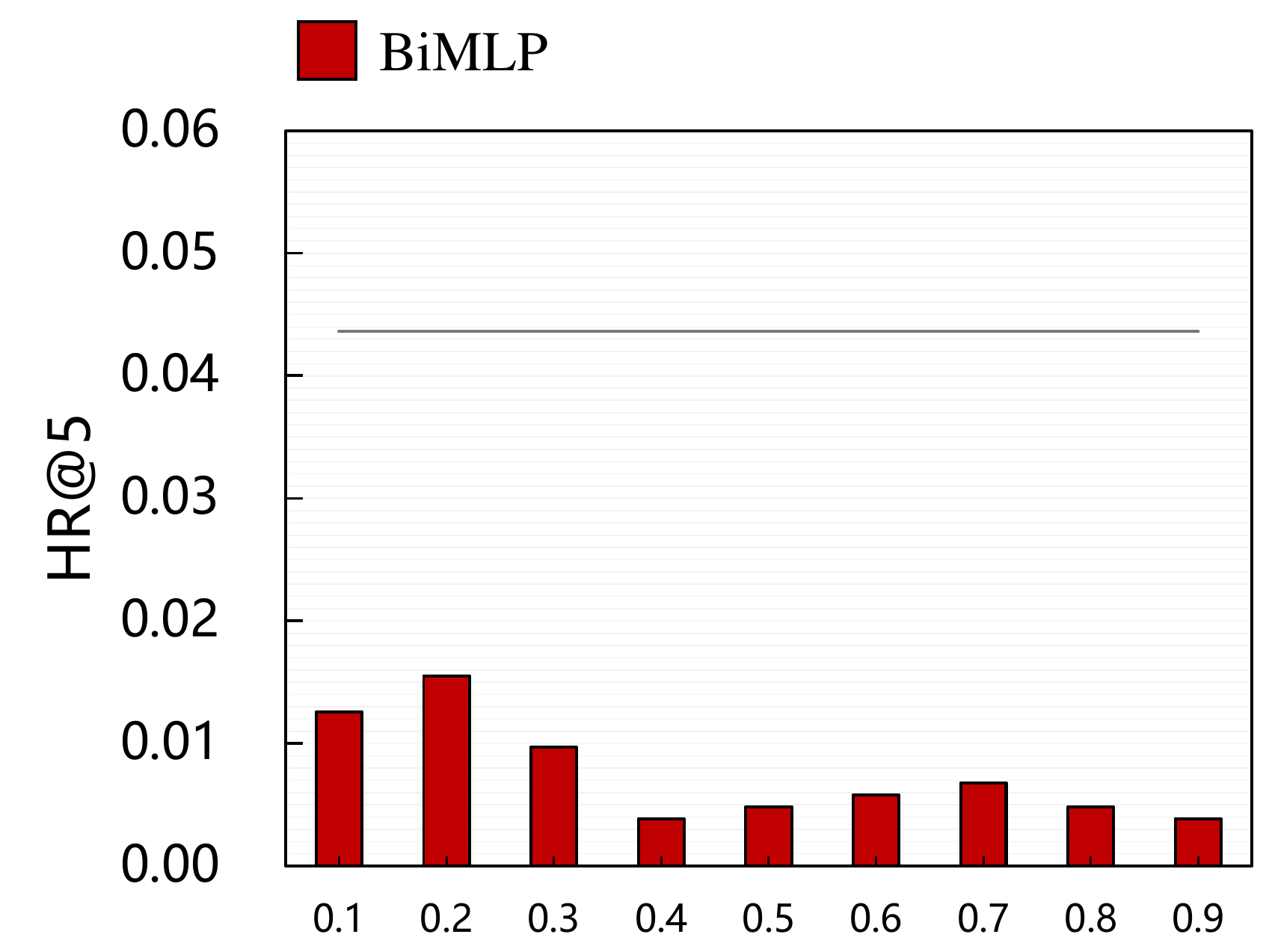}}
	\subfigure[ML-1M]{
	\includegraphics[scale=0.14725]{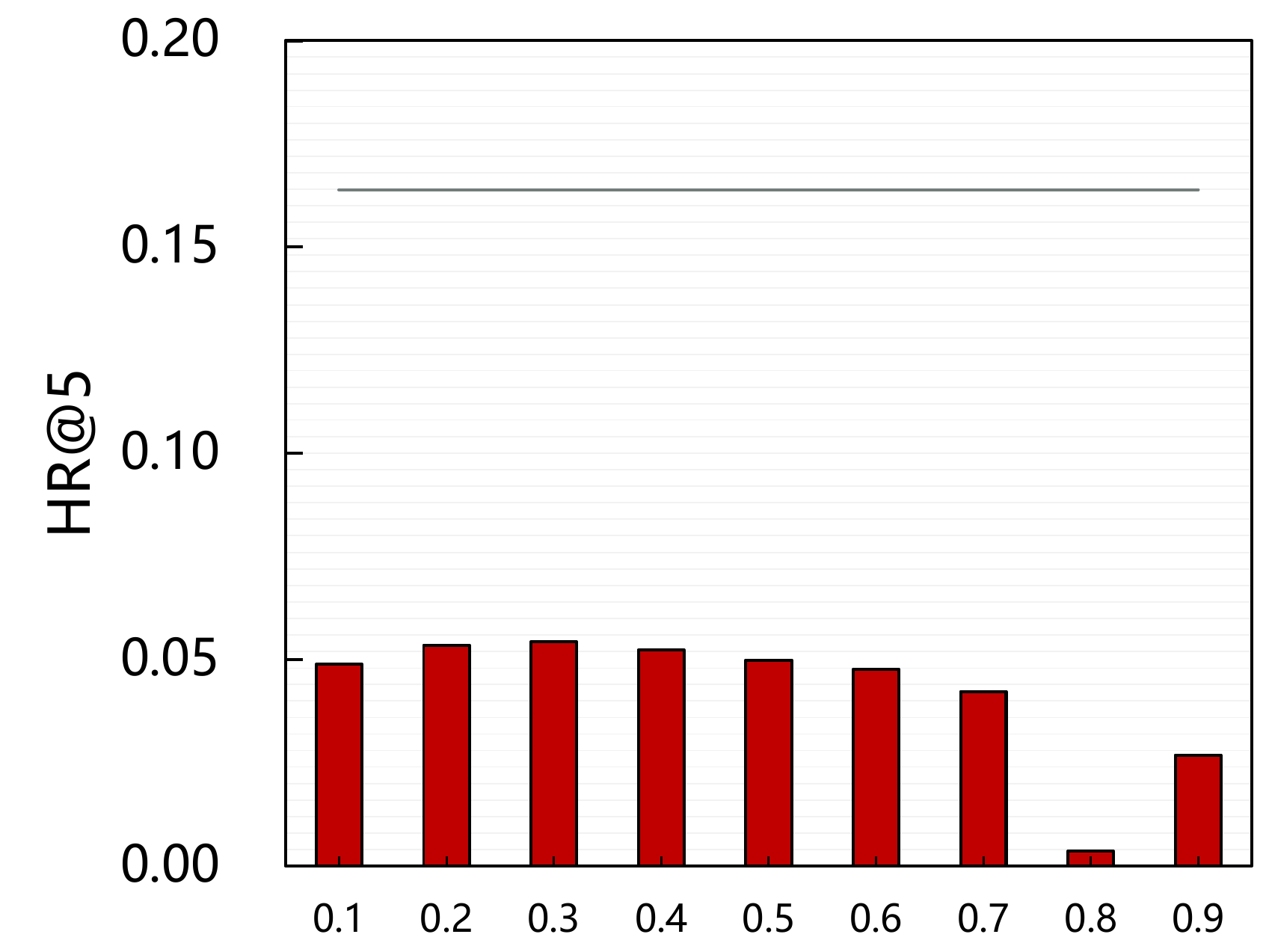}}
	\subfigure[Yelp]{
	\includegraphics[scale=0.14725]{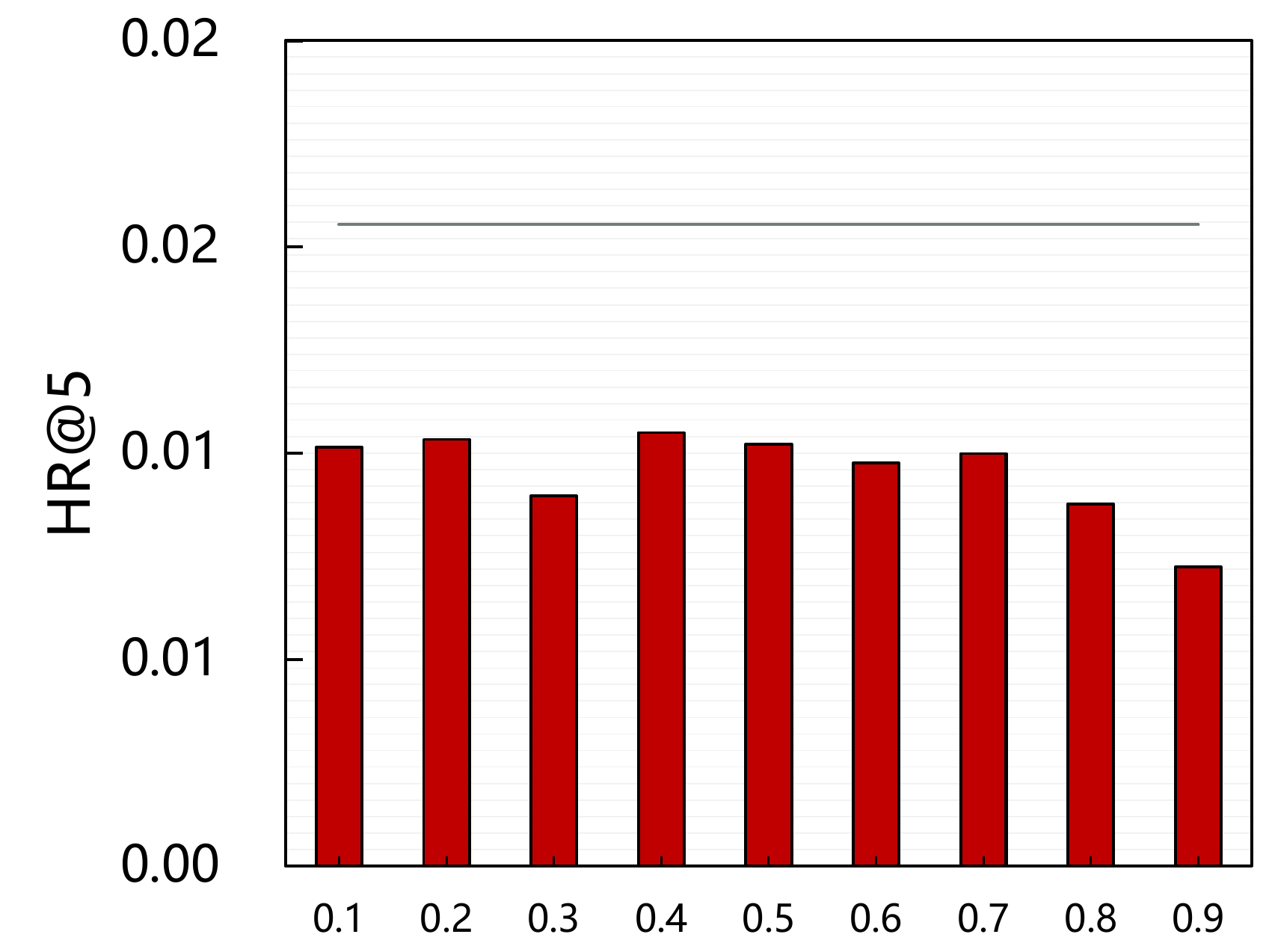}}
\caption{HR@5 comparison between auto-regressive and auto-encoding. The axes of all sub-figures denote the variable mask ratio $r$.}
\label{atec}
\end{figure}

Figure \ref{atec} reports the HR@5 scores correlated with different mask ratio. We find that the recommendation performance of BiMLP stands far behind TriMLP. The possible reason lies in that the auto-encoding training mode is more inclined to suffer from the data-hungry issue. It is more compatible with dense datasets for learning better mask representations, while sequential recommendation datasets are always extremely sparse (sparsity is usually larger than 99\%). Moreover, this observation is also in line with \cite{tenrec, f-mlp4rec, bi1, bi2}, that unidirectional models offer better results than bidirectional ones. Thus, we exploit the triangular design and put forward TriMLP under the unidirectional auto-regressive training scheme.

\section{Conclusion}
In this paper, we make our exploration to study the capacity of MLP in sequential recommendation. We present the MLP-like sequential recommender TriMLP with a novel Triangular Mixer. Credited to the chronological cross-token communication and the serial mixing structure in Triangular Mixer, TriMLP successfully realizes the fine-grained modeling of sequential dependency. The experimental results on 12 datasets demonstrate that TriMLP attains stable, competitive and even better performance than several state-of-the-art baselines under the essential auto-regressive training mode with prominent less inference time, which well performs the ``Revenge of MLP in Sequential Recommendation''.

In the future, we will further improve TriMLP by introducing auxiliary information like temporal factors and item attributes, data augmentation and pre-training techniques. Moreover, it is intriguing to deliberate how to decouple the strong correlation between the sequence length and MLP shape, which will enable MLP to flexibly handle sequences of different lengths.

\bibliographystyle{IEEEtran}
\bibliography{ieeeart}
\end{document}